\documentclass[runningheads]{llncs}

\usepackage[mobile]{eccv}

\usepackage{eccvabbrv}

\usepackage{graphicx}
\usepackage{booktabs}

\usepackage[accsupp]{axessibility}  %

\usepackage{hyperref}

\usepackage{orcidlink}

\usepackage{graphicx}
\usepackage{amsmath}
\usepackage{amssymb}
\usepackage{booktabs}
\usepackage{subfiles}
\graphicspath{{imgs/}{../imgs/}}

\usepackage{bbm}

\usepackage{url}

\usepackage{float}
\floatstyle{plaintop}
\restylefloat{table}

\usepackage{esvect}
\usepackage{booktabs,makecell}
\usepackage{colortbl}
\usepackage{xcolor}
\usepackage{wrapfig}
\usepackage{float}
\restylefloat{table}
\usepackage{placeins}
\usepackage{stfloats}
\usepackage{multicol}
\usepackage{tabu,multirow}
\usepackage{tabularx}
\usepackage{nicematrix}
\usepackage{booktabs}
\usepackage{placeins}
\usepackage{booktabs}
\usepackage{footnote}
\usepackage{afterpage}
\usepackage{adjustbox}
\usepackage{arydshln}
\usepackage{gensymb}
\usepackage{caption}
\usepackage{subcaption}

\usepackage{mathtools}  

\usepackage{siunitx} %

\usepackage{pifont}%

\definecolor{lightgrey}{rgb}{0.7, 0.7, 0.7}

\newcolumntype{"}{@{\hskip\tabcolsep\vrule width 1.3pt\hskip\tabcolsep}}

\newcolumntype{P}[1]{>{\centering\arraybackslash}m{#1}}

\let\svthefootnote\thefootnote
\newcommand\freefootnote[1]{%
  \let\thefootnote\relax%
  \footnotetext{#1}%
  \let\thefootnote\svthefootnote%
}

\usepackage{tikz}

\newcommand{\ourmethod}{6DGS\xspace}
\newcommand{\raygenerationmethod}{Ellicell\xspace}
\newcommand{\raygenerationmethodplural}{Ellicells\xspace}

\begin{document}

\title{\ourmethod: 6D Pose Estimation from a Single \\Image and a 3D Gaussian Splatting Model}

\titlerunning{\ourmethod: 6D Pose Estimation from a Single Image and a 3DGS Model}
\author{Bortolon Matteo\inst{1, 2, 3}\orcidlink{0000-0001-8620-1193} \and
Theodore Tsesmelis\inst{1}\orcidlink{0000-0001-9290-2383} \and
Stuart James\inst{1,4}\orcidlink{0000-0002-2649-2133} \and \\
Fabio Poiesi\inst{2}\orcidlink{0000-0002-9769-1279} \and
Alessio {Del Bue}\inst{1}\orcidlink{0000-0002-2262-4872} }

\institute{PAVIS, Fondazione Istituto Italiano di Tecnologia (IIT), Genoa, IT \and
TeV, Fondazione Bruno Kessler (FBK), Trento, IT \and
Università di Trento, Trento, IT \and
Durham University, Durham, UK}

\authorrunning{M.~Bortolon \etal}

\maketitle
\begin{abstract}
We propose \ourmethod to estimate the camera pose of a target RGB image given a 3D Gaussian Splatting (3DGS) model representing the scene.
\ourmethod avoids the iterative process typical of analysis-by-synthesis methods (\eg iNeRF) that also require an initialization of the camera pose in order to converge.
Instead, our method estimates a 6DoF pose by inverting the 3DGS rendering process. Starting from the object surface, we define a radiant \raygenerationmethod that uniformly generates rays departing from each ellipsoid that parameterize the 3DGS model. 
Each \raygenerationmethod ray is associated with the rendering parameters of each ellipsoid, which in turn is used to obtain the best bindings between the target image pixels  and the cast rays.
These pixel-ray bindings are then ranked to select the best scoring bundle of rays, which their intersection provides the camera center and, in turn, the camera rotation.
The proposed solution obviates the necessity of an \textit{``a priori''} pose for initialization, and it solves 6DoF pose estimation in closed form, without the need for iterations. %
Moreover, compared to the existing Novel View Synthesis (NVS) baselines for pose estimation, \ourmethod can improve the overall average rotational accuracy by $\mathbf{12\%}$ and translation accuracy by $\mathbf{22\%}$ on real scenes, despite not requiring any initialization pose. 
At the same time, our method operates near real-time, reaching $\mathbf{15} fps$ on consumer hardware.
\freefootnote{Project page: \url{https://mbortolon97.github.io/6dgs/}\\
Corresponding author: \email{mbortolon@fbk.eu}}

\end{abstract}

\section{Introduction}\label{sec:intro}

\begin{figure}[t]
\centering
\includegraphics[width=\linewidth]{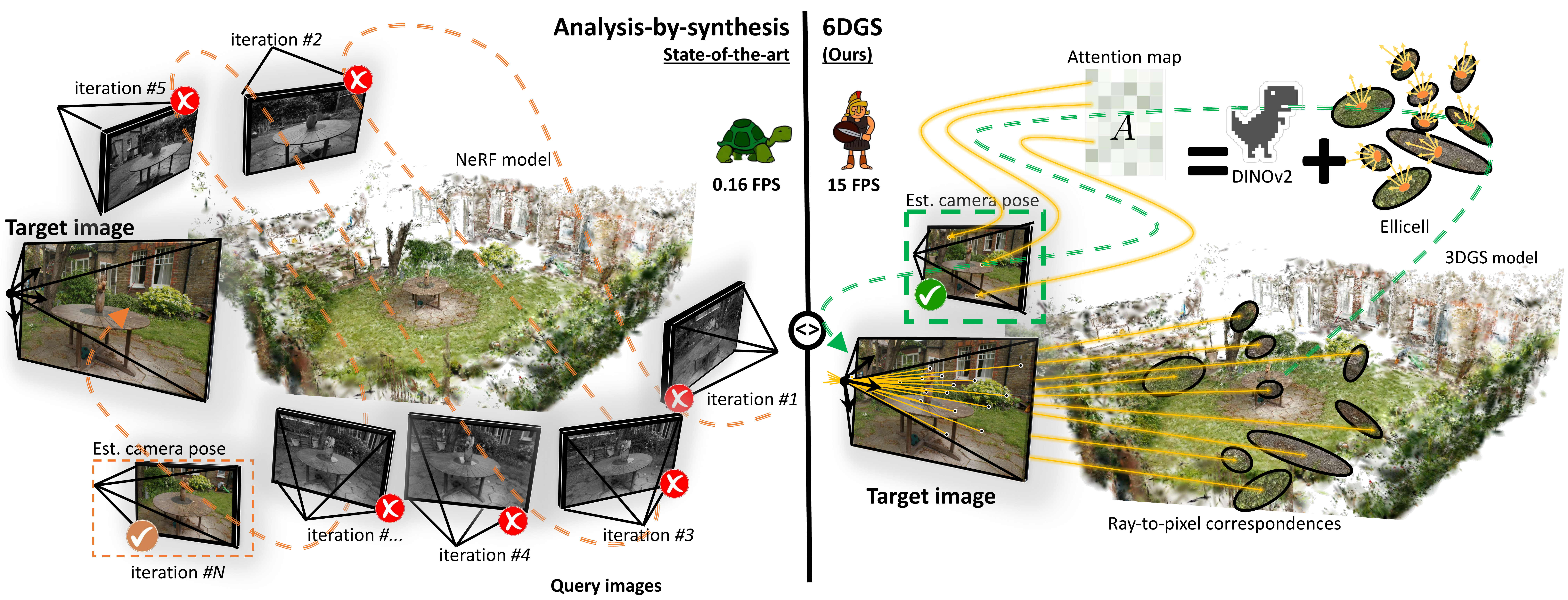}
\vspace{-1.0em}
\caption{
Our 6DGS method introduces a novel approach to 6DoF pose estimation, departing from conventional analysis-by-synthesis methodologies. 
Standard NeRF-based methods \textbf{(left)}  employ an iterative process, rendering candidate poses and comparing them with the target image before updating the pose, which often results in slow performance and limited precision. 
In contrast, 6DGS \textbf{(right)} estimates the camera pose by selecting a bundle of rays projected from the ellipsoid surface (a \textit{radiant Ellicell}) and learning an attention map to output ray/image pixel correspondences (based on DINOv2). The optimal bundle of rays should intersect the optical center of the camera and then are used to estimate the camera rotation in closed-form. %
Our 6GDS method offers significantly improved accuracy and speed, enabling the recovery of the pose within a one-shot estimate.
}
\label{fig:teaser}
\vspace{-1.0em}
\end{figure}

Neural and geometrical 3D representations for Novel View Synthesis (NVS) have recently surged in popularity ~\cite{mildenhall2020nerf,kerbl20233Dgaussians}, and they have been quickly integrated into daily applications, \eg mapping services \cite{Gmaps}.
The change in 3D representation creates new challenges on how to solve classical problems, such as 6D pose estimation, and on how to leverage NVS implicit advantages~\cite{yen2020inerf, lin2023parallelinerf, moreau2023crossfire, wang2022voge, maggio2023verf}.

The method of iNeRF~\cite{yen2020inerf} pioneered 6D pose estimation using an NVS model by proposing an iterative analysis-by-synthesis, as illustrated in the left panel of Fig.~\ref{fig:teaser}.
Given a nearby pose initialization (iteration \#\textit{1}), the NVS model is used to render the image related to the initial pose.
Then iteratively, the rendered image is compared with the target image using a photometric loss, and the initial pose guess is updated so that the two views achieve the best image overlap at the final step (iteration \#\textit{N}).
The authors in iNeRF~\cite{yen2020inerf} use the popular NeRF~\cite{mildenhall2020nerf} NVS model where backpropagation updates every new pose guess. This procedure leverages the remarkable NeRF capabilities in synthesizing realistic novel views, however, at the computational expense of synthesizing a newly rendered image at each iteration.
This limitation restricts iNeRF to offline use while requiring a close initial pose estimate for a successful convergence.

Recent works in 3D Gaussian Splatting (3DGS)~\cite{kerbl20233Dgaussians, luiten2023dynamic3DGS, xie2023physgaussian} are an alternative to Neural NVS models, providing fast rendering speed through the use of explicit geometric primitives that do not require the optimization of a neural network. %
3DGS represents a 3D scene as a set of ellipsoids paired with photometric information, such as color and opacity.
The ellipsoids are first initialized using Structure from Motion (SfM), and then they are optimized to reduce the photometric error between the rasterized ellipsoids and a set of known images.
During the rasterization stage, the 3DGS model is projected onto the image plane as ellipses and for each pixel the algorithm computes its photometric contribution.

By leveraging the 3DGS model properties, we design a novel 6DoF pose estimation method (6DGS) that surpasses the limitations of NeRF-based iterative approaches. 
6DGS does not require any pose initialization, and it estimates the camera translation and rotation without an iterating analysis-by-synthesis walkthrough. 
This is a key factor for achieving near real-time performance (15\textit{fps}), also due to the quick rendering capabilities of 3DGS.
The right panel of Fig. \ref{fig:teaser} presents the gist of our approach for 6DoF pose estimation.
If we knew the camera pose, the first NVS step of 3DGS would be to project the ellipsoid centers onto the image plane. 
Practically, this is a ray casting through the camera's optical center. Our 6DGS attempts to invert this process and, by doing so, to estimate the camera pose. 
If the target image camera pose is unknown, and thus neither where the optical center is, we are unable to cast the single ray from each ellipsoid that passes through the correct target image pixels. 
For this reason, instead, we radiate uniformly distributed rays from each ellipsoid through the introduction of a novel casting procedure named \raygenerationmethod. 
Only one radiated ray per ellipsoid would be accurate, \ie, the one that renders the pixel photometrically by projecting the correct ellipse onto the target image plane.

Now, the 6DGS problem is to select, given all the casted rays from the Ellicells, the correct bundle of rays that can generate most of the target image pixels with high confidence. 
This selection stage is addressed by binding pixels and rays through the learning of an attention map.
Notice that this step is also unsupervised, as it leverages the known camera poses and images used to compute the 3DGS model to obtain the pixel and ray pairs used for training.
After the bundle of rays is selected, the intersection of these rays identifies the camera center, which is solved using weighted Least Squares (wLS), with the weights being the scores from the previous selection stage.
After the optical center is estimated, the optical axis can be used to obtain the camera rotation degrees of freedom from the rays bundle, thus solving the 6DoF pose. 
By design, \ourmethod eliminates the need for an initial camera pose, which is one of the limitations of analysis-by-synthesis pose estimation methods~\cite{yen2020inerf, moreau2023crossfire, wang2022voge}, as well as the tendency to converge to local minima during the iteration procedure, especially if the initial pose is initialized far from the optimal position.

We evaluate \ourmethod on datasets featuring real-world objects and scenes, comparing against the current NVS state-of-the-art approaches such as iNeRF~\cite{yen2020inerf}, Parallel iNeRF~\cite{lin2023parallelinerf} and NeMO + VoGE~\cite{wang2022voge}. 
Our experimental results show that 6DGS is competitive, especially if the initial pose is not provided \textit{``a priori''}.
Finally, we achieve near real-time 6DoF pose estimation on consumer hardware, which is one rather challenging limitation in the practical application of NVS-based approaches for camera pose estimation.
To summarize, \ourmethod contributions are threefold:
\begin{itemize}
    \item Our approach for 6DoF camera pose estimation eliminates the need for an initial camera pose and iterations to converge, which is typically required in analysis-by-synthesis approaches;
    \item \ourmethod employs a novel ray casting pipeline, \ie Ellicell, and an attention-based mechanism that efficiently matches pixel-level image information with 3DGS ellipsoids;
    \item The proposed method is state-of-the-art in the NVS benchmarks for camera pose estimation both for accuracy and real-time performance.
\end{itemize}

\section{Related works}\label{sec:related_works}
We review relevant works on 6DoF camera pose estimation based on Neural Radiance Fields (NeRF) models, ellipsoid-based approaches, and correspondence matching methods that are related to key components of \ourmethod.

\noindent\textbf{Pose estimation from neural radiance fields.} iNeRF~\cite{yen2020inerf} pioneered NeRF-based 6D camera pose estimation, using iterative alignment of target and rendered images based on photometric error. 
However, iNeRF is prone to local minima in the optimization function, leading to recent developments like Parallel iNeRF~\cite{lin2023parallelinerf}, which employs parallel optimization of multiple candidate poses. 
While these approaches rely on NeRF-based models, NeMo+VoGe~\cite{wang2022voge, wang2020nemo} have explored 6D camera pose estimation using object models based on volumetric Gaussian reconstruction kernels as geometric primitives. 
The rendering strategy (VoGE) differs from 3DGS as it is based on ray marching. 
Therefore, NeMo+VoGe iteratively aligns learned features from target and rendered images. 
Notably, NeMo+VoGe's training requires multiple objects, in contrast to our method, which leverages a single object 3DGS model.
Alternatively, CROSSFIRE~\cite{moreau2023crossfire} addresses the local minima issue by integrating learned local features, which
describes not only the visual content but also the 3D location of the scene in the NeRF model.
Despite these advancements, analysis-by-synthesis approaches often struggle with inefficient pose updates due to the nature of the optimization refinement and the dependence on accurate initial pose priors.
These factors can limit their real-world applicability.
Recently, IFFNeRF~\cite{bortolon2024iffnerf} utilized a method that inverts the NeRF model to re-render an image to match a target one. 
However, unlike our approach, it does not consider the specificities of 3DGS, which include ellipsoid elongation and rotation, and their non-uniform distribution across the scene surface.

\noindent\textbf{Pose estimation from ellipsoids.} Recovery of the camera pose from ellipsoids has been explored for both SfM \cite{Gay_2017_ICCV,crocco2016structure,gay2019visual,gaudilliere2020perspective,shan2021ellipsdf,chen2021robust} and SLAM \cite{hosseinzadeh2019structure, laidlow2022simultaneous,liao2022so, gaudilliere2019camera, meng2022ellipsoid, zins2022oa} scenarios, where methods frequently recover the object's ellipsoid representation as well as the camera 6DoF. Such approaches typically solve linear systems to recover the pose, most commonly minimizing a loss of the projection to and from an object detection.
However, this methodological framework often presents limitations when confronted with large numbers of ellipsoids, as they are more indicated for handling few large ellipsoids that model a single object occupancy, 3D position and orientation.

\noindent\textbf{Correspondences Matching.} 
In traditional 6DoF image matching, feature-based approaches are used, which often rely on hand-crafted features, \eg, SIFT~\cite{lowe1999object} or more recent deep approaches such as SuperGlue~\cite{sarlin2020superglue} and TransforMatcher~\cite{kim2022transformatcher}.
SuperGlue utilizes a Graph Neural Network (GNN) for feature attention and Sinkhorn~\cite{sinkhorn} for matching, while LightGlue replaces the GNN with a lightweight transformer.
Unlike these, TransforMatcher ~\cite{kim2022transformatcher}  performs global match-to-match attention, allowing for accurate match localization.
In addition, there is a body of work around feature equivariance \cite{lee2023equiFeatures,lee2022self} for improving the robustness of matching.
However, these methods rely on the hypothesis that both feature sets exist in a homogeneous feature space, \ie extracted from the image, while in 6DGS we have the specific problem to match pixel to rays emitted from the Ellicells.
Therefore, we rely on the proposed attention model to handle these ray-to-pixel bindings.
OnePose++~\cite{he2022onepose++} instead adopts a multi-modal approach matching a point cloud with an image.
Another proposed alternative is to regress directly the pose parameters, as in CamNet~\cite{ding2019camnet}.
Nevertheless, these approaches require a large amount of training data ($\approx500$ or more images), sometimes across multiple scenes and, like with CamNet, these need to be available also at inference time.
\ourmethod however, requires only $\approx100$ or less images, which are utilized only once during training.

\section{Preliminaries}\label{sec:prelim}

We first review 3D Gaussian Splatting (3DGS)~\cite{kerbl20233Dgaussians} to understand the underlying principles and provide the mathematical formalization of the model.
3DGS objective is to synthesize novel views of a scene by optimizing the position, the orientation and the color of a set of 3D Gaussians approximated as ellipsoids $\mathcal{Q} = \{\mathbf{Q}\}_{i=1}^{K}$ from a given set of input images $\mathcal{I} = \{\mathbf{I}\}_{i=1}^{J}$ and their corresponding camera projection matrices $\mathcal{P} = \{\mathbf{P}\}_{i=1}^{J} \in \mathbb{R}^{3 \times 4}$.
A point $\mathbf{d}$ for being on the surface of an ellipsoid must satisfy the equation $(\mathbf{d} -\mathbf{x})\mathbf{\Sigma}(\mathbf{d} -\mathbf{x})^{T} = 1$, where $\mathbf{x} \in \mathbb{R}^{3}$ is the ellipsoid center and $\mathbf{\Sigma} \in \mathbb{R}^{3 \times 3}$ its covariance matrix.
We can further decompose the covariance of the ellipsoid $\mathbf{\Sigma}$ as:
\begin{equation}
\label{eqn:covariance_3D_equation}
    \mathbf{\Sigma} = \mathbf{R}\mathbf{U}\mathbf{U}^T\mathbf{R}^T,
\end{equation}
where $\mathbf{R} \in \mathbb{R}^{3 \times 3}$ is the ellipsoid rotation matrix and $\mathbf{U}^{3 \times 3}$ denotes the scaling matrix.
The projection matrix $\mathbf{P} \in \mathbb{R}^{3 \times 4}$ allows the projection of the ellipsoid $\mathbf{Q}$ onto the image plane generating the corresponding ellipse representation:
\begin{equation}
\label{eqn:splatting_equation}
    \breve{\mathbf{y}} = \mathbf{P}\breve{\mathbf{x}}^T,\ \breve{\mathbf{E}} = \mathbf{P}\mathbf{\Sigma}\mathbf{P}^T,
\end{equation}
where $\mathbf{y} \in \mathbb{R}^{2}$ and $\breve{\mathbf{y}} \in \mathbb{R}^{3}$ correspond to the Euclidean and homogeneous coordinates of the ellipse center point.
The homogeneous coordinates $\breve{\mathbf{y}}$ originate from the projection of the corresponding ellipsoid center in the homogeneous coordinates $\breve{\mathbf{x}} \in \mathbb{R}^{4}$.
The matrix $\breve{\mathbf{E}} \in \mathbb{R}^{3 \times 3}$ is the ellipse covariance in homogeneous space.
The covariance of the ellipse $\mathbf{E} \in \mathbb{R}^{2 \times 2}$, is derived by selecting only the first two rows and columns of $\breve{\mathbf{E}}$ and dividing by the last element on $\breve{\mathbf{E}}$ diagonal.

The splatted ellipses, denoted as $\mathcal{B} = \{\left< \mathbf{y}, \mathbf{E} \right>\}_{i=1}^{K}$, generate a pixel color with the  rendering function $\phi$ using rasterization techniques~\cite{akenine2018realtimeRendering, kerbl20233Dgaussians}.
The function $\phi$ acts independently on every single pixel of the image $\mathbf{p}$.
The pixel value depends on the neighboring projected ellipses, taking into account their center points' distances to the pixel coordinates, as well as their orientations and scales.
$\phi$ assumes that the ellipses are ordered based on the depth, so they should be sorted.
Formally, $\phi$ can be expressed as:
\begin{equation}
\label{eqn:psi_rendering_equation}
    \phi(\mathcal{B}, \mathbf{p}) = \sum^{K}_{i = 1} \mathbf{\rho}_i \alpha_i e^{-\tau(\mathcal{B}_i, \mathbf{p})} \gamma(i, \mathbf{p}),
\end{equation}
where $\mathbf{\rho}$ and $\alpha$ represent the color and opacity attributes associated with the ellipsoid, which are inherited by the splatted ellipse.
Similar to the volumetric rendering equation in NeRF, $\gamma$ denotes the inverse of the volume density accumulated up to the $i^{th}$ ellipse on pixel $\mathbf{p}$ and is defined as:
\begin{equation}
\label{eqn:inverse_volume_density}
\gamma(i, \mathbf{p}) = \prod_{j=1}^{i-1}(1 - \alpha_j e^{-\tau(\mathcal{B}_i, \mathbf{p})}).
\end{equation}
The purpose of $\tau$ is to determine the light absorption by the ellipse when represented as a 2D Gaussian. Light absorption depends on the orientation and distance between the ellipse center, denoted as $\mathbf{y}$, and the pixel location, expressed as $\mathbf{d} = \mathbf{p} - \mathbf{y}$.
Consequently, we can formally define $\tau$ as:
\begin{equation}
\label{eqn:tau_alpha_equation}
    \tau(\mathbf{B}, \mathbf{p}) = \frac{1}{2} \left ( \mathbf{1}_2 \mathbf{d}^T \mathbf{E} \mathbf{d} \mathbf{1}_2^T \right ),
\end{equation}
where $\mathbf{1}_2 \in \mathbb{R}^{2}$ denotes a vector filled with ones.
Following the processing of all pixels onto the image plane, the rendering function $\phi$ generates an image $\hat{\mathbf{I}} \in \mathbb{R}_{+}^{H \times W}$, where $W$ and $H$ represent the width and height of the image.

\section{Our approach}\label{sec:approach}
\begin{figure*}[t]
\begin{center}
\includegraphics[width=0.93\linewidth]{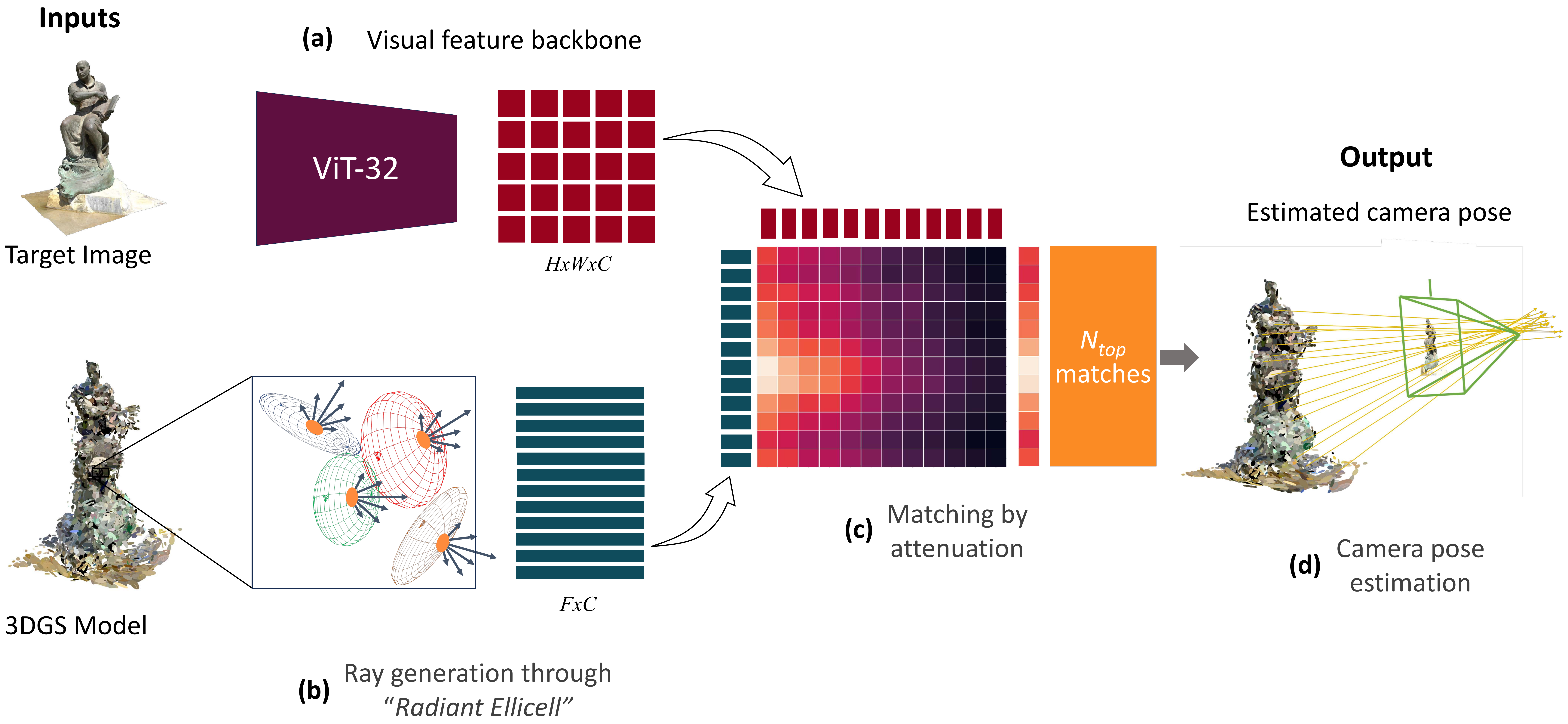}
\end{center}
\vspace{-6mm}
\caption{
The figure illustrates the pipeline of our \ourmethod methodology. 
The image is encoded using a visual backbone \textbf{$\mathbf{(a)}$}.
Concurrently, rays are uniformly projected from the center of the 3DGS ellipsoids \textbf{$\mathbf{(b)}$}, and their corresponding color is estimated.  
Subsequently, an attention map mechanism is employed to compare the encoded ray and image features \textbf{$\mathbf{(c)}$}. 
Following this comparison, the $N_{top}$ matches are selected via attenuation, and the camera location is estimated \textbf{$\mathbf{(d)}$} as the solution of a weighted Least Squares problem, resulting in a distinct 6DoF pose for the image.
}
\label{fig:architecture_diagram}
\vspace{-10pt}
\end{figure*}

\subsection{Overview}

\ourmethod estimates the camera pose $\hat{\mathbf{P}} \in \mathbb{R}^{3 \times 4}$, given a target image $\mathbf{I}_t$ and a set of ellipsoids $\mathcal{Q}$ from a pre-computed 3DGS model (Fig.~\ref{fig:architecture_diagram}).
To solve for the camera pose, we propose a casting method from the ellipsoid's surface, called \raygenerationmethod, that divides it in equal area cells (Sec.~\ref{sec:isocell_generation}).
The ellipsoids cast a set of $N$ rays, denoted as $\mathcal{V} = \{\left< \mathbf{v}_{o}, \mathbf{v}_{d}, \mathbf{v}_{c} \right>\}_{i=1}^{N}$, one for each of the generated cell (Fig.~\ref{fig:ellipcell_3d_rays}).
Each ray is identified by 
\textit{i)} the origin $\mathbf{v}_{o} \in \mathbb{R}^{3}$, 
\textit{ii)} the center point of each ellipsoid, 
\textit{iii)} the direction $\mathbf{v}_{d} \in \mathbb{R}^{3}$ originating from the ellipsoid center to the cell center and through the space, and 
\textit{iv)} the color information $\mathbf{v}_{c} \in \mathbb{R}^{3}$ as RGB values.
We synthesize the rays' color using the 3DGS rendering function $\phi$ (Eq.~\ref{eqn:psi_rendering_equation}).
A subset of these rays, depending on the view perspective, may intersect the camera's optical center.
For binding the rays to the image pixels we compute the target image pixels features $\psi(\textbf{I}_t)$ (Fig.~\ref{fig:architecture_diagram}$\mathbf{a}$) and the rays features $\psi(\mathcal{V})$ (Fig.~\ref{fig:architecture_diagram}$\mathbf{b}$).
These features are used to identify the intersecting rays by using an attention map $\mathcal{A}$ (Fig.~\ref{fig:architecture_diagram}$\mathbf{c}$), see Sec.~\ref{sec:attention_map}.
The higher the attention value for a ray-pixel pair is, the more likely the intersection on the image plane is a valid one.
Lastly, we determine $\hat{\textbf{P}_t}$ (Fig.~\ref{fig:architecture_diagram}$\mathbf{d}$) by computing the intersection point of rays using the weighted Least Squares algorithm (Sec.~\ref{sec:pose_recovery}).

\subsection{Radiant \raygenerationmethod}\label{sec:isocell_generation}

\begin{figure}[t]
    \begin{center}
        \subfloat[][{\scriptsize \raygenerationmethod components}]{\includegraphics[width=.30\linewidth]{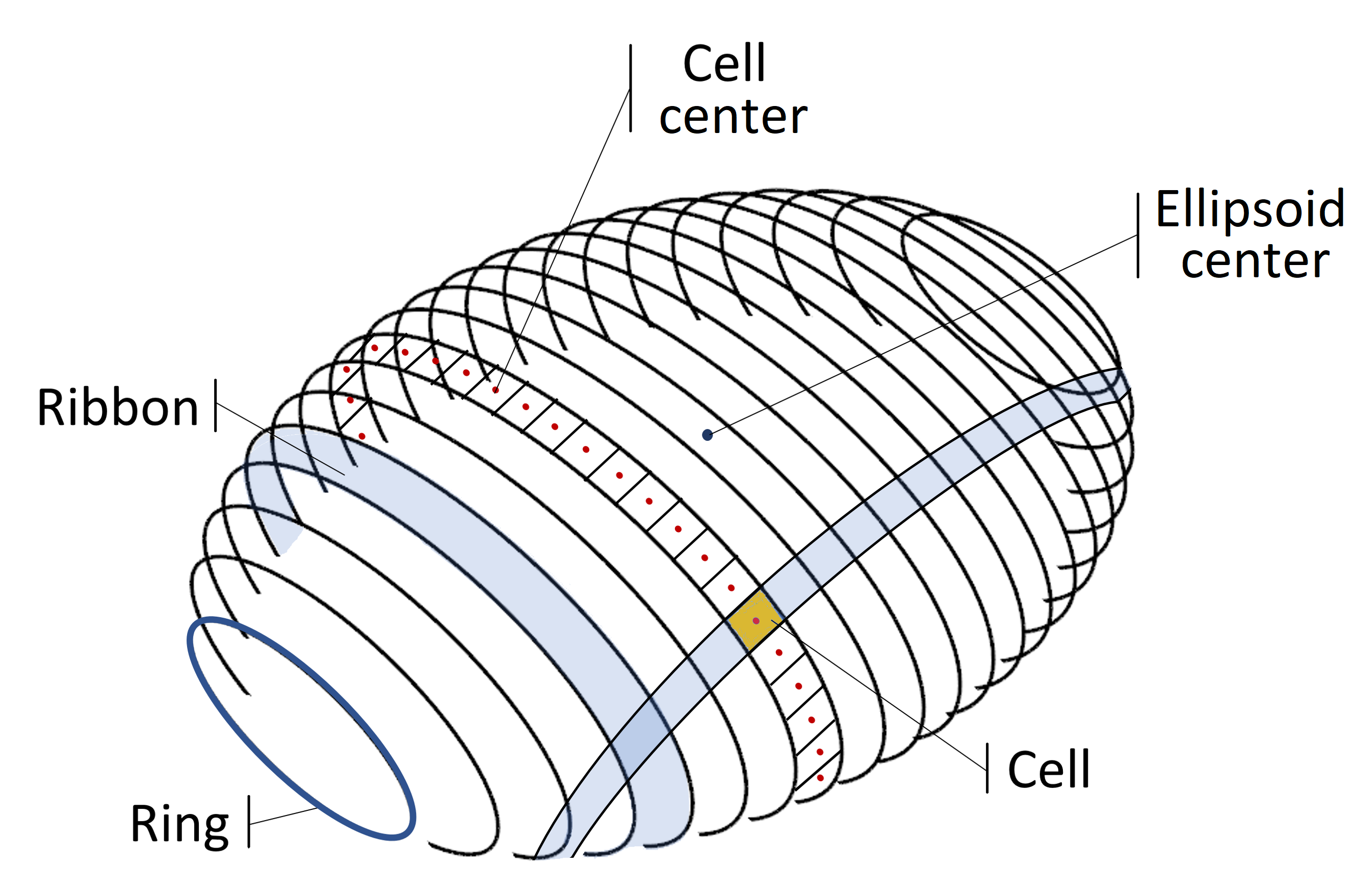}\label{fig:ellipcell_rings_and_ribbons}}
        \hspace{1.2em}
        \subfloat[][{\scriptsize 3D \raygenerationmethod grid}]{\includegraphics[width=.22\linewidth]{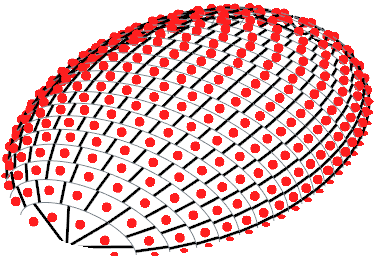}\label{fig:ellipsoidell_3d}}
        \hspace{1.8em}
        \subfloat[][{\scriptsize 3D radiant \raygenerationmethod}]{\includegraphics[width=0.23\linewidth]{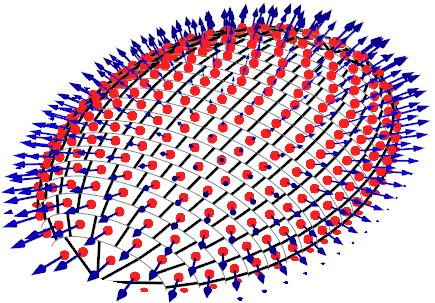}
        \label{fig:ellipcell_3d_rays}}
        \caption{
        The illustration depicts the three primary stages involved in the radiant \raygenerationmethod generation. 
        Firstly, \textbf{(a)} delineates the formulation of components required to compute the geometric information for each cell. Secondly, \textbf{(b)} shows the resulting \raygenerationmethod grid positioned on the surface of the ellipsoid along with their respective center points. 
        Finally, \textbf{(c)} demonstrates the generation of rays originating from the center point of the ellipsoid going through the \raygenerationmethod center.
        }
    \label{fig:ray_methods}
    \end{center}
    \vspace{-15pt}
\end{figure}

We create rays spanning in every direction allowing \ourmethod to recover the camera pose.  
We introduce the concept of radiant \raygenerationmethod for generating rays that uniformly emanate from the ellipsoid surface, as illustrated in Fig.~\ref{fig:ray_methods}.
\raygenerationmethod generation is deterministic \cite{masset2011, beckers2016} and achieves higher precision with fewer rays~\cite{jacques2015isocellEfficiency, Tsesmelis2018RGBD2luxDL} compared to other sampling methods like Monte-Carlo~\cite{malley1988}.

First, we compute the area of each \raygenerationmethod.
This is achieved by calculating the ellipsoid surface area, using a computationally efficient approach, namely Ramanujan approximation~\cite{almkvist1988ramanujan}:
\begin{equation}\label{eqn:ellipsoid_surface}
    h = 4 \pi \left ( \frac{(ab)^{1.6} + (ac)^{1.6} + (bc)^{1.6} }{3} \right )^{\frac{1}{1.6}},
\end{equation}
where $a, b, c = diag(\mathcal{S})$ are the ellipsoid axis scales.
Each \raygenerationmethod cell's target area equals $\mu = h / G$, with $G$ being the number of cells dividing each ellipsoid.
Approximating each cell as a square with side $z = \sqrt{\mu}$ we slice the ellipsoids along the major axis into ribbons, each as wide as $z$ (Fig. ~\ref{fig:ellipcell_rings_and_ribbons}).
The extremity of each ribbon is called a ring.
The total number of rings is $e = \left \lfloor \kappa(a, b) / { ( 2 z ) } \right \rfloor \in \mathbb{N}$, where $\kappa(a, b)$ computes the ring perimeter.
Ignoring ellipsoid's rotation, we compute the ring perimeter by treating them as 2D ellipses, thus defining $\kappa(a, b)$ as:
{\small
\begin{equation}\label{eqn:ramanujan_approximation}
    \kappa(a, b) = \pi \left ( ( a + b ) + \frac{3 ( a - b )^2 }{10 (a + b) + \sqrt{a^2 + 14 a b + b^2}} \right ).
\end{equation}
}

Given the total number of rings $e$ it is possible to compute the ribbon's centerline geometric parameters.
In particular, we compute the scale parameter of each ribbon as:
\begin{equation}\label{eqn:scaling_function}
    \varrho(n, \Delta r, a, b) = \sqrt{1 - \frac{(0.5 \Delta r + n \Delta r - a)^2}{b^2}},
\end{equation}
where $\Delta r = a / e $ is the distance between two consecutive rings.
This equation derives from the manipulation of the standard ellipse equation.
While ribbon size $z$ should be equal to $\Delta r$, these two values will likely differ due to the need for the number of rings being a natural number.
Eq.~\ref{eqn:scaling_function} is also used to compute the other ribbon scaling parameter by replacing $b$ with $c$.
$\varrho$ is then used to compute the number of cells inside each ribbon as:
\begin{equation}\label{eqn:element_per_ring}
\xi(n, e, a, b, c) = \left \lfloor \frac{\kappa \left ( \varrho(n, e, a, b), \varrho(n, e, a, c) \right )}{z} \right \rfloor,
\end{equation}
where $\xi$ is the number of cells inside the ring.
We compute the center of each cell, equally spaced along the ribbon's centerline, by sampling $\xi$ points along it.
This is challenging as the perimeter distance does not linearly correlate with the $x$ and $y$ variations.
However, we can solve this by using a statistical method.
Knowing a distribution's Cumulative Distribution Function (CDF) allows us to sample uniformly between 0 and 1 and then use the CDF inverse to map the sample to the distribution space.
This approach applies to our case, where samples are distributed as follows:
\begin{equation}\label{eqn:distance_on_ellipse}
ds^2 = dx^2 + dy^2,
\end{equation}
and, by taking its inverse, we can retrieve the coordinates of each cell center.
To simplify the equations, we define $r = \varrho(n, e, a, b)$ and $w = \varrho(n, e, a, c)$ to indicate the scale of the ellipse under consideration.
Then we express Eq.~\ref{eqn:distance_on_ellipse} in polar coordinates to simplify the differentiation:
\begin{equation}\label{eqn:cdf_function}
\frac{ds}{d\theta} = \sqrt{r^2 \sin^2 \theta + w^2 \cos^2 \theta},
\end{equation}
then, we can express the set of points on the perimeter of the ribbon centerline as an angular position in the polar coordinate system as:
\begin{equation}\label{eqn:theta_prime}
\theta^{\prime} = \left ( \frac{ds}{d\theta} \right )^{-1} \left ( g \cdot \frac{1}{\xi(n, e, a, b, c)} \right ),
\end{equation}
with $g$ being the cell identifier.
Given $\theta^{\prime}$ we can use it inside the ellipse equation in polar coordinates to obtain the 3D position of each cell center:
\begin{equation}\label{eqn:ellipse_points}
\mathbf{u} = \begin{pmatrix}
w \cos (\theta^{\prime})\\ 
g \sin(\theta^{\prime}) \\
- a + n \Delta r
\end{pmatrix}.
\end{equation}

\subsection{Ray generation}

Once we have divided each ellipsoid of the 3DGS model into equidistant cells, we cast the rays originating from the center point of the ellipsoid \ie $\mathbf{v}_{o} = \mathbf{x}$ and oriented towards the \raygenerationmethod center $\mathbf{v}_{d} = \mathbf{u} - \mathbf{x}$.
We reduce the number of potential rays cast from each ellipsoid by considering only the rays oriented in the same hemisphere as the estimated surface normal of the ellipsoid.
We obtain the surface normals by treating the ellipsoid centroids as a point cloud, and the surface normal is estimated using the nearby points~\cite{Tombari2010UniqueSO}.

Finally, each ray has also been associated with the color information $\mathbf{v}_{c}$, which we compute through the same pixel-level approach of 3DGS (Eq.~\ref{eqn:tau_alpha_equation}).
We note that the application of the volumetric rendering function of Eq.~\ref{eqn:tau_alpha_equation} produces a single pixel for each ray.
The generated rays represent a collection of potential hypotheses, meaning that a subset of them will intersect the target image $\mathbf{I}_t$.

\subsection{Binding by attenuation of rays to image}\label{sec:attention_map}
Given all the cast rays $\mathbf{v}$, we identify a subset of $\mathbf{v}$ correlating with the target image $\textbf{I}_t$.
A learned attention map $\mathcal{A}$ assigns scores $\hat{\mathbf{s}}$ based on the highest correlation to image pixels; higher similarity results in higher scores.
Based on scores $\hat{\mathbf{s}}$, we select the top candidate's rays ($N_{top}$) that present maximal association and use them to recover the pose ($\hat{\mathbf{P}}_t$).

To select rays with similar appearance and position, we use a Multi-Layer Perceptron (MLP) defined as $\textbf{V} = \psi(\mathbf{v})$, where $\textbf{V} \in \mathbb{R}^{N \times C}$ with $C$ being the feature size and $N$ the overall number of rays.
The MLP input is enriched by incorporating Positional Encoding that maps the data in the Fourier domain~\cite{tancik2020fourfeat} to better distinguish between similar data.

We generate features from $\mathbf{I}_t$ using DINOv2~\cite{oquab2023dinov2} as a pre-trained backbone feature extractor.
This results in a set of features $\mathbf{F}_{t} \in \mathbb{R}^{M \times C}$, where $M = W \times H$. 
Both the image and ray features sets are processed by a single attention module $\mathcal{A}(\mathbf{V}_{f}, \mathbf{F}_{t}) \in \mathbb{R}^{M \times N}$ producing a set of scores.
Inside the attention module the ray features, $\mathbf{V}$, are used as queries and the image features, $\mathbf{F}_{t}$, as a key.  
We optimize the attention map by summing along the rows and converting it into a per-ray correlation score 
as follows $\hat{\mathbf{s}}= \sum_{i=1}^{M} \mathcal{A}_i.$
The higher the score value given by $\hat{\mathbf{s}}$, the better the association between the rays and image pixels.
At test-time we select the $N_{top}$ rays with the highest ranking scores.

Because a ray and an image pixel should be associated with each other based on the distance between the camera origin and its projection onto the corresponding ray, we supervise the predicted scores $\hat{\mathbf{s}}$ using the same images used to estimate the 3DGS model at training time.
We compute the projection of the point on the line as $l = max((\mathbf{O} - \mathbf{v}_{o}) \mathbf{v}_{d}, 0)$,
where $\mathbf{O}$ is the camera position, $\mathbf{v}_{o}$ the generated ray origin and $\mathbf{v}_{d}$ the corresponding direction.
Rays are infinite only in one direction, so we restrict $l \in \mathbb{R}^{+}$ using the max operator.
Then, we can compute the distance between the camera origin and its projection on the ray as $\mathbf{h} = \|(\mathbf{v}_{o} + l  \mathbf{v}_{d}) - \textbf{O}\|_2$.
The value $\mathbf{h}$ can span from $0$ to $+\infty$, with $0$ indicating a ray that passes through the camera's optical center.
We map distances to the attention map score using:
\begin{equation}
\label{eqn:score_regularization}
     \mathbf{\delta} = 1 - tanh\left(\frac{\mathbf{h}}{\lambda}\right),\ 
    \mathbf{s} = \mathbf{\delta} \frac{M}{\sum \mathbf{\delta}},
\end{equation}
where $\lambda$ regulates the number of rays to assign to a specific camera.
Lastly, the softmax inside the attention map computation requires we normalize the ground truth scores.
We use the $L2$ loss to minimize the difference between the predicted $\hat{\mathbf{s}}$ and the computed ground truth $\mathbf{s}$ scores as:
\begin{equation}\label{eqn:optimizer_loss}
\mathcal{L} = \frac{1}{MN} \sum_{i = 1}^{M}\sum_{j = 1}^{N} \| \hat{\mathbf{s}}_{i,j} - \mathbf{s}_{i,j} \|_2,
\end{equation}
where $M,N$ are the size of the attention map $\mathcal{A}$.
During each training iteration, we predict an image and a pose utilized for estimating the 3DGS model.

\subsection{Test-time pose estimation}
\label{sec:pose_recovery}

During the test phase, the predicted scores $\hat{\textbf{s}}$ are used to select the top $N_{top}$ rays, identified as the utmost relevant, and constrained to choose at most one ray per ellipsoid.
Note that only a small set of rays is sufficient to estimate the camera pose. 
However, based on an ablation study we set $N_{top}=100$, see Tab.~\ref{tab:top_k}.

The camera position is found at the intersection of selected rays, solved as a weighted Least Squares problem. Since 3D lines usually do not intersect at a single point due to discretization noise introduced by the \raygenerationmethod, we minimize the sum of squared perpendicular distances instead.

For the selected ray $\mathbf{v}_j$ with $f = 1 \ldots N_{top}$, the error is given by the square of the distance from the camera position to predict $\hat{\mathbf{O}}$ to its projection on $\mathbf{v}_j$:
\begin{equation}\label{eqn:minimization_of_square_diff}
    \sum_{f=1}^{N_{top}}\left( (\hat{\mathbf{O}}-\mathbf{v}_{o,f})^T (\hat{\mathbf{O}}-\mathbf{v}_{o,f}) - ((\hat{\mathbf{O}}-\mathbf{v}_{o,f})^T\mathbf{v}_{d,f})^2 \right),
\end{equation}
where $\mathbf{v}_{o,f}$ indicating the origin of the $f$-th ray and $\mathbf{v}_{d,f}$ the respective direction.
To minimize Eq.~\ref{eqn:minimization_of_square_diff}, we differentiate it with respect to $\hat{\mathbf{O}}$, resulting in
\begin{equation}\label{eqn:minimization_of_square_diff_differentiate}
    \hat{\mathbf{O}} = \sum_{f=1}^{N_{top}}\hat{\mathbf{s}}_{f}(\mathbb{I} - \mathbf{v}_{d,f}\mathbf{v}_{d,f}^T)\mathbf{v}_{o,f},
\end{equation}
where $\mathbb{I}$ is the identity matrix and $\hat{\mathbf{s}}_{f}$ are the predicted ray scores.
This expression can be solved as a weighted system of linear equations.

\section{Results}\label{sec:results}

\begin{table*}[t]
\caption{
Evaluation of 6DoF pose estimation on the Mip-NeRF 360\textdegree~\cite{barron2022mipnerf360} dataset.
We report results in terms of Mean Angular Error (MAE) and Mean Translation Error (MTE) in terms of degrees and units, $u$, respectively.  Where $1u$ is equal to the object's largest dimension.
For both metrics lower is better.
Best-performing results are highlighted in \textbf{bold} and green, while second best values are highlighted in orange.
}
\centering
\begin{adjustbox}{max width=\textwidth}
\setlength\tabcolsep{3.8pt}
\begin{tabular}{l|ll|ll|ll||ll|ll|ll|ll}

& \multicolumn{6}{c||}{Fixed pose prior (eval. protocol by \cite{yen2020inerf})} & \multicolumn{6}{c}{Random pose prior} & \multicolumn{2}{c}{No pose prior} \\ \cmidrule(lr){2-7} \cmidrule(lr){8-13} \cmidrule(lr){14-15}
\multicolumn{1}{c|}{}         & \multicolumn{2}{c}{\makecell[cb]{iNeRF~\cite{yen2020inerf}}} & \multicolumn{2}{c}{\makecell[cb]{NeMo + VoGE~\cite{wang2022voge}}} & \multicolumn{2}{c||}{\makecell[cb]{Parallel iNeRF~\cite{lin2023parallelinerf}}} &  \multicolumn{2}{c}{\makecell[cb]{iNeRF~\cite{yen2020inerf}}}  & \multicolumn{2}{c}{\makecell[cb]{NeMo + VoGE~\cite{wang2022voge}}} & \multicolumn{2}{c}{\makecell[cb]{Parallel iNeRF~\cite{lin2023parallelinerf}}} &         \multicolumn{2}{c}{\makecell[cb]{\ourmethod (Ours)}}                 \\
\cmidrule(lr){2-3} \cmidrule(lr){4-5} \cmidrule(lr){6-7} \cmidrule(lr){8-9} \cmidrule(lr){10-11} \cmidrule(lr){12-13} \cmidrule(lr){14-15}
\makecell[cb]{Scenes}       & MAE $\downarrow$      & MTE $\downarrow$     &  MAE  $\downarrow$    & MTE $\downarrow$       & MAE $\downarrow$     & MTE $\downarrow$       & MAE $\downarrow$ & MTE$\downarrow$ & MAE $\downarrow$  & MTE $\downarrow$  & MAE $\downarrow$      & MTE $\downarrow$     & MAE $\downarrow$ & MTE $\downarrow$ \\  \hline                   
           Bicycle     & \cellcolor{red!5}39.5            & \cellcolor{red!5}0.116       & \cellcolor{red!5}43.8              & \cellcolor{orange!15}0.015    & \cellcolor{orange!15}35.9  & \cellcolor{red!5}0.116         & \cellcolor{red!5}76.6       & \cellcolor{red!5}0.217   & \cellcolor{red!5}111.8       & \cellcolor{red!5}0.038    & \cellcolor{red!5}44.4       & \cellcolor{red!5}0.150 & \cellcolor{green!30}\textbf{12.1}       & \cellcolor{green!30}\textbf{0.010}   \\
                                   Bonsai      & \cellcolor{red!5}51.3            & \cellcolor{red!5}0.228       & \cellcolor{red!5}52.5              & \cellcolor{green!30}\textbf{0.036}    & \cellcolor{orange!15}41.1  & \cellcolor{red!5}0.223         & \cellcolor{red!5}96.7       & \cellcolor{red!5}0.385   & \cellcolor{red!5}98.9        & \cellcolor{red!5}0.073    & \cellcolor{red!5}58.2       & \cellcolor{red!5}0.298 & \cellcolor{green!30}\textbf{10.5}      & \cellcolor{orange!15}0.038   \\
                                   Counter     & \cellcolor{red!5}40.7            & \cellcolor{red!5}0.324       & \cellcolor{red!5}45.6              & \cellcolor{orange!15}0.072    & \cellcolor{orange!15}24.7  & \cellcolor{red!5}0.212         & \cellcolor{red!5}70.3       & \cellcolor{red!5}0.487   & \cellcolor{red!5}98.1        & \cellcolor{red!5}0.139    & \cellcolor{red!5}42.1       & \cellcolor{red!5}0.435 & \cellcolor{green!30}\textbf{19.6}      & \cellcolor{green!30}\textbf{0.043}   \\
                                   Garden      & \cellcolor{orange!15}31.0            & \cellcolor{red!5}0.121       & \cellcolor{red!5}31.8              & \cellcolor{orange!15}0.026    & \cellcolor{green!30}\textbf{18.2}  & \cellcolor{red!5}0.090         & \cellcolor{red!5}72.8       & \cellcolor{red!5}0.210   & \cellcolor{red!5}89.2        & \cellcolor{red!5}0.038    & \cellcolor{red!5}60.0       & \cellcolor{red!5}0.144 & \cellcolor{red!5}37.8      & \cellcolor{green!30}\textbf{0.015}   \\
                                   Kitchen     & \cellcolor{red!5}38.2            & \cellcolor{red!5}0.113       & \cellcolor{red!5}41.6              & \cellcolor{orange!15}0.042    & \cellcolor{orange!15}37.3  & \cellcolor{red!5}0.109         & \cellcolor{red!5}100.2      & \cellcolor{red!5}0.266   & \cellcolor{red!5}122.2       & \cellcolor{red!5}0.082    & \cellcolor{red!5}65.0       & \cellcolor{red!5}0.193 & \cellcolor{green!30}\textbf{23.2}      & \cellcolor{green!30}\textbf{0.018}   \\
                                   Room        & \cellcolor{red!5}38.8            & \cellcolor{red!5}0.274       & \cellcolor{red!5}44.9              & \cellcolor{orange!15}0.045    & \cellcolor{green!30}\textbf{30.7}  & \cellcolor{red!5}0.257         & \cellcolor{red!5}91.6       & \cellcolor{red!5}0.444   & \cellcolor{red!5}110.0       & \cellcolor{red!5}0.010    & \cellcolor{red!5}63.5       & \cellcolor{red!5}0.271 & \cellcolor{orange!15}38.3      & \cellcolor{green!30}\textbf{0.019}   \\
                                   Stump       & \cellcolor{orange!15}21.4            & \cellcolor{red!5}0.030       & \cellcolor{red!5}26.3              & \cellcolor{orange!15}0.016    & \cellcolor{green!30}\textbf{14.8}  & \cellcolor{orange!15}0.016         & \cellcolor{red!5}86.9       & \cellcolor{red!5}0.035   & \cellcolor{red!5}96.3        & \cellcolor{red!5}0.025    & \cellcolor{red!5}72.6       & \cellcolor{red!5}0.033 & \cellcolor{red!5}28.3      & \cellcolor{green!30}\textbf{0.009}   \\
                                   \cellcolor[gray]{0.9} Avg.        & \cellcolor[gray]{0.9}\cellcolor{red!15} 37.3            & \cellcolor[gray]{0.9}\cellcolor{red!15} 0.172       & \cellcolor[gray]{0.9}\cellcolor{red!15} 40.9              & \cellcolor[gray]{0.9}\cellcolor{orange!25} 0.036    & \cellcolor[gray]{0.9}\cellcolor{orange!25} 28.9   & \cellcolor[gray]{0.9}\cellcolor{red!15} 0.146         & \cellcolor[gray]{0.9}\cellcolor{red!15} 85.0       & \cellcolor[gray]{0.9}\cellcolor{red!15} 0.292   & \cellcolor[gray]{0.9}\cellcolor{red!15} 103.8       & \cellcolor[gray]{0.9}\cellcolor{red!15} 0.058    & \cellcolor[gray]{0.9}\cellcolor{red!15} 58.0       & \cellcolor[gray]{0.9}\cellcolor{red!15} 0.218 & \cellcolor[gray]{0.9}\cellcolor{green!50} \textbf{24.3}       & \cellcolor[gray]{0.9}\cellcolor{green!50} \textbf{0.022}     
\end{tabular}
\end{adjustbox}
\vspace{-5pt}
\label{tab:summary_results_mip}
\end{table*}

\subsection{Experimental setup}

We evaluate \ourmethod and compare with other analysis-by-synthesis methods for 6D pose estimation, including iNeRF~\cite{yen2020inerf}, Parallel iNeRF~\cite{lin2023parallelinerf}, and NeMo+VoGE~\cite{wang2022voge, wang2020nemo}.
We reproduce the results using their published code.
We follow iNeRF's evaluation protocol and test on two real-world datasets: Tanks \& Temples~\cite{Knapitsch2017TanksAndTemples} and Mip-NeRF 360\textdegree~\cite{barron2022mipnerf360}.
For each dataset, we use the predefined training-test splits and evaluate them with two pose initialization pipelines: 
\textit{i)} the original iNeRF initialization, where the starting pose is sampled uniformly between $[-40^{\circ}, +40^{\circ}]$ degrees of errors and $[-0.1, +0.1]$ units of translation error from the ground-truth target pose; 
\textit{ii)} by randomly choosing an initialization pose from the ones used to create the 3DGS mode.
Although analysis-by-synthesis methods were tested with a prior, in reality it is rarely available, so we present a second scenario to assess them under more realistic conditions.
We perform multiple ablation studies to assess the sensitivity of \ourmethod to different hyperparameters and settings.
We quantify pose estimation results in terms of mean angular (MAE) and translation (MTE) errors (see Tab.~\ref{tab:summary_results_mip} and Tab.~\ref{tab:summary_results}) and measure the inference time.

\noindent\textbf{Implementation Details.}~\ourmethod is implemented in PyTorch and the attention map was trained for 1.5K iterations ($\sim$45mins) with an NVIDIA GeForce RTX 3090. 
We use the Adafactor optimizer~\cite{shazeer2018adafactor} with weight decay of $10^{-3}$. For speedup training, we uniformly sample 2000 3DGS ellipsoids at each iteration.

\subsection{Datasets}
To demonstrate the applicability of \ourmethod, we test on two datasets featuring real world challenges. 
\textbf{Tanks\&Temples}~\cite{Knapitsch2017TanksAndTemples} was created to evaluate 3D reconstruction methods with challenging real-world objects of varying sizes, acquired from human-like viewpoints and with difficult conditions (illumination, shadows, and reflections).
We use the five scenes (Barn, Caterpillar, Family, Ignatius, Truck) and the train test splits given in~\cite{liu2020nsvf, chen2022tensorf}.
The splits are object dependent, having on average $\approx247$ training images ($87\%$) and $\approx35$ testing images ($12\%$).
\textbf{Mip-NeRF 360\textdegree}~\cite{barron2022mipnerf360} consists of seven scenes: two outdoors and four indoors, with a structured scenario and background.
We use the original train-test splits~\cite{barron2022mipnerf360}, at a ratio of $1{:}8$.
Following~\cite{lin2023parallelinerf}, we resize all the objects to fit inside a unit box. 
The translation error is relative to the object size, defined as a unit.

\begin{table*}[t]
\caption{Evaluation of 6DoF pose estimation on the Tanks\&Temples~\cite{Knapitsch2017TanksAndTemples} dataset. We show the same metrics and analysis as in Table~\ref{tab:summary_results_mip}. 
}
\centering
\begin{adjustbox}{max width=\textwidth}
\setlength\tabcolsep{3.8pt}
\begin{tabular}{l|ll|ll|ll||ll|ll|ll|ll}

 & \multicolumn{6}{c||}{Fixed pose prior (eval. protocol by \cite{yen2020inerf})} & \multicolumn{6}{c}{Random pose prior} & \multicolumn{2}{c}{No pose prior} \\ \cmidrule(lr){2-7} \cmidrule(lr){8-13} \cmidrule(lr){14-15}
\multicolumn{1}{c|}{}               & \multicolumn{2}{c}{\makecell[cb]{iNeRF~\cite{yen2020inerf}}} & \multicolumn{2}{c}{\makecell[cb]{NeMo + VoGE~\cite{wang2022voge}}} & \multicolumn{2}{c||}{\makecell[cb]{Parallel iNeRF~\cite{lin2023parallelinerf}}} &  \multicolumn{2}{c}{\makecell[cb]{iNeRF~\cite{yen2020inerf}}}  & \multicolumn{2}{c}{\makecell[cb]{NeMo + VoGE~\cite{wang2022voge}}} & \multicolumn{2}{c}{\makecell[cb]{Parallel iNeRF~\cite{lin2023parallelinerf}}} &         \multicolumn{2}{c}{\makecell[cb]{\ourmethod (Ours)}}                 \\
\cmidrule(lr){2-3} \cmidrule(lr){4-5} \cmidrule(lr){6-7} \cmidrule(lr){8-9} \cmidrule(lr){10-11} \cmidrule(lr){12-13} \cmidrule(lr){14-15}
\makecell[cb]{Objects}       & MAE $\downarrow$      & MTE $\downarrow$     &  MAE  $\downarrow$    & MTE $\downarrow$       & MAE $\downarrow$     & MTE $\downarrow$       & MAE $\downarrow$ & MTE$\downarrow$ & MAE $\downarrow$  & MTE $\downarrow$  & MAE $\downarrow$      & MTE $\downarrow$     & MAE $\downarrow$ & MTE $\downarrow$ \\  \hline                   
Barn        & \cellcolor{orange!15}26.5            & \cellcolor{red!5}0.208       & \cellcolor{red!5}51.2              & \cellcolor{red!5}0.752    & \cellcolor{green!30}\textbf{22.9}  & \cellcolor{green!30}\textbf{0.131}      & \cellcolor{red!5}89.2       & \cellcolor{red!5}0.682   & \cellcolor{red!5}92.5        & \cellcolor{red!5}0.684    & \cellcolor{red!5}85.2      & \cellcolor{red!5}0.572 & \cellcolor{red!5}30.3       & \cellcolor{orange!15}0.162   \\
                                   Caterpillar & \cellcolor{red!5}42.9            & \cellcolor{red!5}0.166       & \cellcolor{red!5}52.6              & \cellcolor{red!5}0.516    & \cellcolor{orange!15}25.2  & \cellcolor{orange!15}0.138      & \cellcolor{red!5}89.3       & \cellcolor{red!5}2.559   & \cellcolor{red!5}90.5        & \cellcolor{red!5}2.559    & \cellcolor{red!5}86.8     & \cellcolor{red!5}0.843 & \cellcolor{green!30}\textbf{14.5}       & \cellcolor{green!30}\textbf{0.027}   \\
                                   Family      & \cellcolor{red!5}42.8            & \cellcolor{red!5}0.794       & \cellcolor{red!5}58.4              & \cellcolor{red!5}1.130    & \cellcolor{orange!15}22.9  & \cellcolor{orange!15}0.507      & \cellcolor{red!5}93.9       & \cellcolor{red!5}1.505   & \cellcolor{red!5}97.0        & \cellcolor{red!5}1.506    & \cellcolor{red!5}99.0     & \cellcolor{red!5}2.028 & \cellcolor{green!30}\textbf{20.6}       & \cellcolor{green!30}\textbf{0.468}   \\
                                   Ignatius    & \cellcolor{red!5}31.4            & \cellcolor{red!5}0.723       & \cellcolor{red!5}51.2              & \cellcolor{red!5}1.193    & \cellcolor{orange!15}23.4  & \cellcolor{orange!15}0.604      & 8\cellcolor{red!5}4.1       & \cellcolor{red!5}1.489   & \cellcolor{red!5}85.4        & \cellcolor{red!5}1.491    & \cellcolor{red!5}86.9     & \cellcolor{red!5}1.326 & \cellcolor{green!30}\textbf{15.5}       & \cellcolor{green!30}\textbf{0.441}   \\
                                   Truck       & \cellcolor{red!5}31.6            & \cellcolor{red!5}0.370       & \cellcolor{red!5}54.6              & \cellcolor{red!5}1.236    & \cellcolor{orange!15}29.4  & \cellcolor{orange!15}0.351      & \cellcolor{red!5}94.4       & \cellcolor{red!5}1.042   & \cellcolor{red!5}97.7        & \cellcolor{red!5}1.045    & \cellcolor{red!5}97.6     & \cellcolor{red!5}0.883 & \cellcolor{green!30}\textbf{27.5}       & \cellcolor{green!30}\textbf{0.242}   \\
                                   \cellcolor[gray]{0.9} Avg.       & \cellcolor[gray]{0.9}\cellcolor{red!15} 35.0            & \cellcolor[gray]{0.9}\cellcolor{red!15} 0.452       & \cellcolor[gray]{0.9}\cellcolor{red!15} 53.6              & \cellcolor[gray]{0.9}\cellcolor{red!15} 0.965    & \cellcolor[gray]{0.9}\cellcolor{orange!25} 24.7  & \cellcolor[gray]{0.9}\cellcolor{orange!25} 0.346         & \cellcolor[gray]{0.9}\cellcolor{red!15} 90.2       & \cellcolor[gray]{0.9}\cellcolor{red!15} 1.455   & \cellcolor[gray]{0.9}\cellcolor{red!15} 92.6        & \cellcolor[gray]{0.9}\cellcolor{red!15} 1.457    & \cellcolor[gray]{0.9}\cellcolor{red!15} 91.1       & \cellcolor[gray]{0.9}\cellcolor{red!15} 1.130 & \cellcolor[gray]{0.9} \cellcolor{green!50}\textbf{21.7}       & \cellcolor[gray]{0.9} \cellcolor{green!50}\textbf{0.268}   \\ \hline
\end{tabular}
\end{adjustbox}
\vspace{-5pt}
\label{tab:summary_results}
\end{table*}

\subsection{Analysis}
\noindent\textbf{Quantitative Analysis:}
Tab.~\ref{tab:summary_results_mip} and Tab.~\ref{tab:summary_results} present the results obtained across both datasets. 
\ourmethod consistently outperforms baseline methods across all datasets and pose initialization pipelines. 
Notably, \ourmethod achieves lower error rates than the second-best results, especially under identical comparison conditions (\ie, random pose prior). 
Even when initialized from a fixed pose proximal to the known camera, \ourmethod still excels over baselines in most scenes.
Parallel iNeRF demonstrates improvement over iNeRF across all tested scenarios, consistent with its reported enhancements, but both methods' performance drops with random initialization.
Likewise, NeMo+VoGE performs worst, especially with random pose prior due to the utilization of a smaller number of larger ellipsoids in their approach. 
In contrast, \ourmethod leverages approximately 300,000 ellipsoids of varying sizes obtained via 3DGS, as opposed to their mesh-to-ellipsoid method, which utilizes only about 5,000 larger ellipsoids. 
This fundamental disparity in ellipsoid size and quantity is a crucial factor contributing to the performance difference. 
Additionally, \ourmethod exhibits faster processing speeds, operating nearly in real-time at $15\: frames\: per\: second$ ($fps$) compared to the $0.05 fps$ of Parallel iNeRF and $0.16 fps$ of iNeRF.
Please refer to the supplementary material for the complete table on timings.

\begin{figure*}[t]
\centering
{\fontfamily{phv}\selectfont 
\begin{minipage}[outer sep=0]{0.48\textwidth}
    \centering
    {\scriptsize Truck}
    \vspace{.5ex}
    \hrule
    \vspace{.5ex}
    \begin{minipage}[m]{\textwidth}
        \centering
        \includegraphics[width=0.8\textwidth]{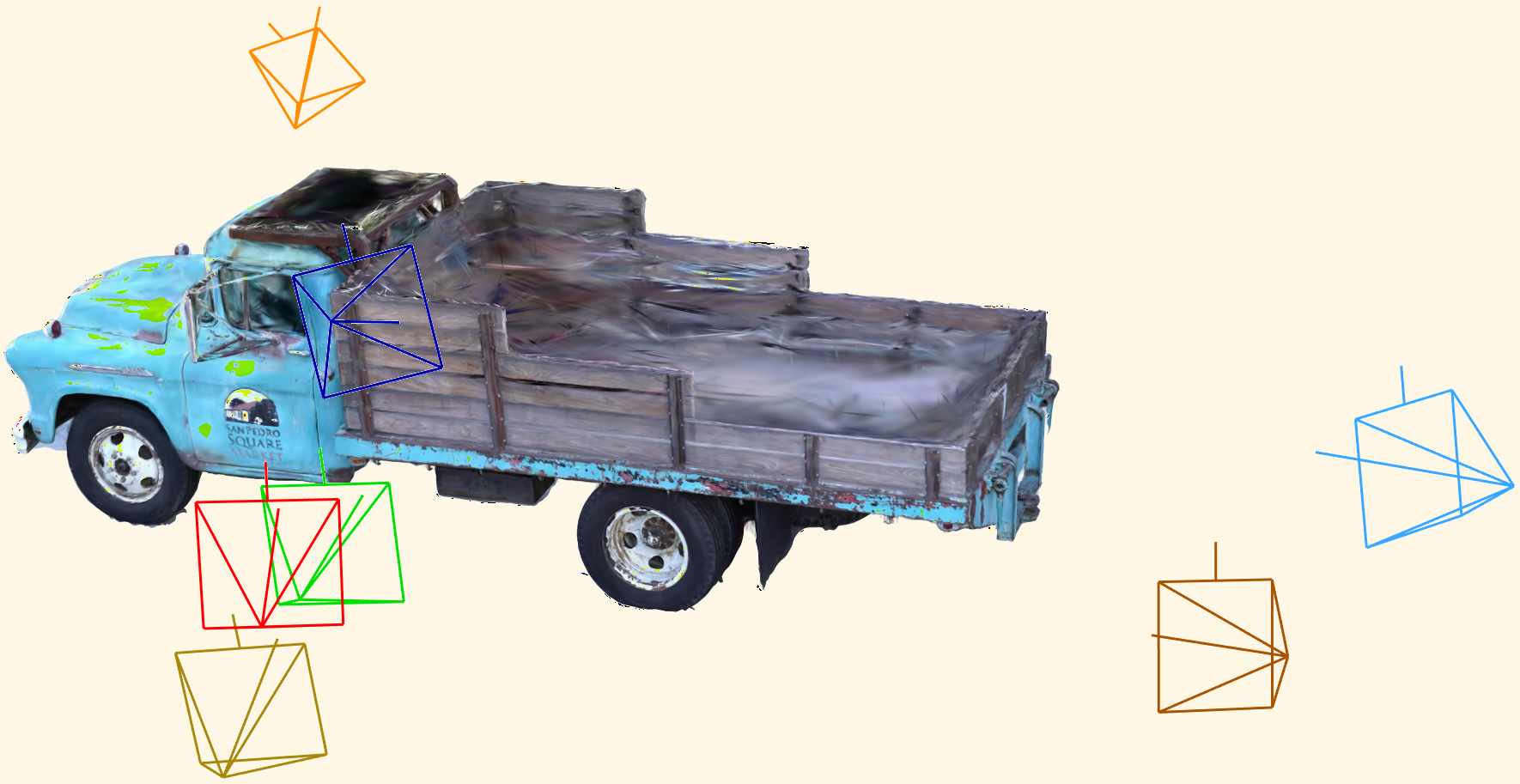} \\
    \end{minipage}
    \begin{minipage}[m]{\textwidth}
        \hspace{+7pt}
        \begin{minipage}[m]{0.49\textwidth}
        \centering
            {\tiny Target image} \\
            \includegraphics[width =0.7\textwidth]{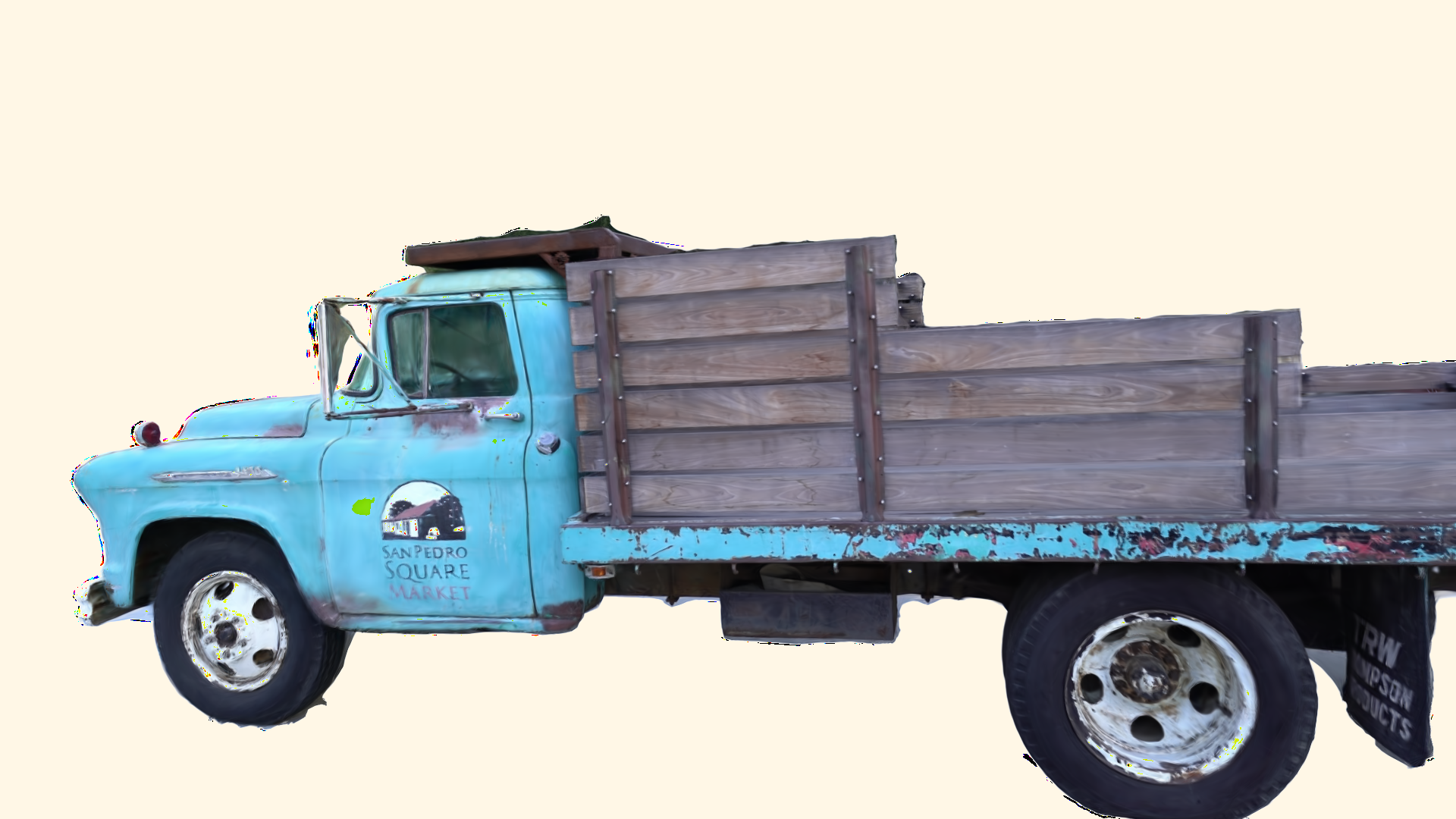}
        \end{minipage}
        \hspace{-25pt}
        \begin{minipage}[m]{0.49\textwidth}
        \centering
        {\tiny Estimated NVS} \\
        \includegraphics[width =0.7\textwidth]{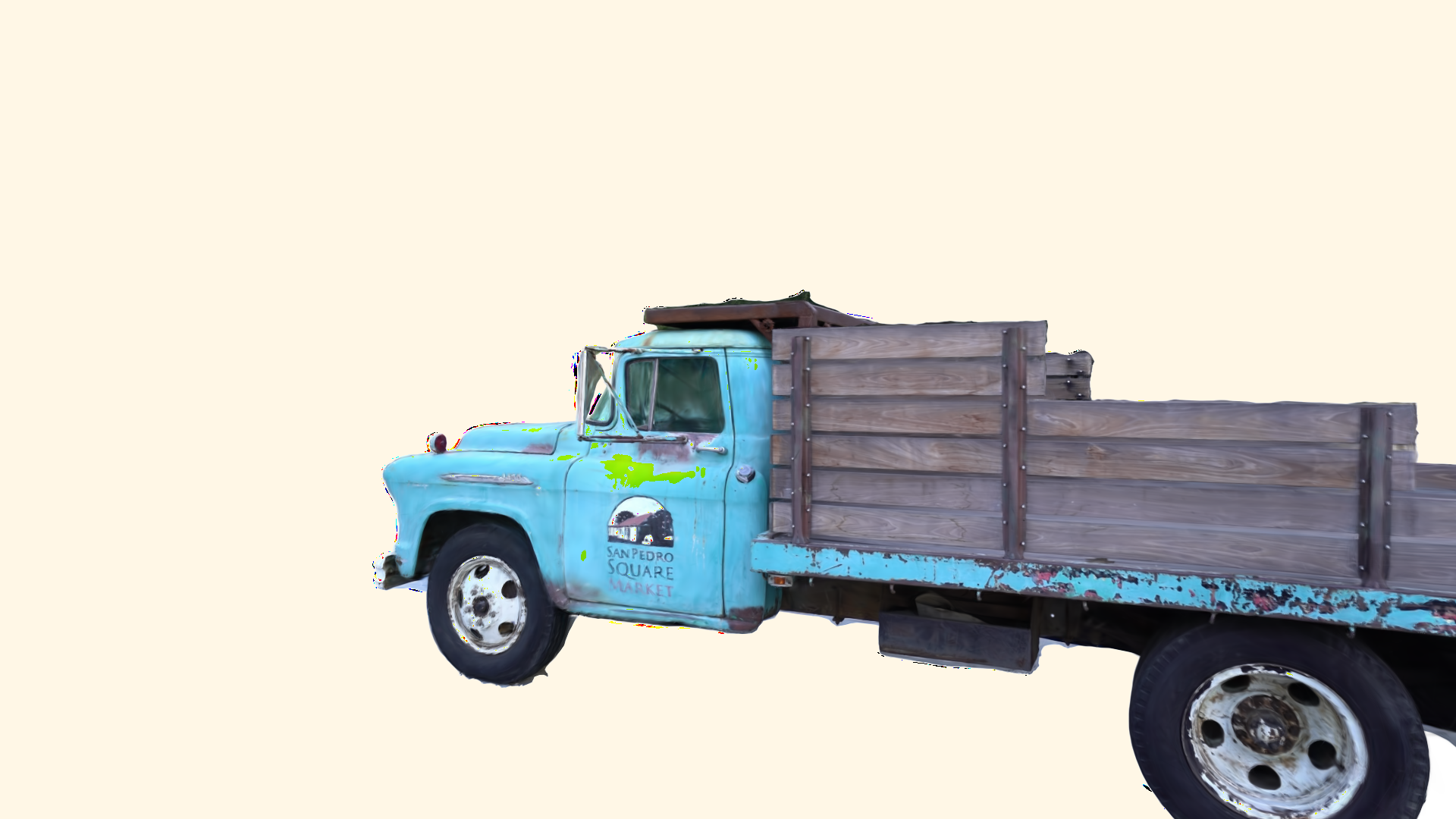}
        \end{minipage}\\[0.5ex]
    \end{minipage}\hfill
    \begin{minipage}[m]{0.6\textwidth}
    \centering
    \end{minipage}\hfill
    
\end{minipage}\hfill
\begin{minipage}[outer sep=0]{0.48\textwidth}
    \centering
    {\scriptsize Family}
    \vspace{.5ex}
    \hrule
    \vspace{.5ex}
    \begin{minipage}[m]{\textwidth}
        \centering
        
        \includegraphics[width=0.8\textwidth]{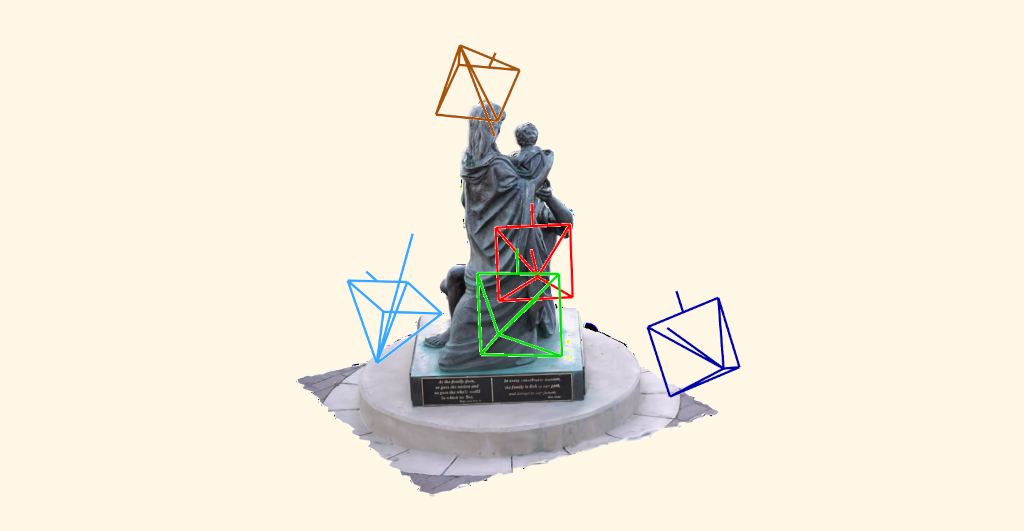} \\
    \end{minipage}
    \begin{minipage}[m]{\textwidth}
        \hspace{+7pt}
        \begin{minipage}[m]{0.49\textwidth}
        \centering
            {\tiny Target image} \\
            \includegraphics[width =0.7\textwidth]{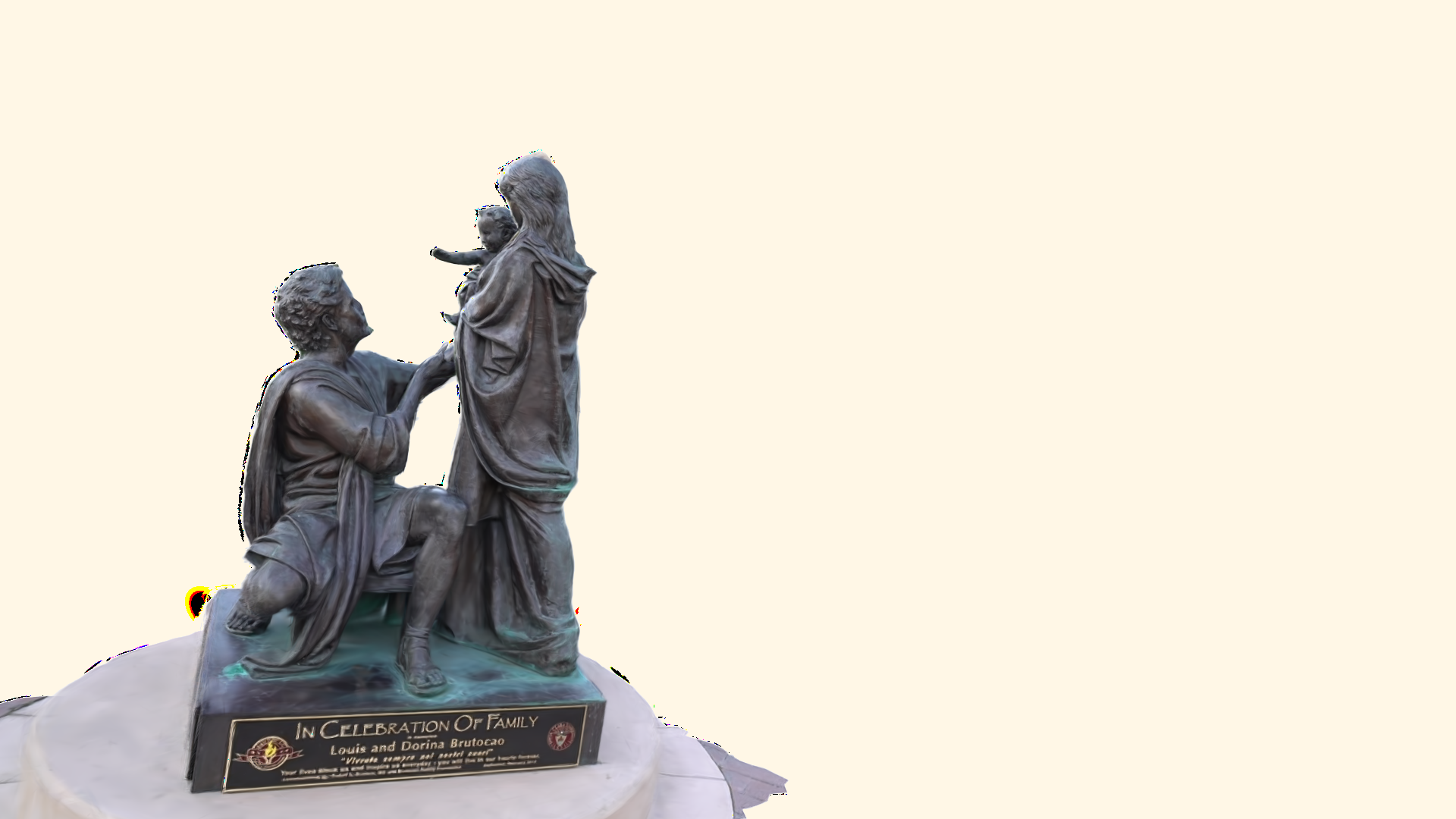}
        \end{minipage}
        \hspace{-25pt}
        \begin{minipage}[m]{0.49\textwidth}
        \centering
        {\tiny Estimated NVS} \\
        \includegraphics[width =0.7\textwidth]{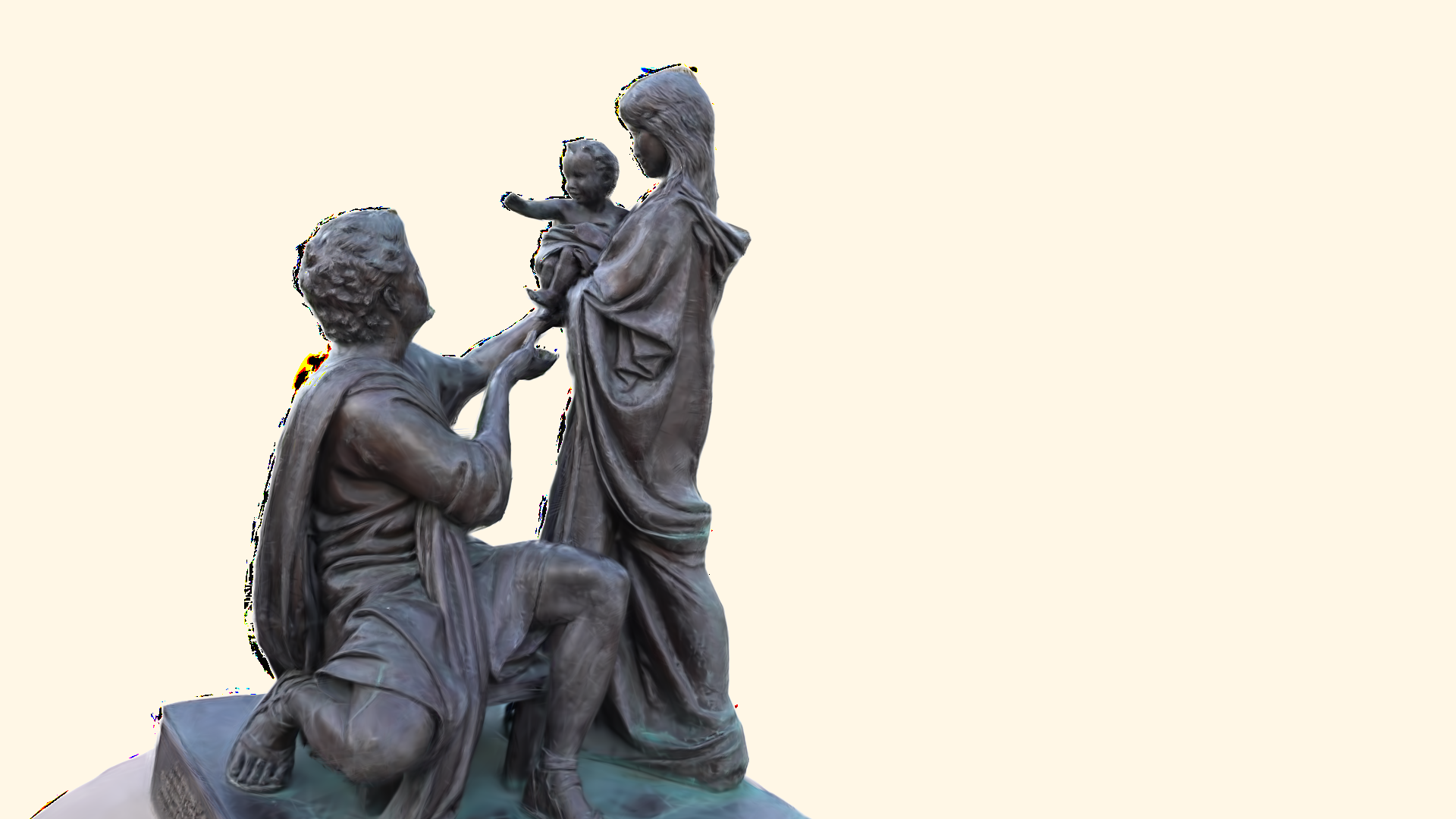}
        \end{minipage}\\[0.5ex]
    \end{minipage}\hfill
    \begin{minipage}[m]{0.6\textwidth}
    \centering
    \end{minipage}\hfill
\end{minipage}\hfill\\[1ex]

\begin{minipage}[outer sep=0]{0.48\textwidth}
    \centering
    {\scriptsize Counter}
    \vspace{.5ex}
    \hrule
    \vspace{.5ex}
    \begin{minipage}[m]{\textwidth}
        \centering
        \includegraphics[width=0.8\textwidth]{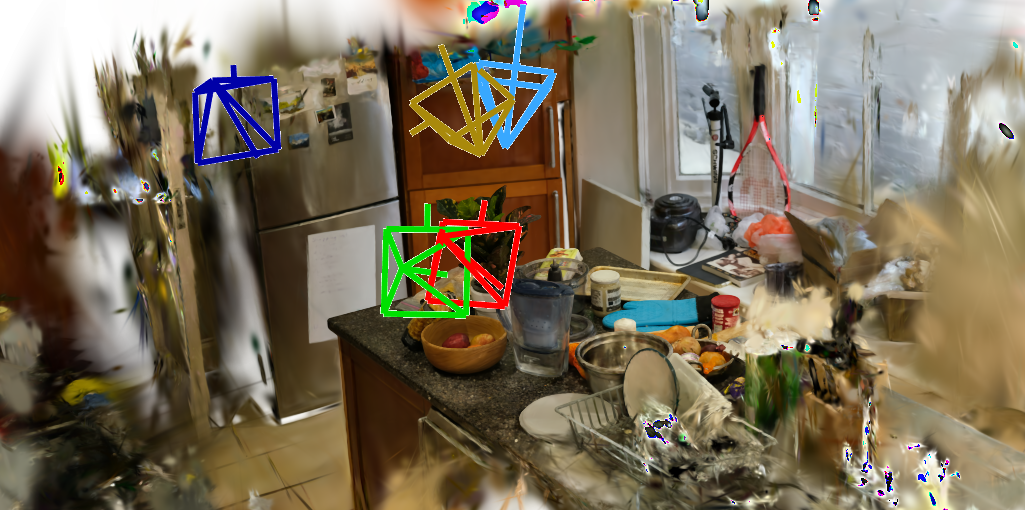} \\
    \end{minipage}
    \begin{minipage}[m]{\textwidth}
        \hspace{+7pt}
        \begin{minipage}[m]{0.49\textwidth}
        \centering
            {\tiny Target image} \\
            \includegraphics[width =0.7\textwidth]{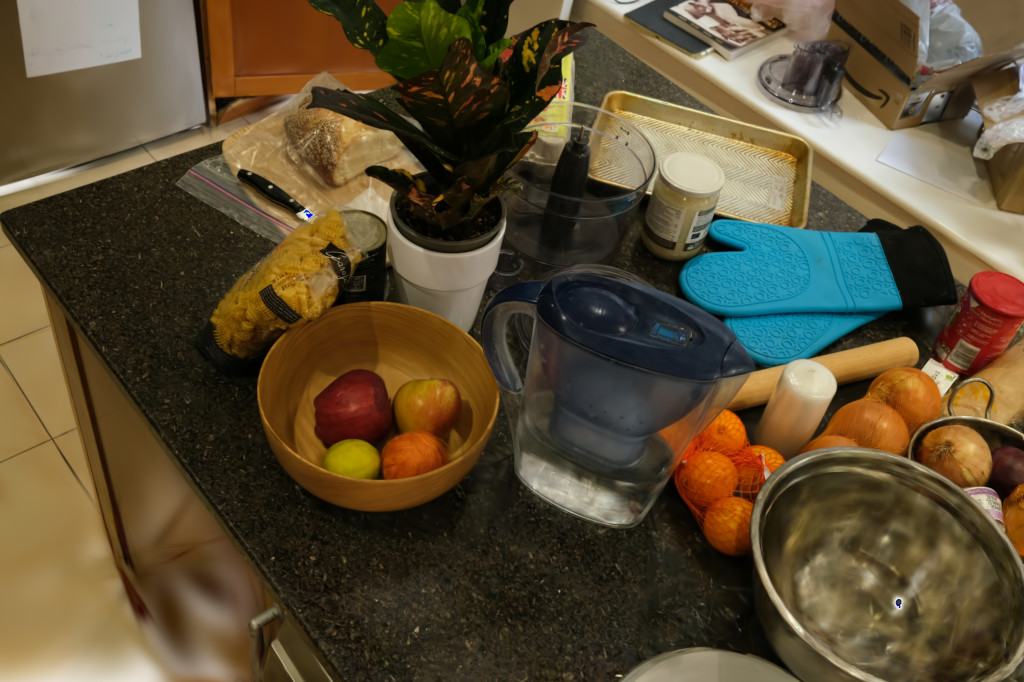}
        \end{minipage}
        \hspace{-25pt}
        \begin{minipage}[m]{0.49\textwidth}
        \centering
        {\tiny Estimated NVS} \\
        \includegraphics[width =0.7\textwidth]{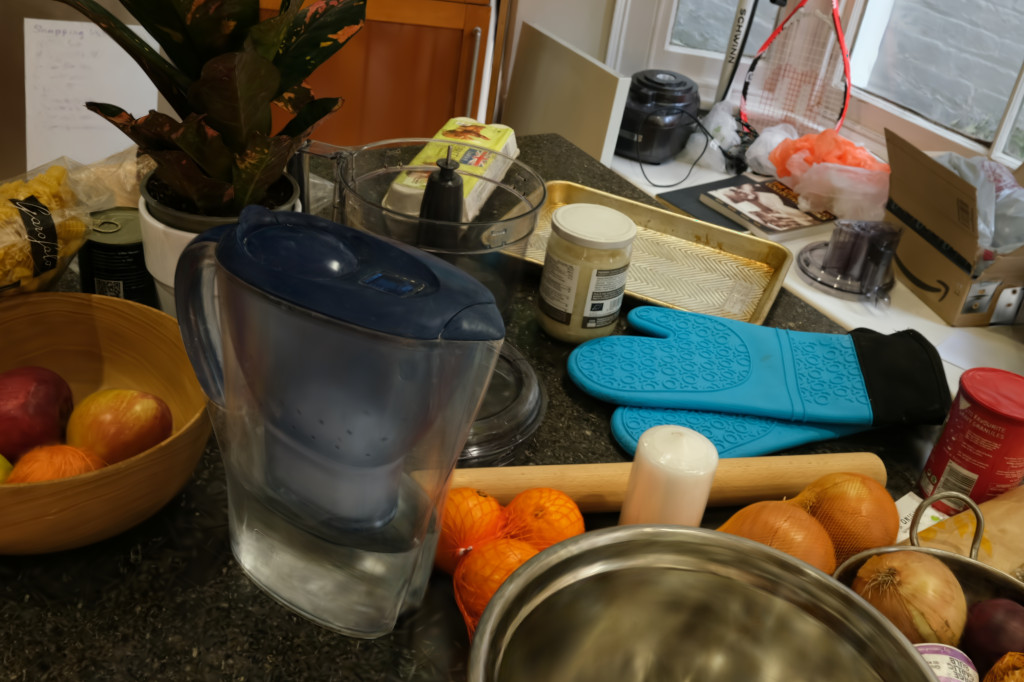}
        \end{minipage}\\[0.5ex]
    \end{minipage}\hfill
    \begin{minipage}[m]{0.6\textwidth}
    \centering
    \end{minipage}\hfill
\end{minipage}
\begin{minipage}[outer sep=0]{0.48\textwidth}
    \centering
    {\scriptsize Bonsai}
    \vspace{.5ex}
    \hrule
    \vspace{.5ex}
    \begin{minipage}[m]{\textwidth}
        \centering
        \includegraphics[width=0.71\textwidth]{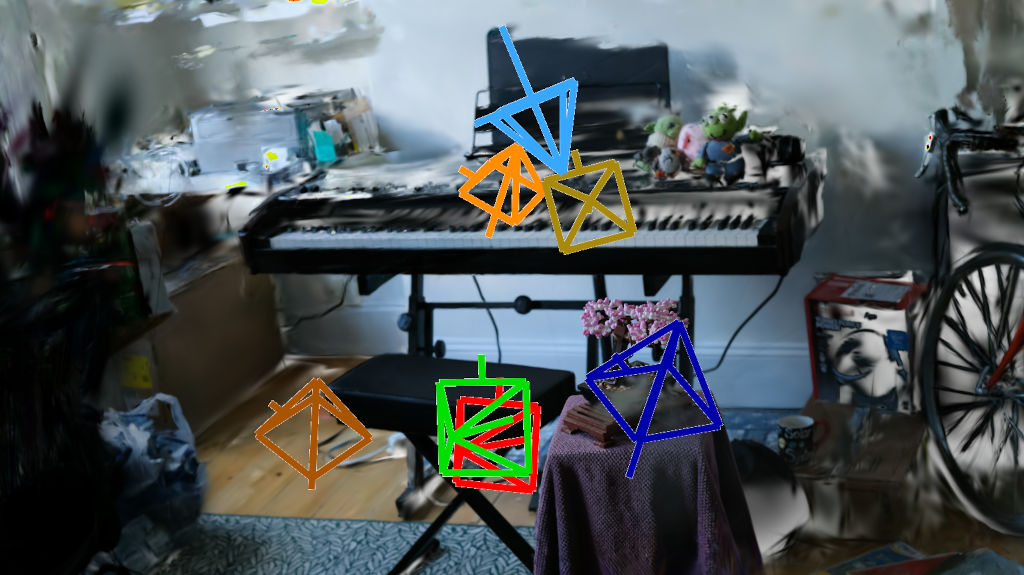} \\
    \end{minipage}
    \begin{minipage}[m]{\textwidth}
        \hspace{+7pt}
        \begin{minipage}[m]{0.49\textwidth}
        \centering
            {\tiny Target image} \\
            \includegraphics[width =0.7\textwidth]{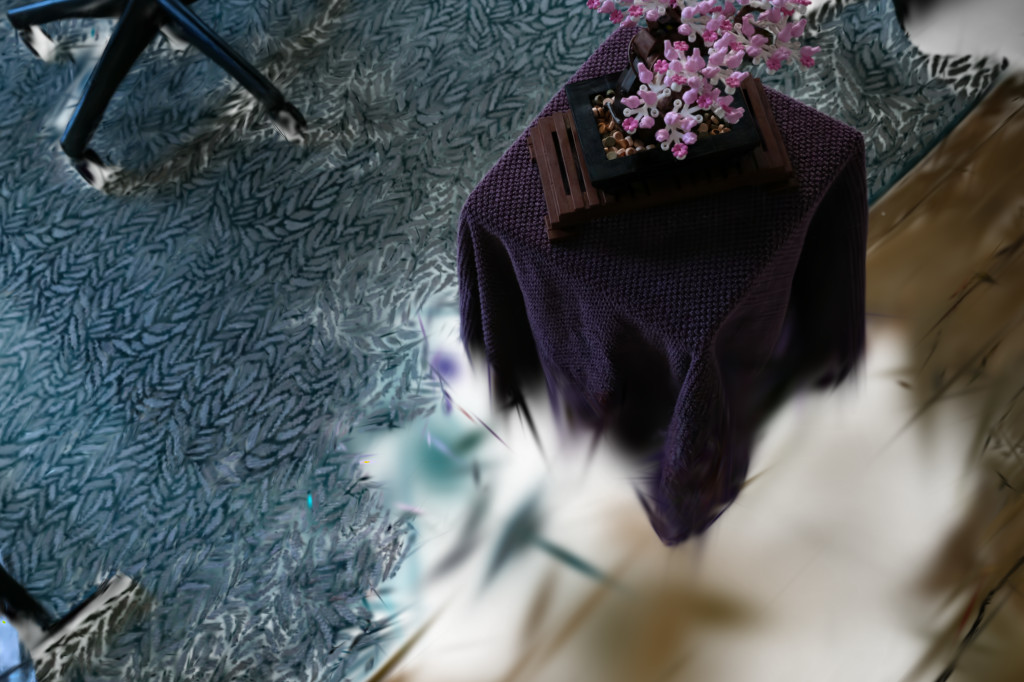}
        \end{minipage}
        \hspace{-25pt}
        \begin{minipage}[m]{0.49\textwidth}
        \centering
        {\tiny Estimated NVS} \\
        \includegraphics[width =0.7\textwidth]{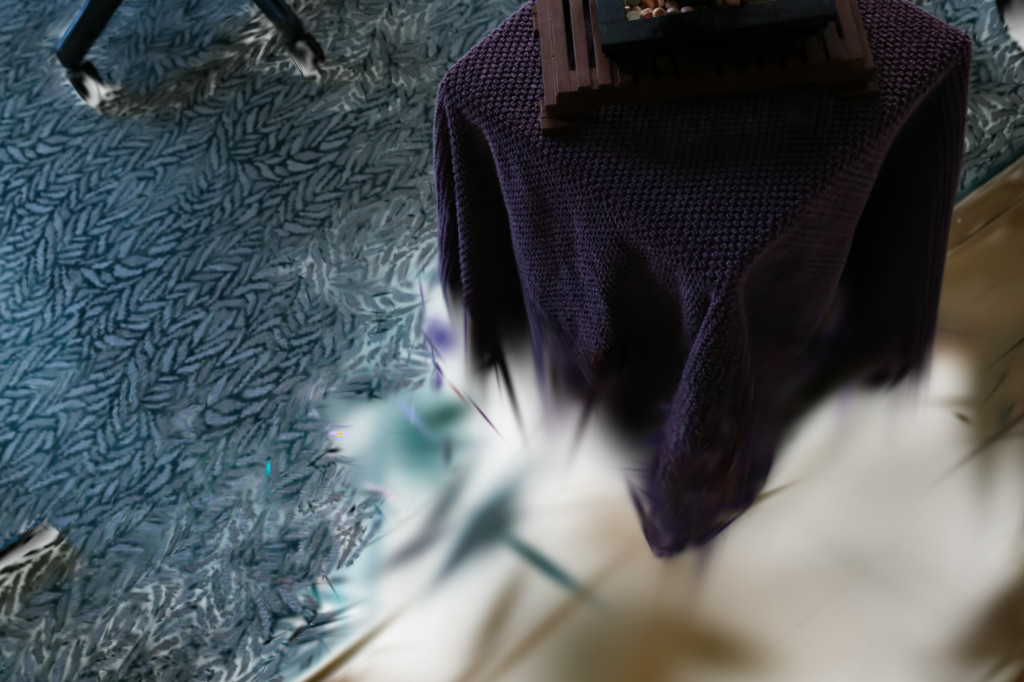}
        \end{minipage}\\[0.5ex]
    \end{minipage}\hfill
    \begin{minipage}[m]{0.6\textwidth}
    \centering
    \end{minipage}\hfill
\end{minipage}\hfill\\[0.5ex]

        \begin{center}
    {\tiny
    \begin{tabular}{ c P{0.03\textwidth} c P{0.05\textwidth} c P{0.06\textwidth} c P{0.07\textwidth} c P{0.06\textwidth} c P{0.07\textwidth} c P{0.09\textwidth} c P{0.09\textwidth} }
     \includegraphics[width=0.04\textwidth]{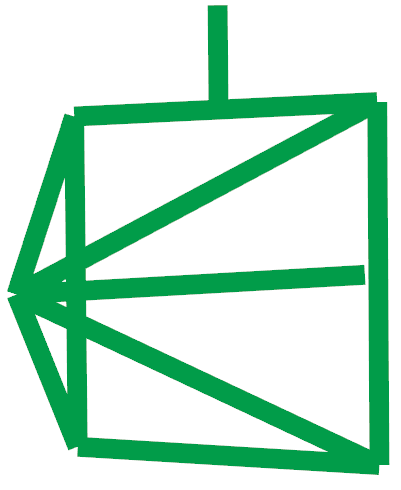} & \vspace{-0.03\textwidth} GT & \includegraphics[width=0.04\textwidth]{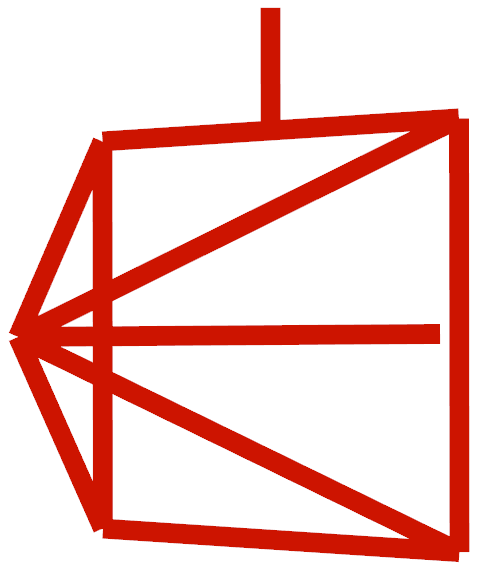} & \vspace{-0.03\textwidth}\ourmethod (Ours) & \includegraphics[width=0.04\textwidth]{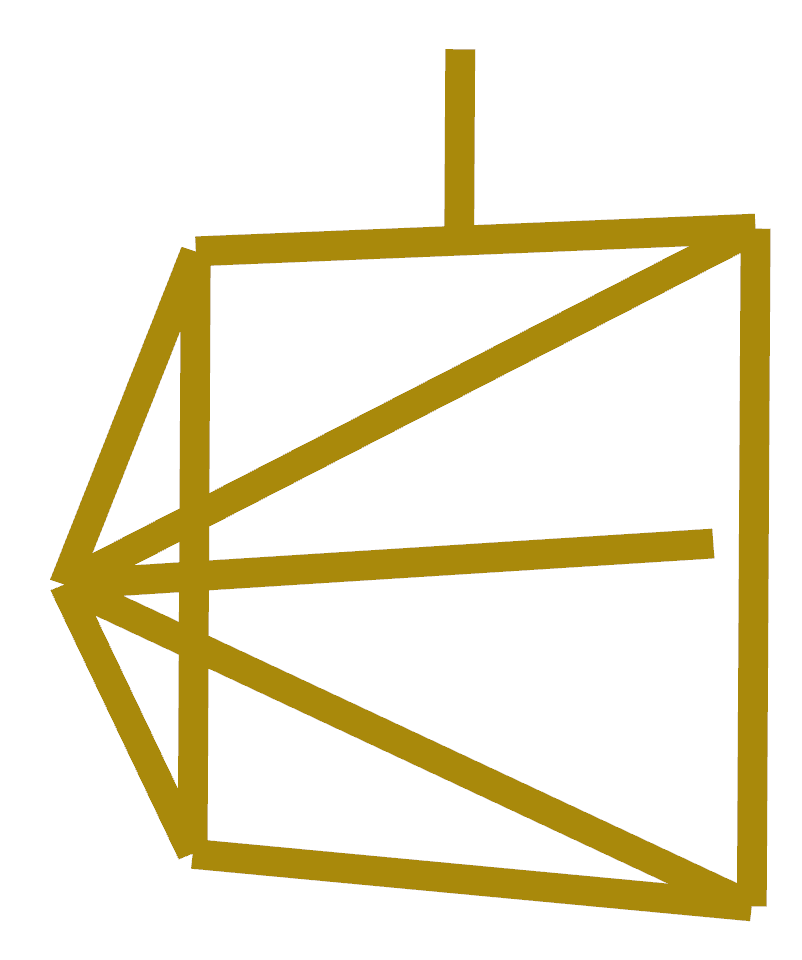} & \vspace{-0.03\textwidth}iNeRF w/ prior & 
     \includegraphics[width=0.04\textwidth]{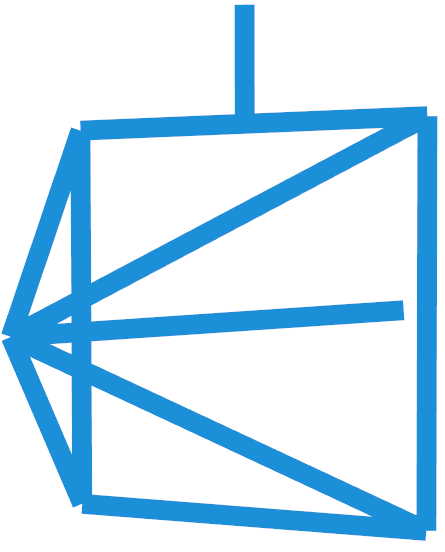} & \vspace{-0.03\textwidth}iNeRF w/o prior & 
     \includegraphics[width=0.04\textwidth]{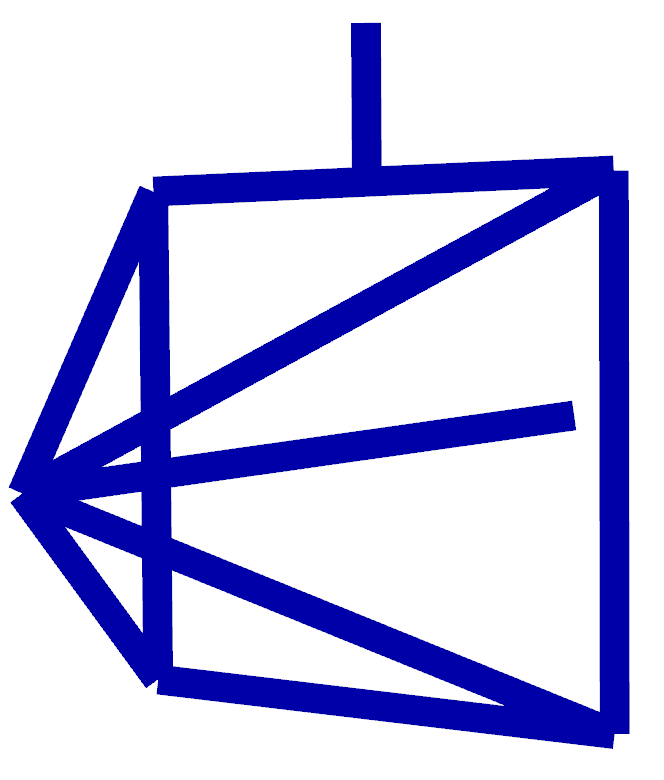} & \vspace{-0.03\textwidth} Parallel iNeRF w/ prior & 
     \includegraphics[width=0.04\textwidth]{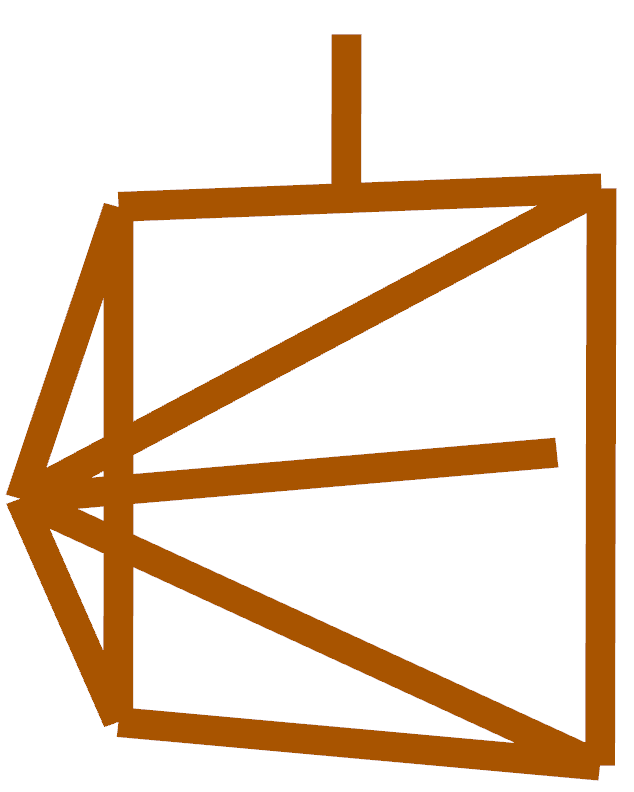} & \vspace{-0.03\textwidth}Parallel iNeRF w/o prior & 
     \includegraphics[width=0.04\textwidth]{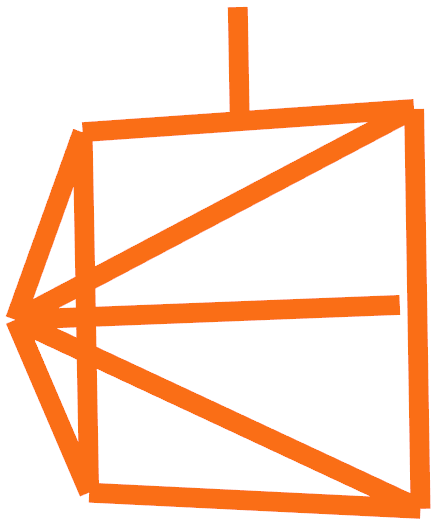} & \vspace{-0.03\textwidth}NeMo + VoGE w/ prior & \includegraphics[width=0.04\textwidth]{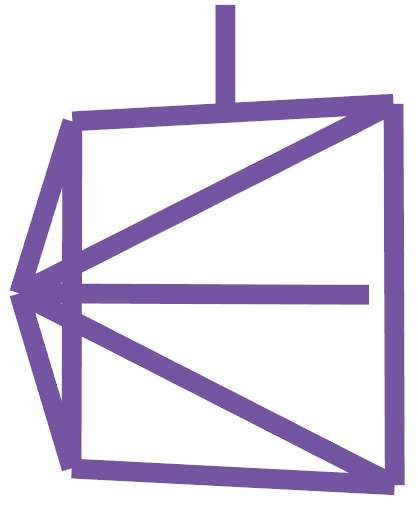} & \vspace{-0.03\textwidth}NeMo + VoGE w/o prior
     \\ 
    \end{tabular}
    }
    \end{center}
    
    }
    \vspace{-2.0ex}
    \caption[XYZ]{\label{fig:qualitative_results} 
    The illustration presents qualitative results from Tanks \& Temple (upper row) and Mip-NeRF 360\textdegree\ (lower row) datasets. 
    Each scene showcases the target images (bottom left) along with their corresponding Novel View Synthesis (NVS) outputs (bottom right), derived from the camera poses estimated by \ourmethod (located on the top). 
    Furthermore, the estimated camera poses from the comparative baselines are visualized, with distinct colors as indicated in the image legend. 
    The NVS of each scene is rendered based on the provided 3DGS model.
    Please check the supplementary material for more qualitative results.
    }
    \vspace{-8pt}
\end{figure*}

\noindent\textbf{Qualitative Analysis:}
Figure~\ref{fig:qualitative_results} illustrates qualitative findings revealing notable observations. 
Particularly, we notice that the estimated poses exhibit proximity to the object relative to ground truth, attributable to the quantization effect introduced by the \raygenerationmethod. 
The qualitative findings verify the quantitative outcomes, albeit occasional inconsistencies in results, such as in the Counter scene, with the analysis-by-synthesis approaches showcasing a total incoherent output in regards to the overall scene (notice how the estimated poses are completely off the target). 
Moreover, the performance of \ourmethod demonstrates consistency across varied scenarios, encompassing single-object instances and indoor settings, despite substantial variations in the models utilized.

\vspace{-2pt}
\subsection{Ablation studies}

Our ablation studies involve the analysis of the number of rays selected for the pose estimation (Tab.~\ref{tab:top_k}), the number of rays that we cast from a \raygenerationmethod (Tab.~\ref{tab:cast_rays}) as well as the different feature size on the MLP channels (Tab.~\ref{tab:mlp_channels}). The supplementary material contains additional ablations that analyze 6DGS performance with low-quality 3DGS models.

We find that the number of selected rays mainly affects the angular error, while the translation error remains relatively stable. 
Increasing the number of rays decreases the angular error but slightly increases the translation error, likely due to less confident rays contributing to the pose estimation. 
The optimal balance between translation and angular errors is achieved between 100 to 150 rays, with 100 being the best.
The slight increase in error with more $N_{top}$ rays is due to introducing rays not pointing precisely to the camera's optical center.
Similar to what we observed in the qualitative examples, the noisy rays make the weighted Least Squares estimating the camera closer to the object than it actually is.

Regarding the impact of the varying number of rays cast from the \raygenerationmethodplural, the angular error tends to remain relatively constant across different configurations. 
In contrast, the translation error decreases when 50 cast rays are used, and then increases again.
This behavior is connected to network generalization capability.
Increasing the number of rays allows the network to fit the training set better, but at test time, it makes the network more prone to noise and consequently selecting the wrong rays, thus increasing the error.
We observe this generalization issue when increasing the MLP channels, see Tab.~\ref{tab:mlp_channels}, particularly given the limited and uneven distribution of training images ($\approx 150$).
Moreover, the processing time increases proportionally with the number of rays and the MLP channels; upon exceeding the default values for rays and feature size, a notable surge in processing time is observed, reaching approximately $10 fps$ and $13 fps$, respectively.

\begin{table}[t]
    \caption{Ablation study on the number of rays selected for pose estimation, on the rays cast from each ellipsoid and on the MLP channels using Mip-NeRF 360~\cite{barron2022mipnerf360}. \underline{Underline} indicates the default values used.}
    \label{tab:ablation_top_k}
    \centering
    \aboverulesep=0ex %
    \belowrulesep=0ex %
    \vspace{-2mm}
    \begin{adjustbox}{max width=1\linewidth}
    \begin{minipage}{0.28\linewidth}
        \begin{subtable}{\textwidth}
            \caption{Number of rays used for pose estimation.}
            \label{tab:top_k}
            \begin{adjustbox}{max width=1\linewidth}
                \begin{NiceTabular}[t]{|cccc|}
                    \CodeBefore
                      \rowcolor{gray!20}{1}
                    \Body
                    \toprule
                    $N_{top}$ & MAE (\textdegree)$\downarrow$ & MTE (u)$\downarrow$ & Time (s) \\
                    \midrule
                    20 & \cellcolor{red!5}29.0 & \cellcolor{red!5}0.0235 & \cellcolor{green!30}0.03 \\
                    50 & \cellcolor{red!5}26.3 & \cellcolor{red!5}0.0227 & \cellcolor{orange!15}0.04 \\
                    \underline{100} & \cellcolor{green!30}\underline{24.3} & \cellcolor{green!30}\underline{0.0217} & \cellcolor{red!5}\underline{0.06} \\
                    150 & \cellcolor{orange!15}24.4 & \cellcolor{orange!15}0.0219 & \cellcolor{red!5}0.9 \\
                    200 & \cellcolor{red!5}24.5 & \cellcolor{red!5}0.0222 & \cellcolor{red!5}0.11 \\
                    \bottomrule
                \end{NiceTabular}
            \end{adjustbox}
        \end{subtable}
    \end{minipage}%
    \hspace{0.5pt}
    \begin{minipage}{0.36\linewidth}
        \begin{subtable}{\textwidth}
        \caption{Number of cast rays per ellipsoid.}
        \label{tab:cast_rays}
            \begin{adjustbox}{max width=1\linewidth}
                \begin{NiceTabular}{|cccc|}
                    \CodeBefore
                      \rowcolor{gray!20}{1}
                    \Body
                    \toprule
                    \# of cast rays & MAE (\textdegree)$\downarrow$ & MTE (u)$\downarrow$ & Time (s) \\
                    \midrule
                    20 & \cellcolor{red!5}29.0 & \cellcolor{red!5}0.0235 & \cellcolor{green!30}0.04 \\
                    35 & \cellcolor{red!5}24.7 & \cellcolor{orange!15}0.0220 & \cellcolor{green!30}0.04 \\
                    \underline{50} & \cellcolor{green!30}\underline{24.3} & \cellcolor{green!30}\underline{0.0217} & \cellcolor{orange!15}\underline{0.06} \\
                    65 & \cellcolor{red!5}25.1 & \cellcolor{orange!15}0.0218 & \cellcolor{red!5}0.09 \\
                    80 & \cellcolor{red!5}25.2 & \cellcolor{red!5}0.0221 & \cellcolor{red!5}0.15 \\
                    \bottomrule
                \end{NiceTabular}
            \end{adjustbox}
        \end{subtable}
    \end{minipage}
    \hspace{-7pt}
    \begin{minipage}{0.38\linewidth}   
    \vspace{-17pt}
        \begin{subtable}{\textwidth}
            \caption{MLP channel feature size.\vspace{9pt}}
            \label{tab:mlp_channels}
            \centering
            \begin{adjustbox}{max width=0.9\linewidth}
                \begin{NiceTabular}{|cccc|}
                    \CodeBefore
                      \rowcolor{gray!20}{1}
                    \Body
                    \toprule
                    MLP channels & MAE (\textdegree)$\downarrow$ & MTE (u)$\downarrow$ & Time (s) \\
                    \midrule
                    256 & \cellcolor{orange!15}29.4 & \cellcolor{red!5}0.0273 & \cellcolor{green!30}0.04 \\
                    \underline{512} & \cellcolor{green!30}\underline{24.3} & \cellcolor{green!30}\underline{0.0217} & \cellcolor{orange!15}\underline{0.06} \\
                    1024 & \cellcolor{red!5}30.1 & \cellcolor{orange!15}0.0228 & \cellcolor{red!5}0.27 \\
                    \bottomrule
                \end{NiceTabular}
            \end{adjustbox}
        \end{subtable}
    \end{minipage}
    \end{adjustbox}
    \vspace{-5pt}
\end{table}

\vspace{-4pt}
\section{Conclusions}\label{sec:conclusion}

In this study, we proposed a novel ray sampling by attention method for estimating 6DoF camera poses from a single image and a 3DGS scene model. 
Our analytical evaluation demonstrates its robustness and efficiency without requiring initialization, up to $22\%$ in accuracy and while being faster by a big margin, approx. 94x faster. 
Furthermore, the proposed method formulates and utilizes a novel ray generation methodology in order to explore diverse camera pose hypotheses in accordance to an effective attention mechanism.
Our method exhibits enhanced robustness across real-world datasets and holds promise for real-time deployment in robotics and other fields. 
Future research endeavors will focus on improving accuracy and extending applicability to diverse scenes and objects.

\noindent\textbf{Limitations.} 
The main constraint of \ourmethod is the need for retraining with each new scene. This could be mitigated with meta-learning, particularly when similar objects or scenes are under consideration.

\newpage
\section*{Acknowledgments}
This work is part of the RePAIR project that has received funding from the European Union's Horizon 2020 research and innovation programme under grant agreement No.~964854.
This work has also received funding from the European Union’s Horizon Europe research and innovation programme under grant agreement No.~101092043, project AGILEHAND (Smart Grading, Handling and Packaging Solutions for Soft and Deformable Products in Agile and Reconfigurable Lines).
We thank S. Fiorini for the discussion on the optimizers.

\bibliographystyle{splncs04}
\bibliography{main}

\title{\ourmethod: 6D Pose Estimation from a Single \\Image and a 3D Gaussian Splatting Model - Supplementary material}

\titlerunning{\ourmethod}

\author{Bortolon Matteo\inst{1, 2, 3}\orcidlink{0000-0001-8620-1193} \and
Theodore Tsesmelis\inst{1}\orcidlink{0000-0001-9290-2383} \and
Stuart James\inst{1,4}\orcidlink{0000-0002-2649-2133} \and \\
Fabio Poiesi\inst{2}\orcidlink{0000-0002-9769-1279} \and
Alessio {Del Bue}\inst{1}\orcidlink{0000-0002-2262-4872} }

\institute{PAVIS, Fondazione Istituto Italiano di Tecnologia (IIT), Genoa, IT \and
TeV, Fondazione Bruno Kessler (FBK), Trento, IT \and
Università di Trento, Trento, IT \and
Durham University, Durham, UK}

\authorrunning{M.~Bortolon \etal}

\maketitle

\section{Introduction}
The provided supplementary material shows extra details, results, and figures to support the main findings presented in the main manuscript. 
In Section~\ref{sec:inference_time} we present the inference time performance of each method for the task of 6D pose estimation. 
Section~\ref{sec:additional_qualitative_results} offers additional figures showing qualitative results and discussions as mentioned in the main paper. 
These extra results aim to make the study clearer and more complete, providing further useful insights into our proposed 6DGS pipeline.
\freefootnote{Project page: \url{https://mbortolon97.github.io/6dgs/}\\
Corresponding author: \email{mbortolon@fbk.eu}}

\section{Inference time}
\label{sec:inference_time}

\begin{table}[!b]
\centering
\caption{Average computation time to estimate the pose of an image. 
Comparison between our method and state-of-the-art approaches across the two datasets.
\textbf{Bold} font indicates best performance. 
Time is reported in seconds.}
\label{tab:timing_results}
\begin{tabular}{lccc}
\toprule
 & Tanks and Temples & Mip-NeRF 360 \\
 \midrule
iNeRF (pose prior by \cite{yen2020inerf}) & 6.321 & 8.214 \\
\midrule
NeMo+VoGE (pose prior by \cite{yen2020inerf}) & 251.4 & 290.4 \\
\midrule
Parallel iNeRF (pose prior by \cite{yen2020inerf}) & 16.9 & 16.8 \\
\midrule
iNeRF (random pose) & 6.2 & 8.5 \\
\midrule
NeMo+VoGE (random pose) & 240.5 & 284.4 \\
\midrule
Parallel iNeRF (random pose) & 16.5 & 17.1 \\
\midrule
\ourmethod (Ours) (no pose) & \textbf{0.05} & \textbf{0.06} \\
\bottomrule
\end{tabular}
\end{table}

In Tab.~\ref{tab:timing_results} we detail the timings of each method.
It can be noted from the table that our approach is the only one that is able to run under $1s$ and near to real-time performance at 16FPS.
In general, it is possible to notice how time slightly increases in the Mip-NeRF 360\textdegree dataset due to the higher complexity of the scenes.
The increase in computational complexity can be attributed to several factors, including the size of the neural models required for iNeRF and Parallel iNeRF, as well as the quantity of ellipsoids used in NeMo+VoGE for each of the two datasets (\ie $\approx150000$ ellipsoids for Tanks\&Temples and $\approx300000$ for Mip-NeRF 360\textdegree, while in NeRF models Tanks\&Temples use $\approx50\%$ less parameters than Mip-NeRF 360\textdegree). 
As it can be noticed \ourmethod is able to handle both datasets and especially the the Mip-NeRF 360\textdegree scenes without issues.

Considering the different methods, we can notice that NeMo+VoGE requires the most time.
This derive from the processing algorithm that struggles with large number of ellipsoids.
In regards to the analysis-by-synthesis methods, iNeRF represents the fastest method, with Parallel NeRF being the second fastest. However, all methods show a considerable higher running time in contrast to \ourmethod.

\section{Additional ablations studies}

To assess our method's robustness to 3DGS model quality, we conducted two experiments using the Mip360° dataset's Bonsai and Bicycle scenes.
First, we investigated the effect of 3DGS model accuracy on pose estimation by introducing varying noise levels to the Gaussian centers. Second, we analyzed the performance of our method under sparse viewpoint conditions.

The results below show that \ourmethod is resilient for translation and minimal increase for rotational error, both with high noise level and even with only six views ($\approx60\degree$ between viewpoints).
\begin{table}[h]
\renewcommand{\arraystretch}{0.8}
\tabcolsep 5pt
\resizebox{\columnwidth}{!}{%
\begin{tabular}{cccccc}
    \toprule
     Gauss. noise (\textperthousand) & MAE (\textdegree) & MTE (u) & No. views & MAE (\textdegree) & MTE (u) \\ 
     \cmidrule(l{0.5em}r{0.5em}){1-3}
     \cmidrule(l{0.5em}r{0.5em}){4-6}
    0.0\textperthousand u (None) & 11.3 & 0.024 & All & 11.3 & 0.024 \\
    0.1\textperthousand u & 18.3 & 0.025 & 12 & 15.5 & 0.024 \\
    0.5\textperthousand u & 22.5 & 0.025 & 6 & 18.6 & 0.026 \\
    \bottomrule
\end{tabular}
}
\vspace{-3.0mm}
\end{table}

\section{Additional qualitative results}
\label{sec:additional_qualitative_results}

\begin{figure*}[!b]
\centering
{\fontfamily{phv}\selectfont 
\begin{minipage}[outer sep=0]{0.48\textwidth}
    \centering
    {\footnotesize Barn}
    \vspace{.5ex}
    \hrule
    \vspace{.5ex}
    \begin{minipage}[m]{\textwidth}
        \centering
        \includegraphics[width=0.7\textwidth]{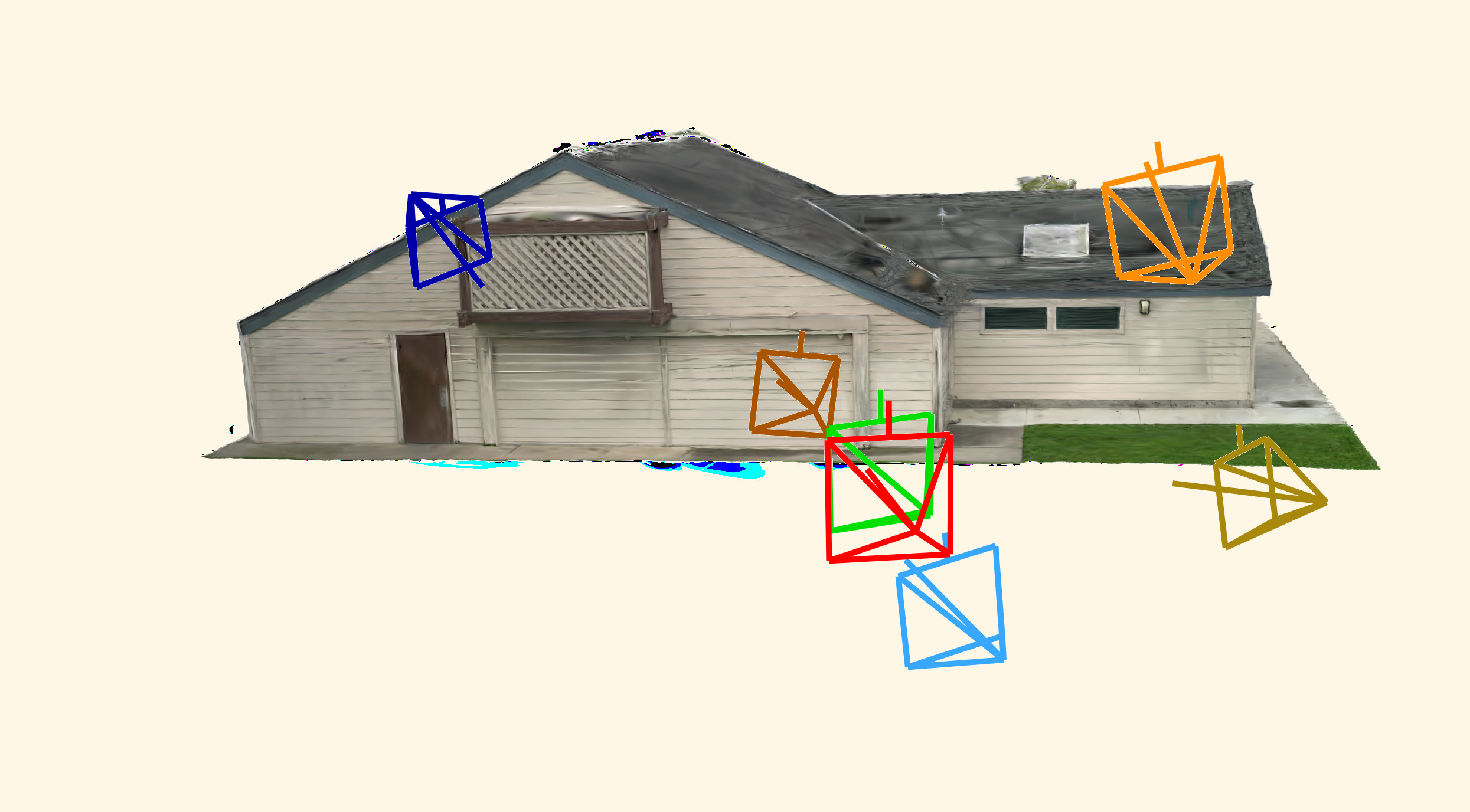} \\
    \end{minipage}
    \begin{minipage}[m]{\textwidth}
        \hspace{+7pt}
        \begin{minipage}[m]{0.49\textwidth}
        \centering
            {\scriptsize Target image} \\
            \includegraphics[width =0.7\textwidth]{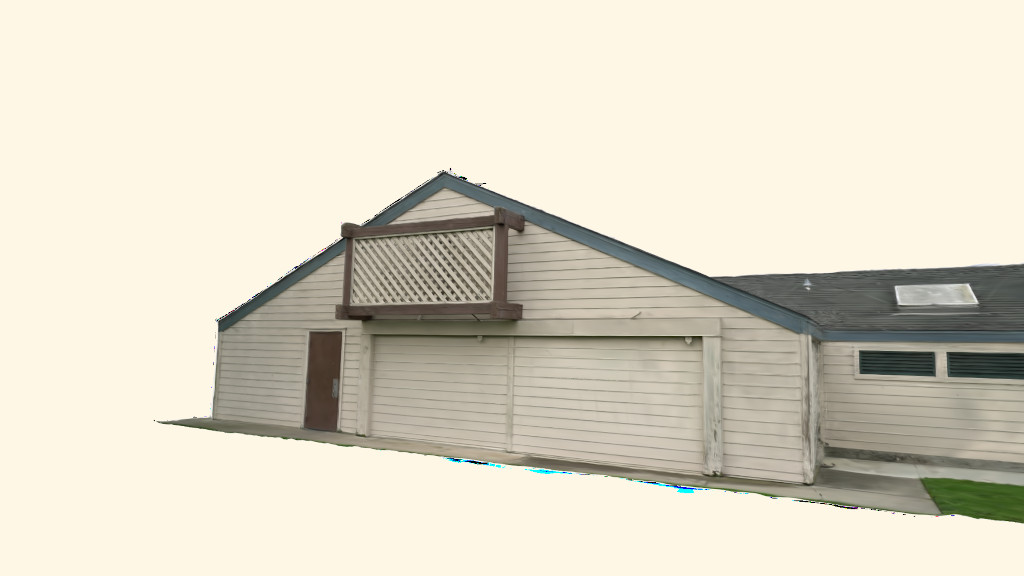}
        \end{minipage}
        \hspace{-25pt}
        \begin{minipage}[m]{0.49\textwidth}
        \centering
        {\scriptsize Estimated NVS} \\
        \includegraphics[width =0.7\textwidth]{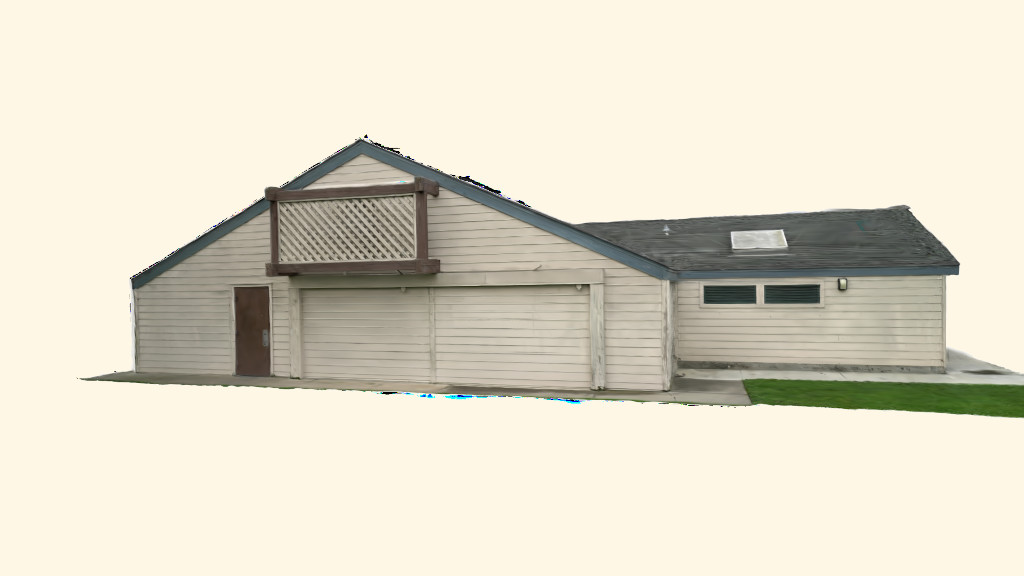}
        \end{minipage}\\[0.5ex]
    \end{minipage}\hfill
    \begin{minipage}[m]{0.6\textwidth}
    \centering
    \end{minipage}\hfill
\end{minipage}
\begin{minipage}[outer sep=0]{0.48\textwidth}
    \centering
    {\footnotesize Caterpillar}
    \vspace{.5ex}
    \hrule
    \vspace{.5ex}
    \begin{minipage}[m]{\textwidth}
        \centering
        
        \includegraphics[width=0.7\textwidth]{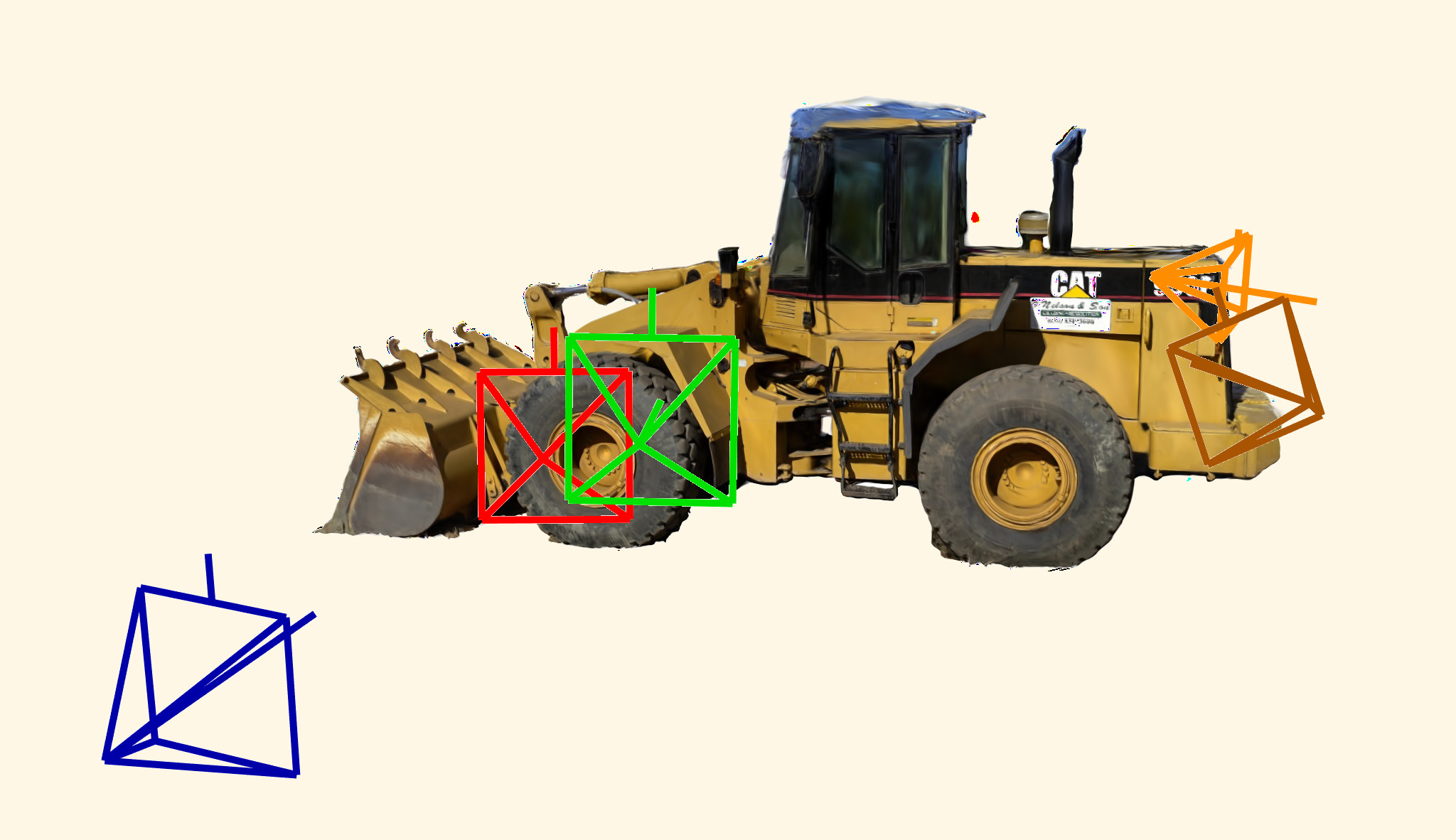} \\
    \end{minipage}
    \begin{minipage}[m]{\textwidth}
        \hspace{+7pt}
        \begin{minipage}[m]{0.49\textwidth}
        \centering
            {\scriptsize Target image} \\
            \includegraphics[width =0.7\textwidth]{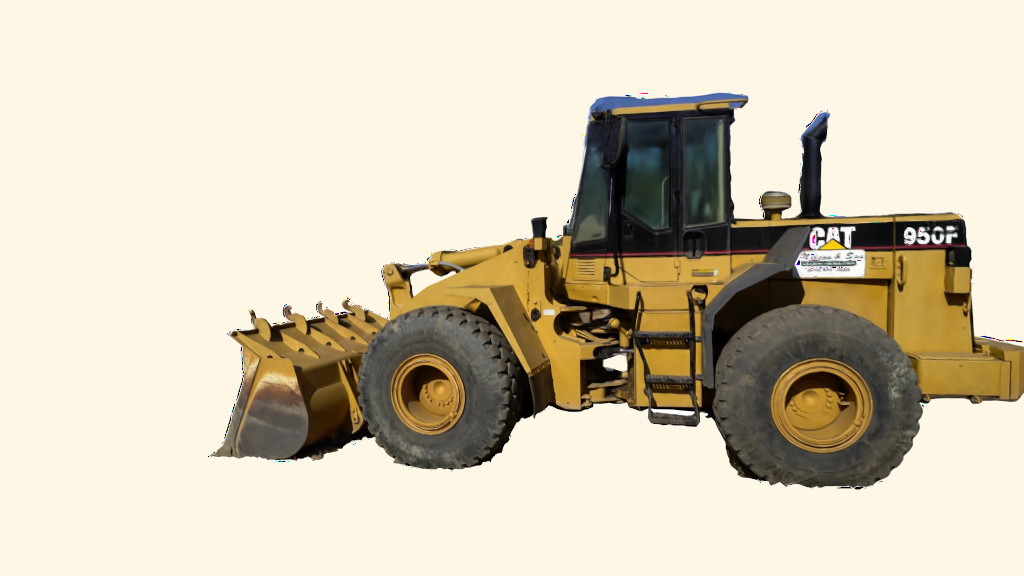}
        \end{minipage}
        \hspace{-25pt}
        \begin{minipage}[m]{0.49\textwidth}
        \centering
        {\scriptsize Estimated NVS} \\
        \includegraphics[width =0.7\textwidth]{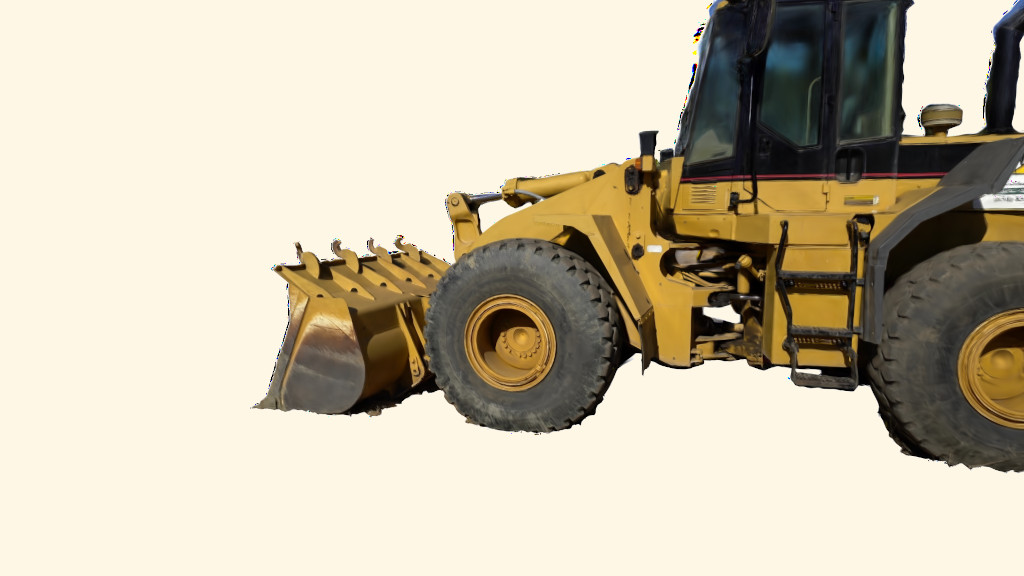}
        \end{minipage}\\[0.5ex]
    \end{minipage}\hfill
    \begin{minipage}[m]{0.6\textwidth}
    \centering
    \end{minipage}\hfill
\end{minipage}\hfill

\begin{minipage}[outer sep=0]{0.48\textwidth}
    \centering
    {\footnotesize Ignatius}
    \vspace{.5ex}
    \hrule
    \vspace{.5ex}
    \begin{minipage}[m]{\textwidth}
        \centering
        \includegraphics[width=0.7\textwidth]{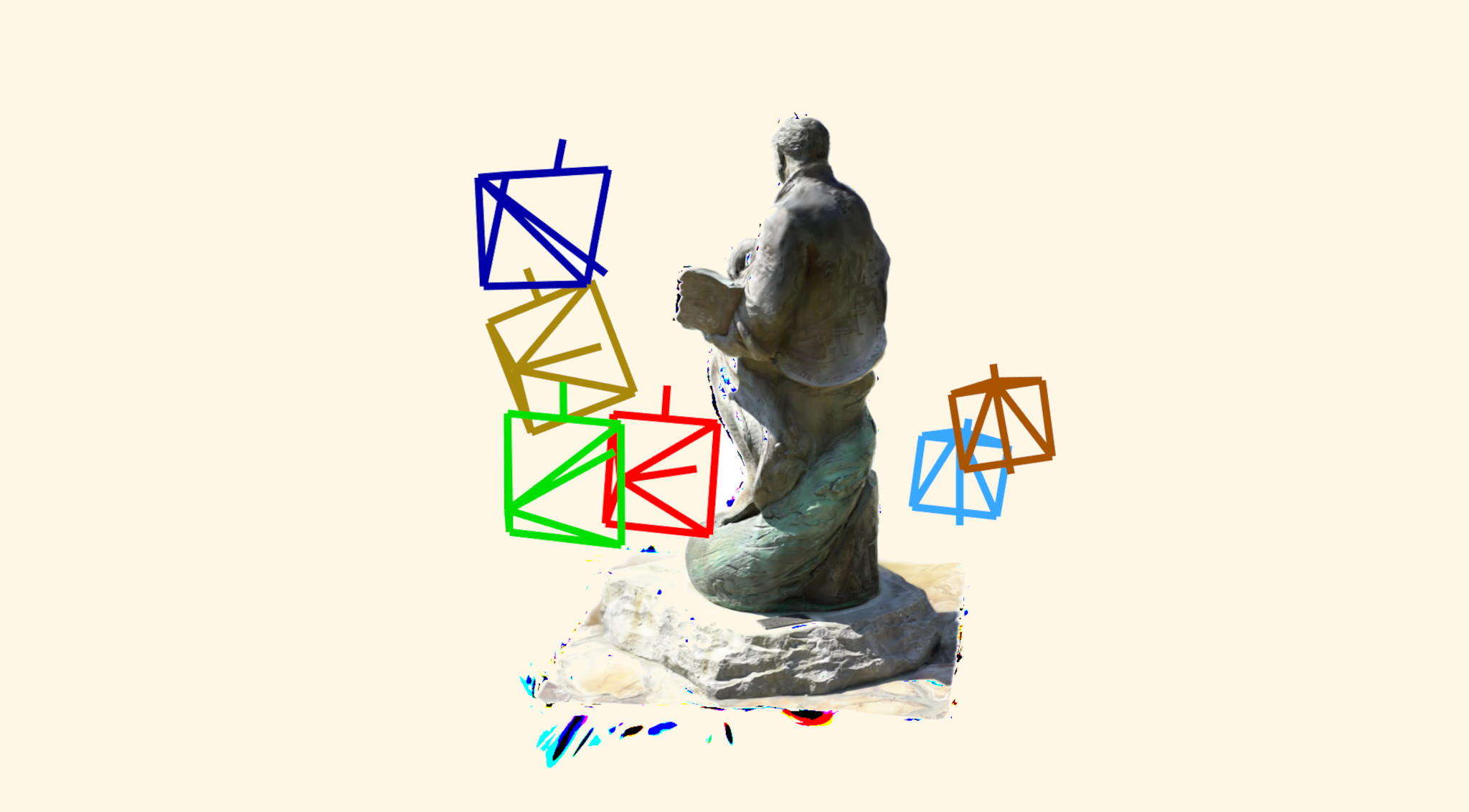} \\
    \end{minipage}
    \begin{minipage}[m]{\textwidth}
        \hspace{+7pt}
        \begin{minipage}[m]{0.49\textwidth}
        \centering
            {\scriptsize Target image} \\
            \includegraphics[width =0.7\textwidth]{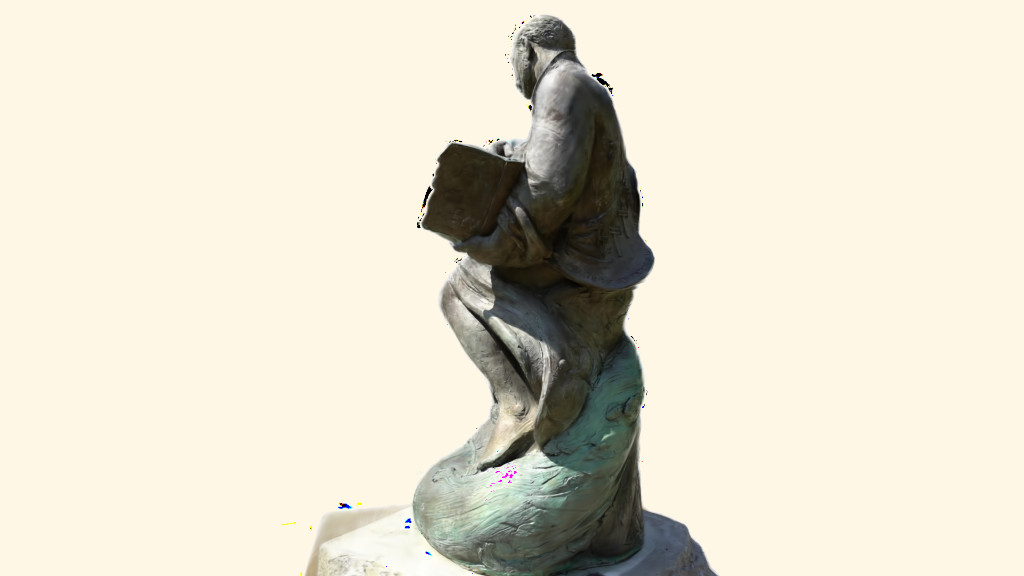}
        \end{minipage}
        \hspace{-25pt}
        \begin{minipage}[m]{0.49\textwidth}
        \centering
        {\scriptsize Estimated NVS} \\
        \includegraphics[width =0.7\textwidth]{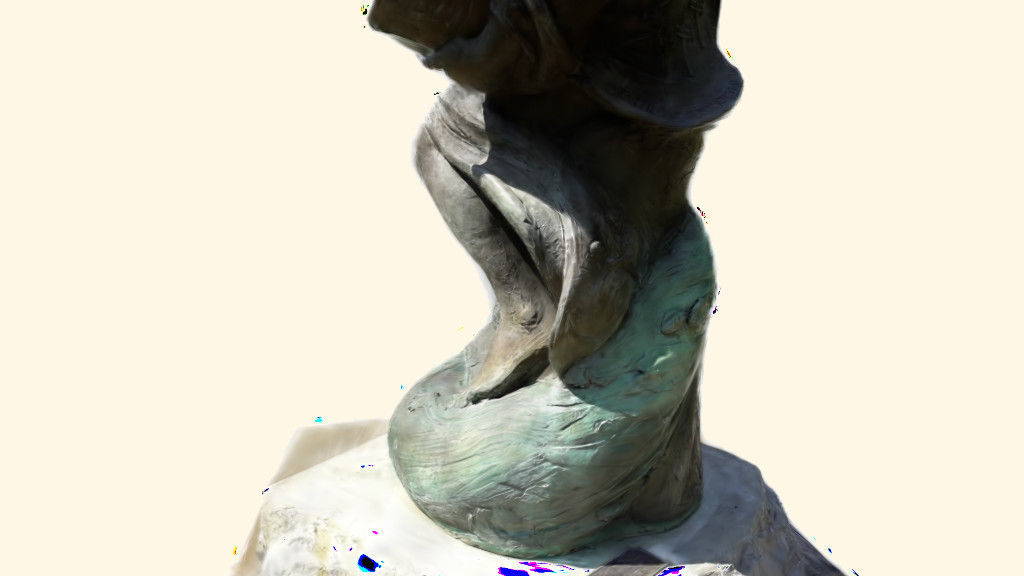}
        \end{minipage}\\[0.5ex]
    \end{minipage}\hfill
    \begin{minipage}[m]{0.6\textwidth}
    \centering
    \end{minipage}\hfill
\end{minipage}
\begin{minipage}[outer sep=0]{0.48\textwidth}
    \centering
    {\footnotesize Truck}
    \vspace{.5ex}
    \hrule
    \vspace{.5ex}
    \begin{minipage}[m]{\textwidth}
        \centering
        \includegraphics[width=0.7\textwidth]{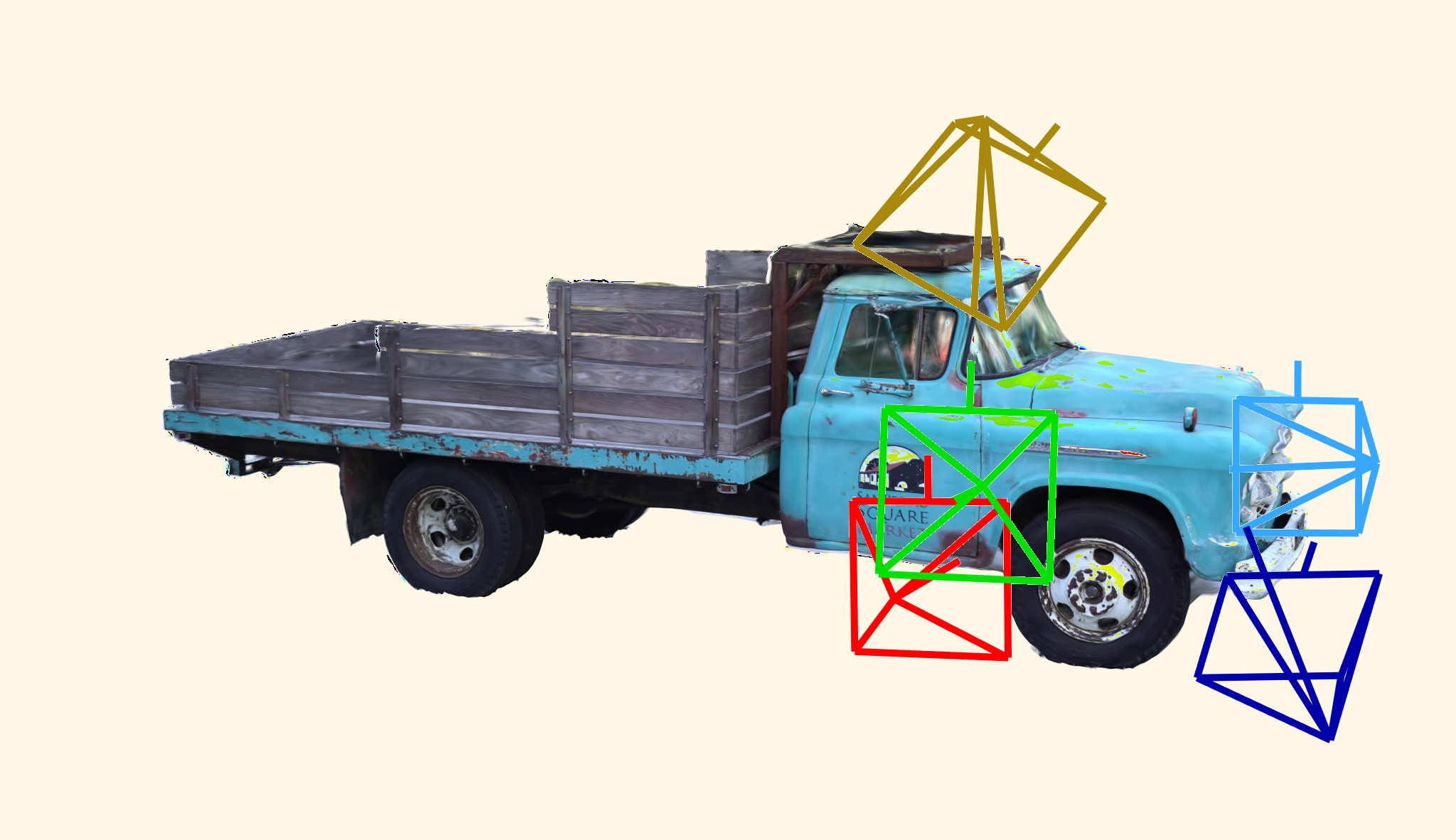} \\
    \end{minipage}
    \begin{minipage}[m]{\textwidth}
        \hspace{+7pt}
        \begin{minipage}[m]{0.49\textwidth}
        \centering
            {\scriptsize Target image} \\
            \includegraphics[width =0.7\textwidth]{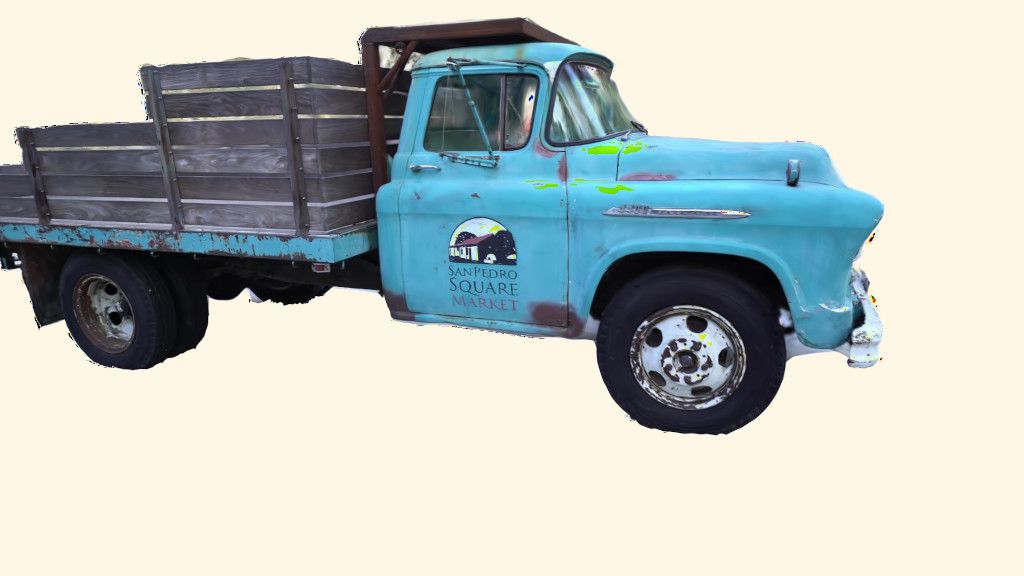}
        \end{minipage}
        \hspace{-25pt}
        \begin{minipage}[m]{0.49\textwidth}
        \centering
        {\scriptsize Estimated NVS} \\
        \includegraphics[width =0.7\textwidth]{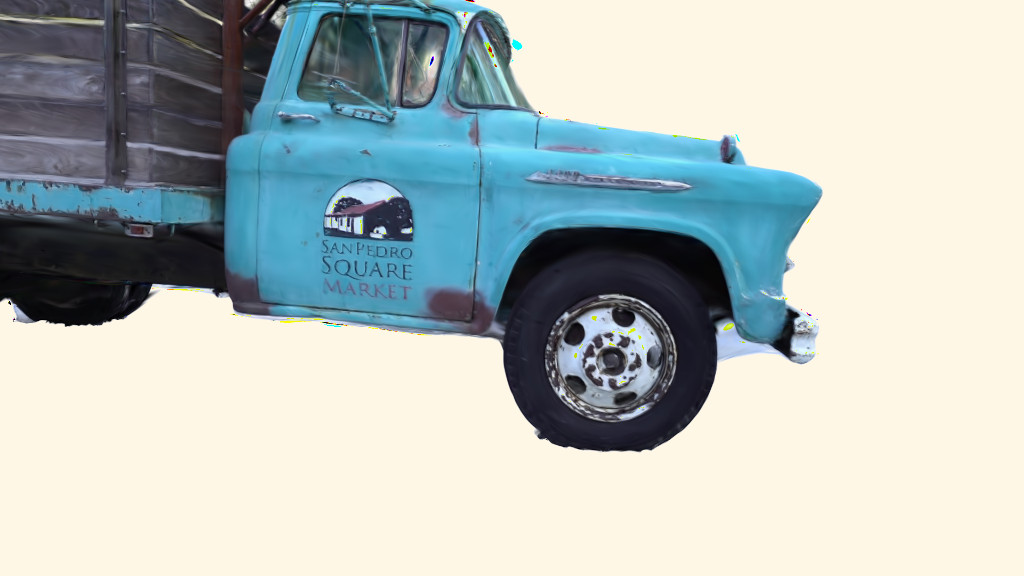}
        \end{minipage}\\[0.5ex]
    \end{minipage}\hfill
    \begin{minipage}[m]{0.6\textwidth}
    \centering
    \end{minipage}\hfill
\end{minipage}\hfill\\[0.5ex]

        \begin{center}
    {\tiny
    \begin{tabular}{ c P{0.03\textwidth} c P{0.05\textwidth} c P{0.06\textwidth} c P{0.07\textwidth} c P{0.06\textwidth} c P{0.07\textwidth} c P{0.09\textwidth} c P{0.09\textwidth} }
     \includegraphics[width=0.04\textwidth]{imgs/qualitative_visualization/cameras/gt.png} & \vspace{-0.03\textwidth} GT & \includegraphics[width=0.04\textwidth]{imgs/qualitative_visualization/cameras/ours.png} & \vspace{-0.03\textwidth}\ourmethod (Ours) & \includegraphics[width=0.04\textwidth]{imgs/qualitative_visualization_2/inerf_wprior.png} & \vspace{-0.03\textwidth}iNeRF w/ prior & 
     \includegraphics[width=0.04\textwidth]{imgs/qualitative_visualization/cameras/inerf_random.png} & \vspace{-0.03\textwidth}iNeRF w/o prior & 
     \includegraphics[width=0.04\textwidth]{imgs/qualitative_visualization_2/pinerf_wprior.png} & \vspace{-0.03\textwidth} Parallel iNeRF w/ prior & 
     \includegraphics[width=0.04\textwidth]{imgs/qualitative_visualization_2/pinerf_woprior.png} & \vspace{-0.03\textwidth}Parallel iNeRF w/o prior & 
     \includegraphics[width=0.04\textwidth]{imgs/qualitative_visualization/cameras/nemo_voge_init.png} & \vspace{-0.03\textwidth}NeMo + VoGE w/ prior & \includegraphics[width=0.04\textwidth]{imgs/qualitative_visualization/cameras/nemo_voge_random.png} & \vspace{-0.03\textwidth}NeMo + VoGE w/o prior
     \\ 
    \end{tabular}
    }
    \end{center}
    
    }
    \vspace{-2.0ex}
    \caption[XYZ]{\label{fig:supp_qualitative_results_tt} 
    Additional scenes from the Tanks\&Temple dataset. 
    For each scene, we show a visualization of the camera poses in regards to the model (top) for \ourmethod as well as the baselines, which are visualized with different colors as indicated in the image legend. 
    In addition, for each scene, we showcase the target image (bottom left) along with their corresponding Novel View Synthesis (NVS) output (bottom right) of the estimated camera pose by \ourmethod.     }
    \vspace{-8pt}
\end{figure*}

\begin{figure*}[!ht]
\centering
{\fontfamily{phv}\selectfont 
\begin{minipage}[outer sep=0]{0.48\textwidth}
    \centering
    {\footnotesize Garden}
    \vspace{.5ex}
    \hrule
    \vspace{.5ex}
    \begin{minipage}[m]{\textwidth}
        \centering
        \includegraphics[width=0.8\textwidth]{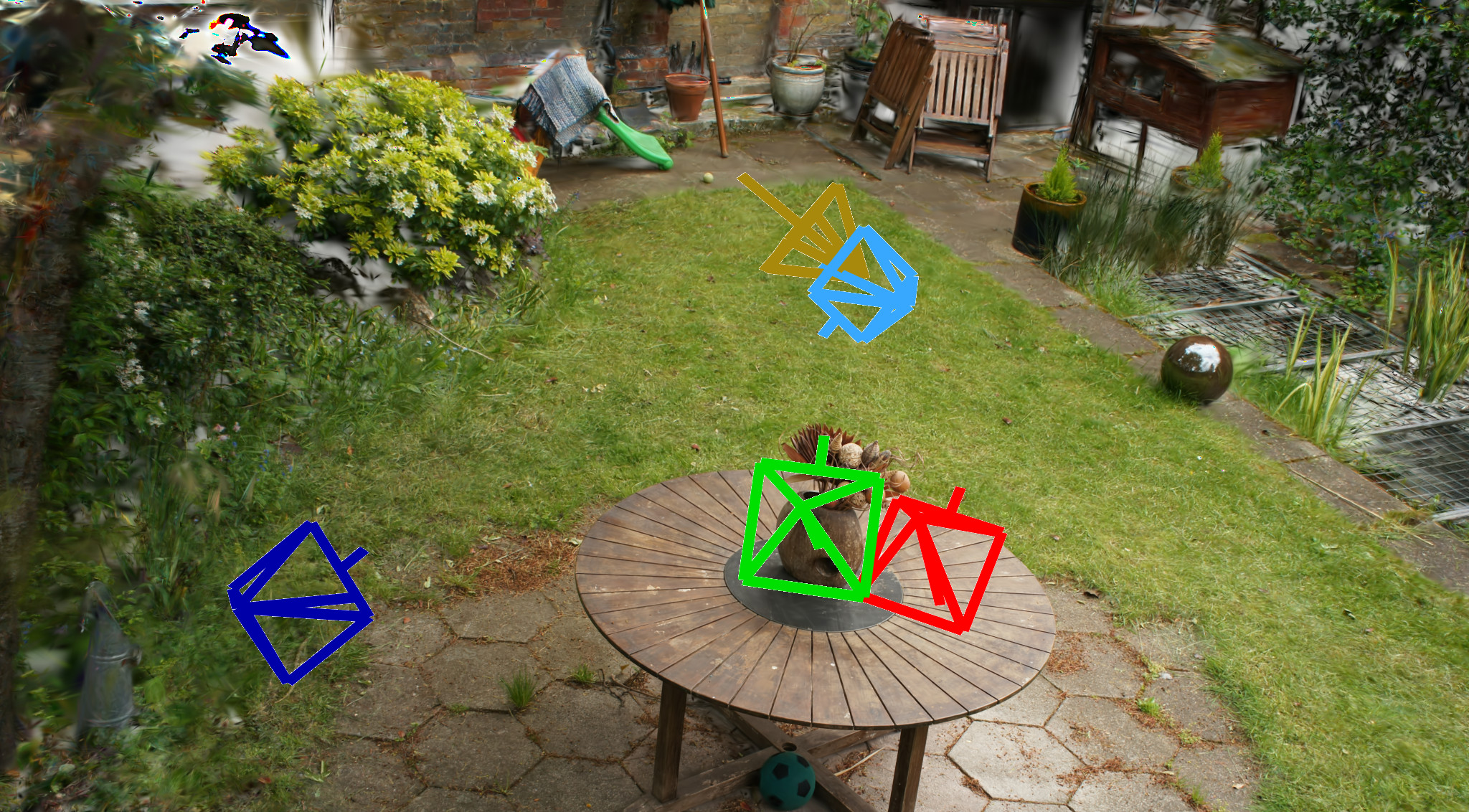} \\
    \end{minipage}
    \begin{minipage}[m]{\textwidth}
        \hspace{+15pt}
        \begin{minipage}[m]{0.49\textwidth}
        \centering
            {\scriptsize Target image} \\
            \includegraphics[width =0.7\textwidth]{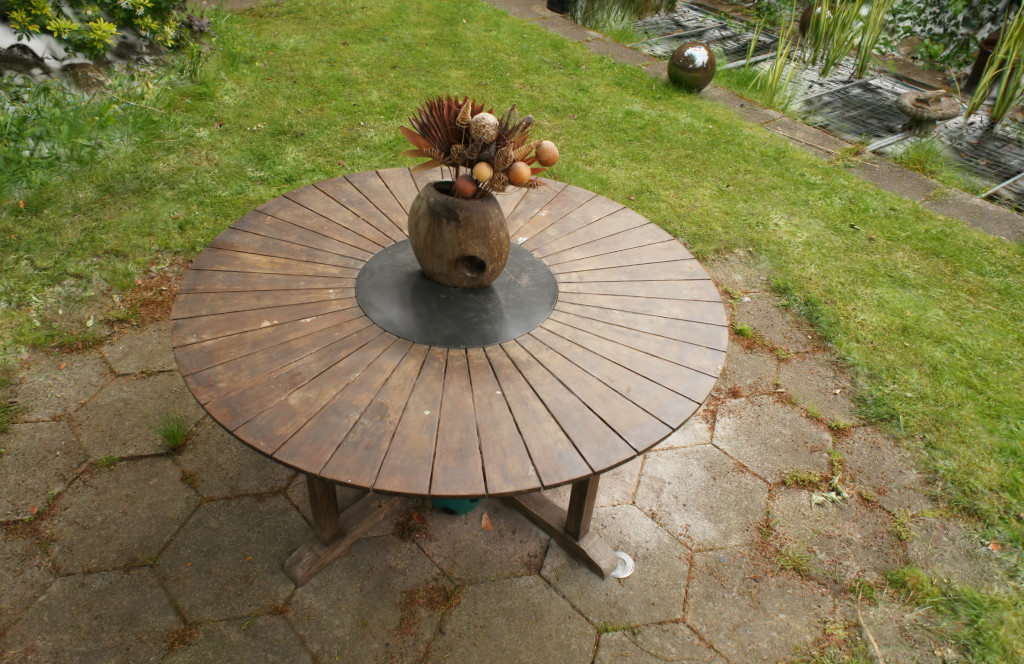}
        \end{minipage}
        \hspace{-25pt}
        \begin{minipage}[m]{0.49\textwidth}
        \centering
        {\scriptsize Estimated NVS} \\
        \includegraphics[width =0.7\textwidth]{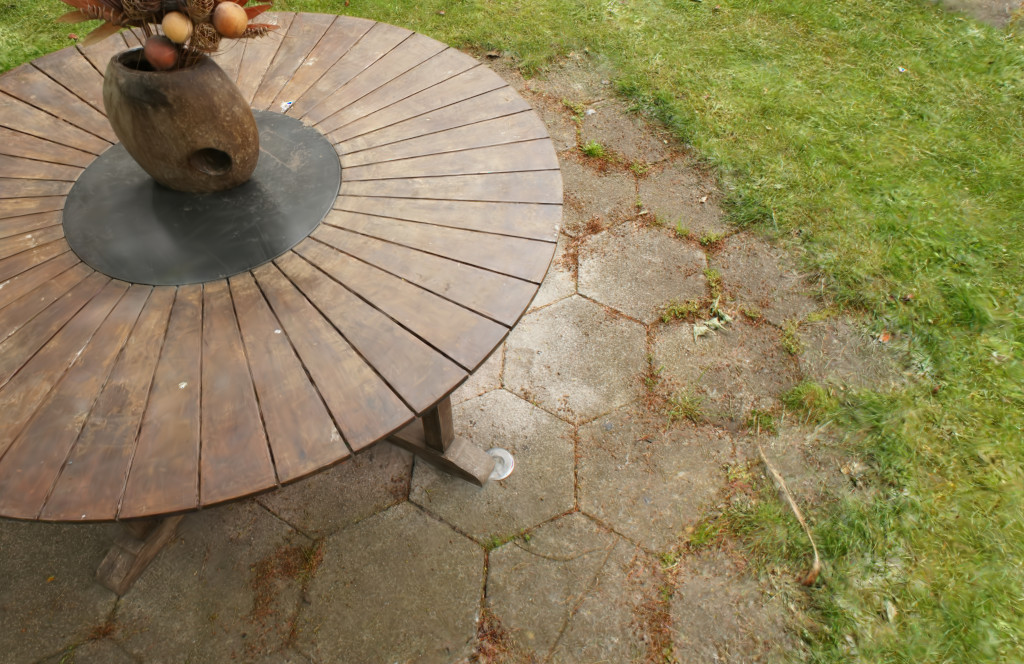}
        \end{minipage}\\[0.5ex]
    \end{minipage}\hfill
    \begin{minipage}[m]{0.6\textwidth}
    \centering
    \end{minipage}\hfill
\end{minipage}\hfill
\begin{minipage}[outer sep=0]{0.48\textwidth}
    \centering
    {\footnotesize Kitchen}
    \vspace{.5ex}
    \hrule
    \vspace{.5ex}
    \begin{minipage}[m]{\textwidth}
        \centering
        \includegraphics[width=0.83\textwidth]{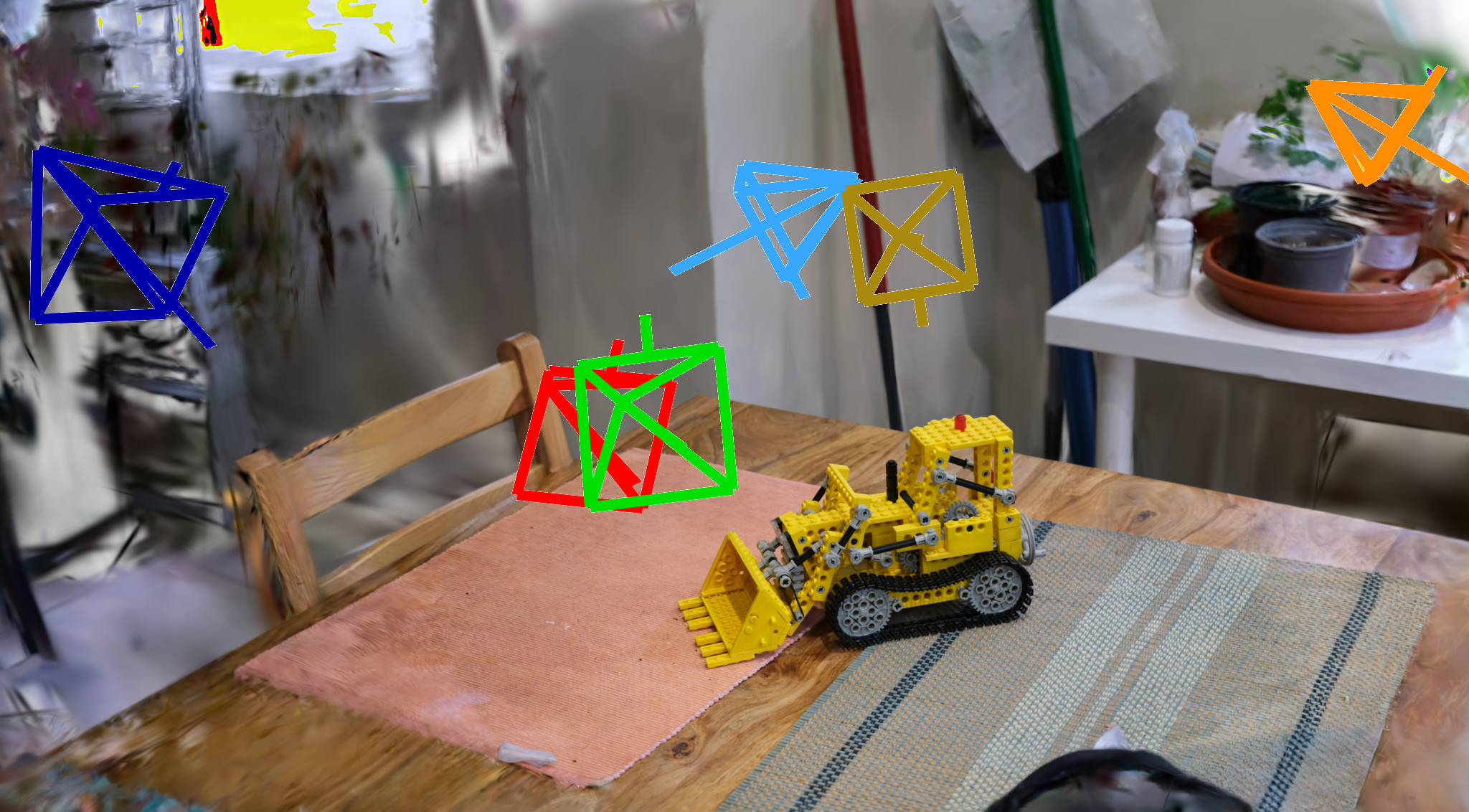} \\
    \end{minipage}
    \begin{minipage}[m]{\textwidth}
        \hspace{+7pt}
        \begin{minipage}[m]{0.49\textwidth}
        \centering
            {\scriptsize Target image} \\
            \includegraphics[width =0.7\textwidth]{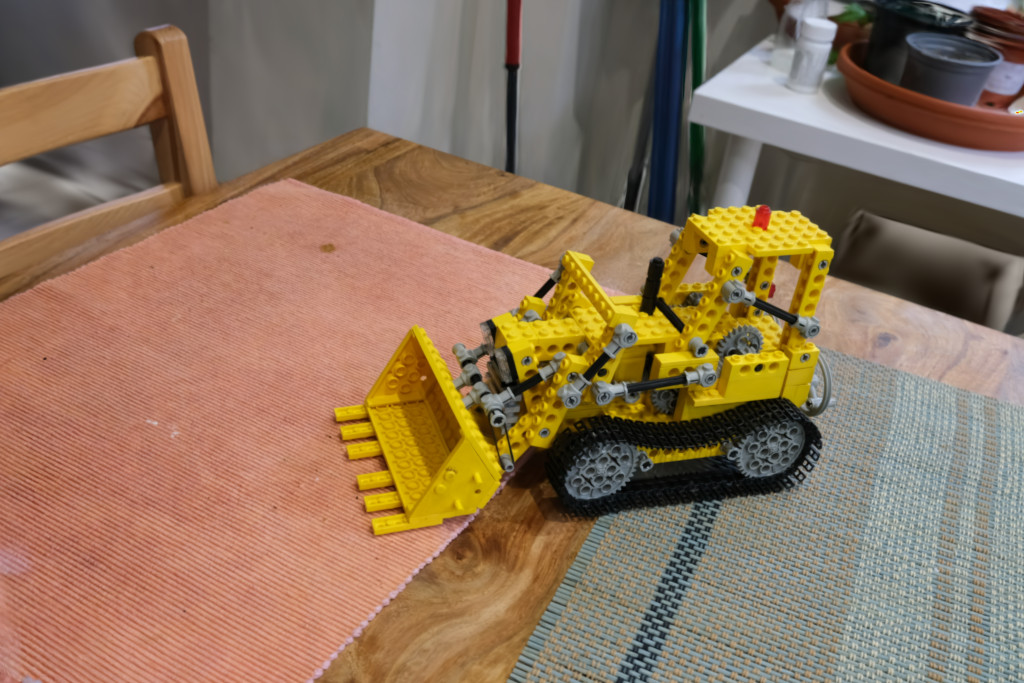}
        \end{minipage}
        \hspace{-25pt}
        \begin{minipage}[m]{0.49\textwidth}
        \centering
        {\scriptsize Estimated NVS} \\
        \includegraphics[width =0.7\textwidth]{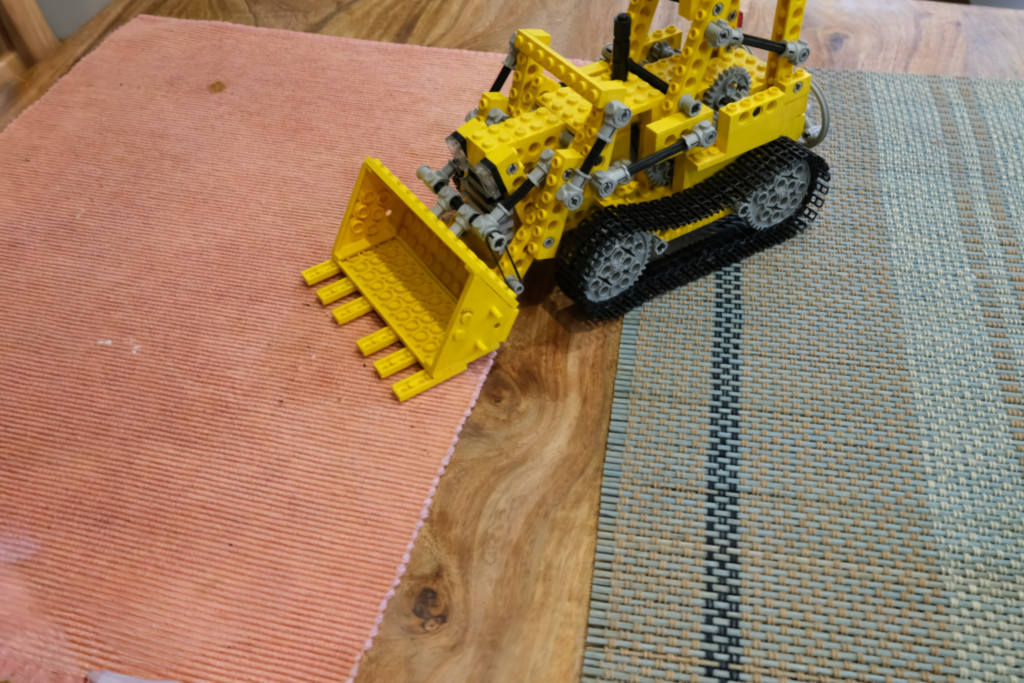}
        \end{minipage}\\[0.5ex]
    \end{minipage}\hfill
    \begin{minipage}[m]{0.6\textwidth}
    \centering
    \end{minipage}\hfill
\end{minipage}\hfill

\begin{minipage}[outer sep=0]{0.48\textwidth}
    \centering
    {\footnotesize Stump}
    \vspace{.5ex}
    \hrule
    \vspace{.5ex}
    \begin{minipage}[m]{\textwidth}
        \centering
        \includegraphics[width=0.8\textwidth]{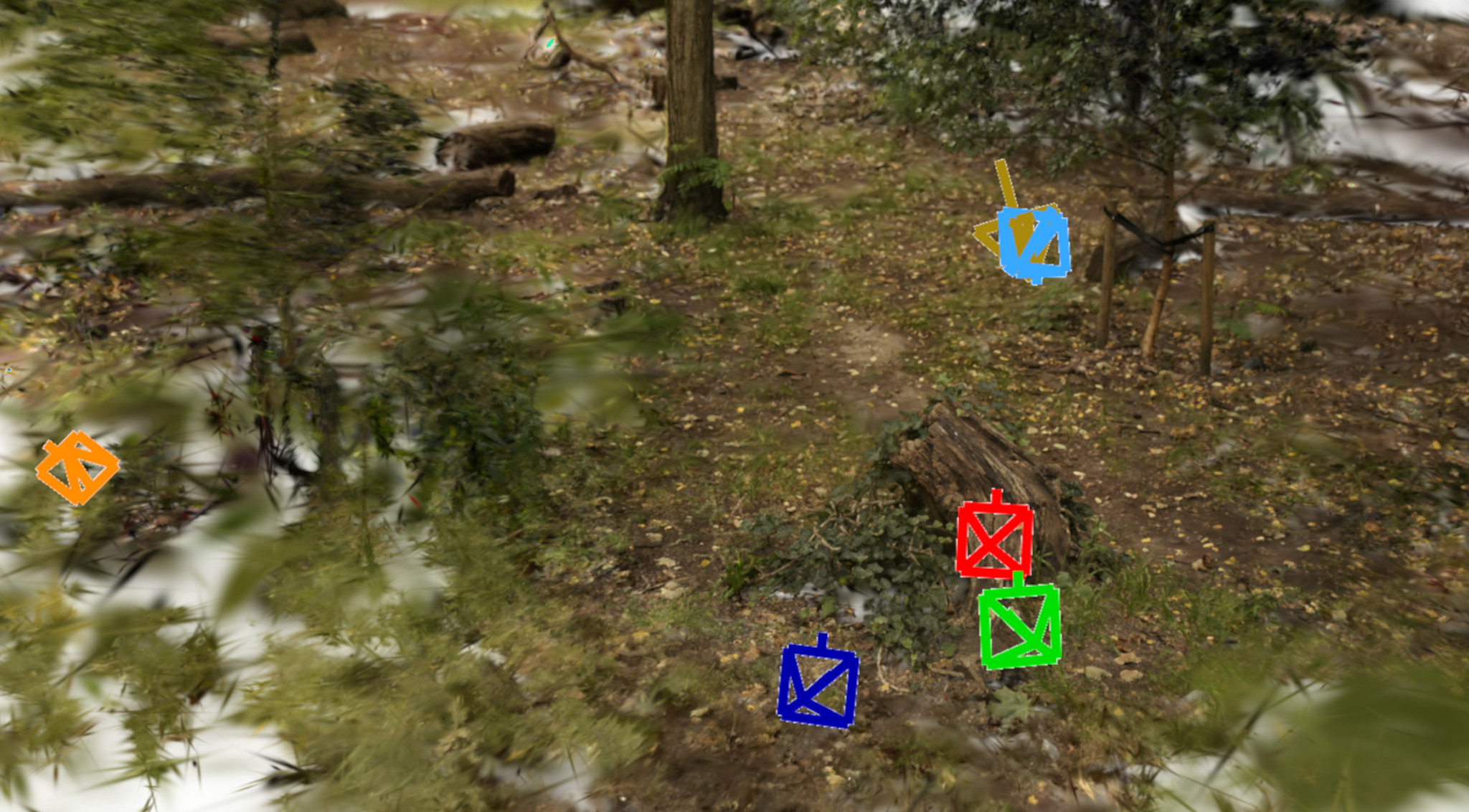} \\
    \end{minipage}
    \begin{minipage}[m]{\textwidth}
        \hspace{+7pt}
        \begin{minipage}[m]{0.49\textwidth}
        \centering
            {\scriptsize Target image} \\
            \includegraphics[width =0.7\textwidth]{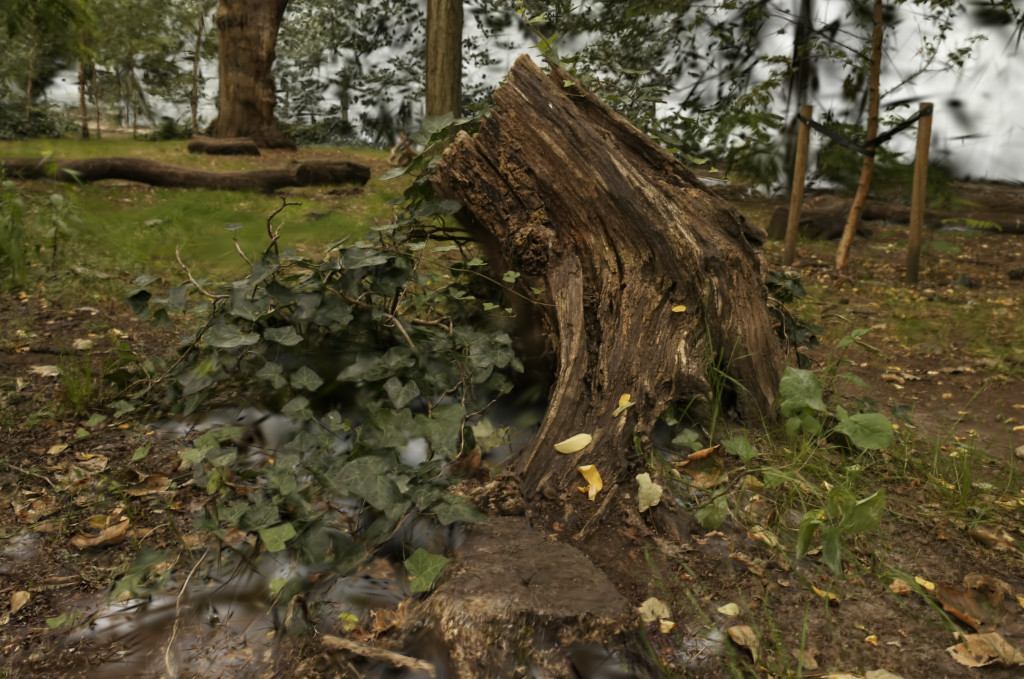}
        \end{minipage}
        \hspace{-25pt}
        \begin{minipage}[m]{0.49\textwidth}
        \centering
        {\scriptsize Estimated NVS} \\
        \includegraphics[width =0.7\textwidth]{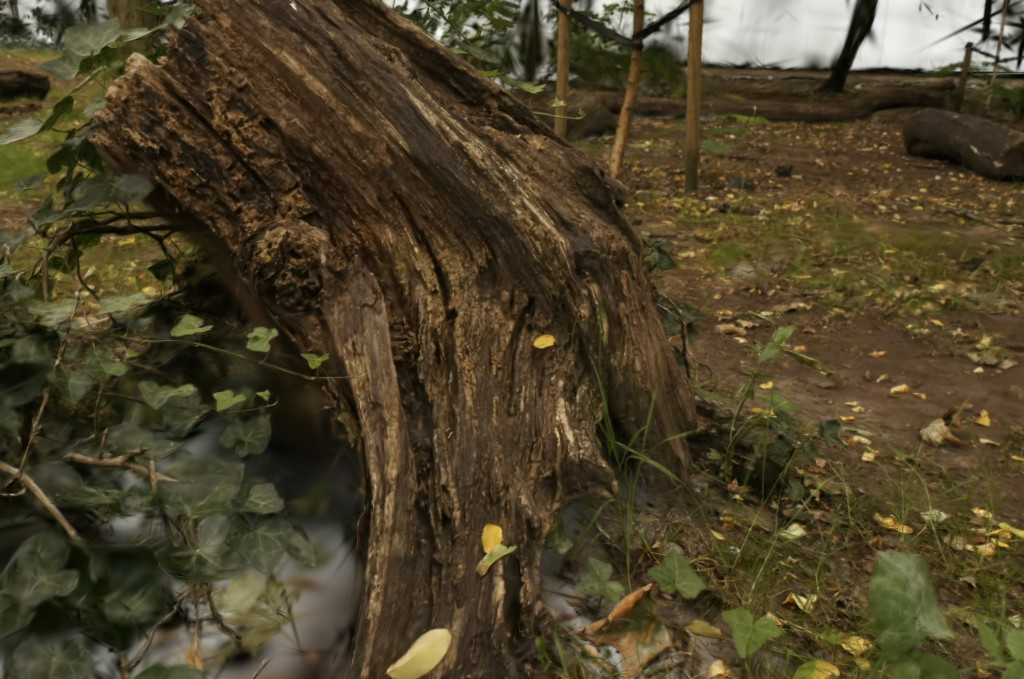}
        \end{minipage}\\[0.5ex]
    \end{minipage}\hfill
    \begin{minipage}[m]{0.6\textwidth}
    \centering
    \end{minipage}\hfill
\end{minipage}
\begin{minipage}[outer sep=0]{0.48\textwidth}
    \centering
    {\footnotesize Room}
    \vspace{.5ex}
    \hrule
    \vspace{.5ex}
    \begin{minipage}[m]{\textwidth}
        \centering
        \includegraphics[width=0.8\textwidth]{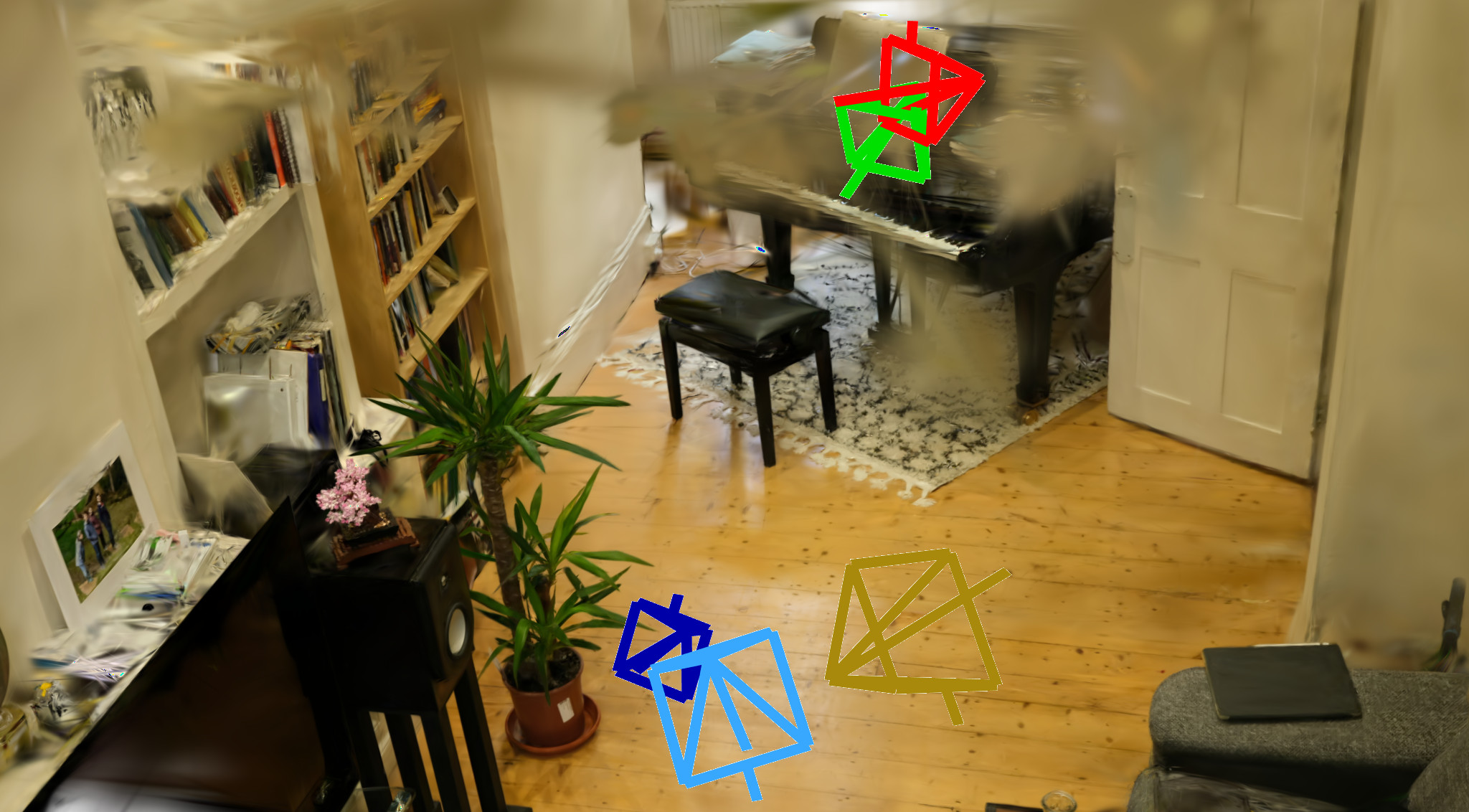} \\
    \end{minipage}
    \begin{minipage}[m]{\textwidth}
        \hspace{+7pt}
        \begin{minipage}[m]{0.49\textwidth}
        \centering
            {\scriptsize Target image} \\
            \includegraphics[width =0.7\textwidth]{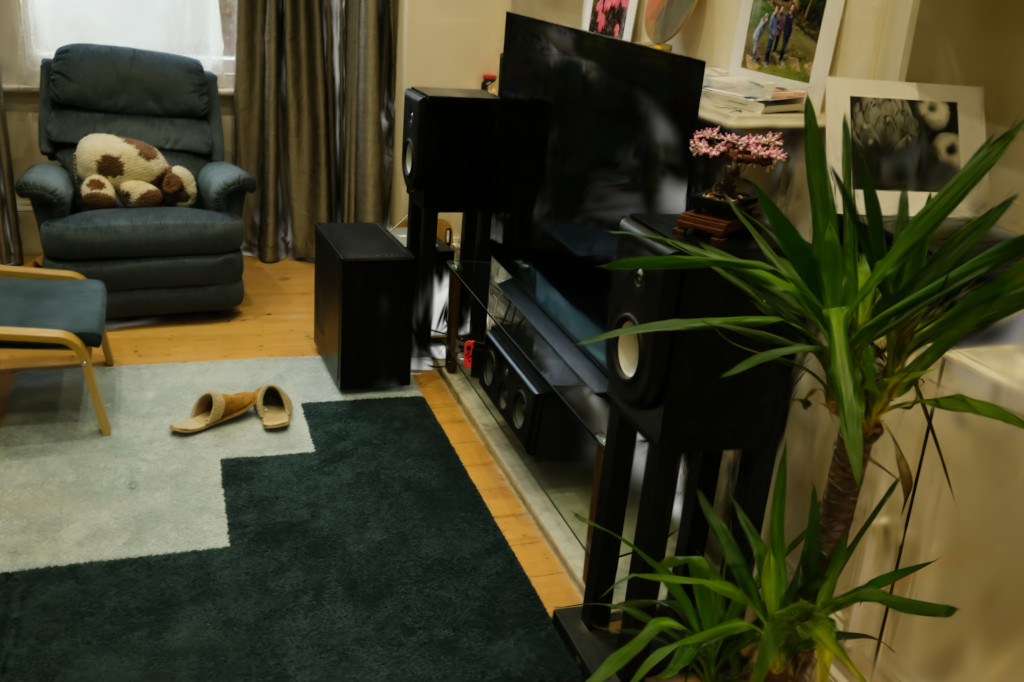}
        \end{minipage}
        \hspace{-25pt}
        \begin{minipage}[m]{0.49\textwidth}
        \centering
        {\scriptsize Estimated NVS} \\
        \includegraphics[width =0.7\textwidth]{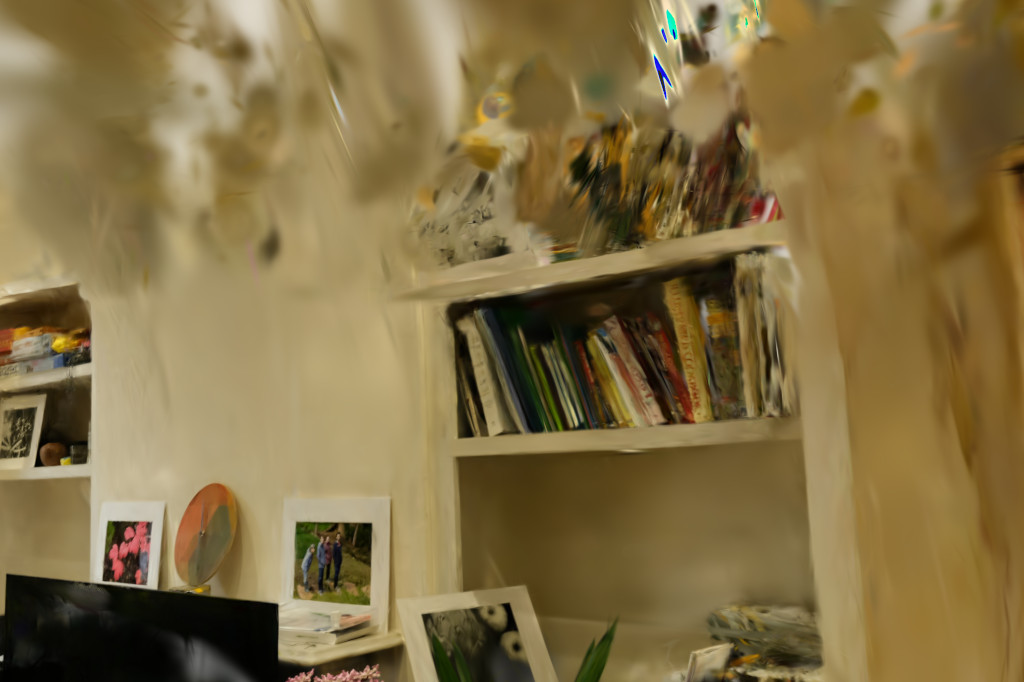}
        \end{minipage}\\[0.5ex]
    \end{minipage}\hfill
    \begin{minipage}[m]{0.6\textwidth}
    \centering
    \end{minipage}\hfill
\end{minipage}\hfill
\begin{minipage}[outer sep=0]{0.48\textwidth}
    \centering
    {\footnotesize Bicycle}
    \vspace{.5ex}
    \hrule
    \vspace{.5ex}
    \begin{minipage}[m]{\textwidth}
        \centering
        \includegraphics[width=0.8\textwidth]{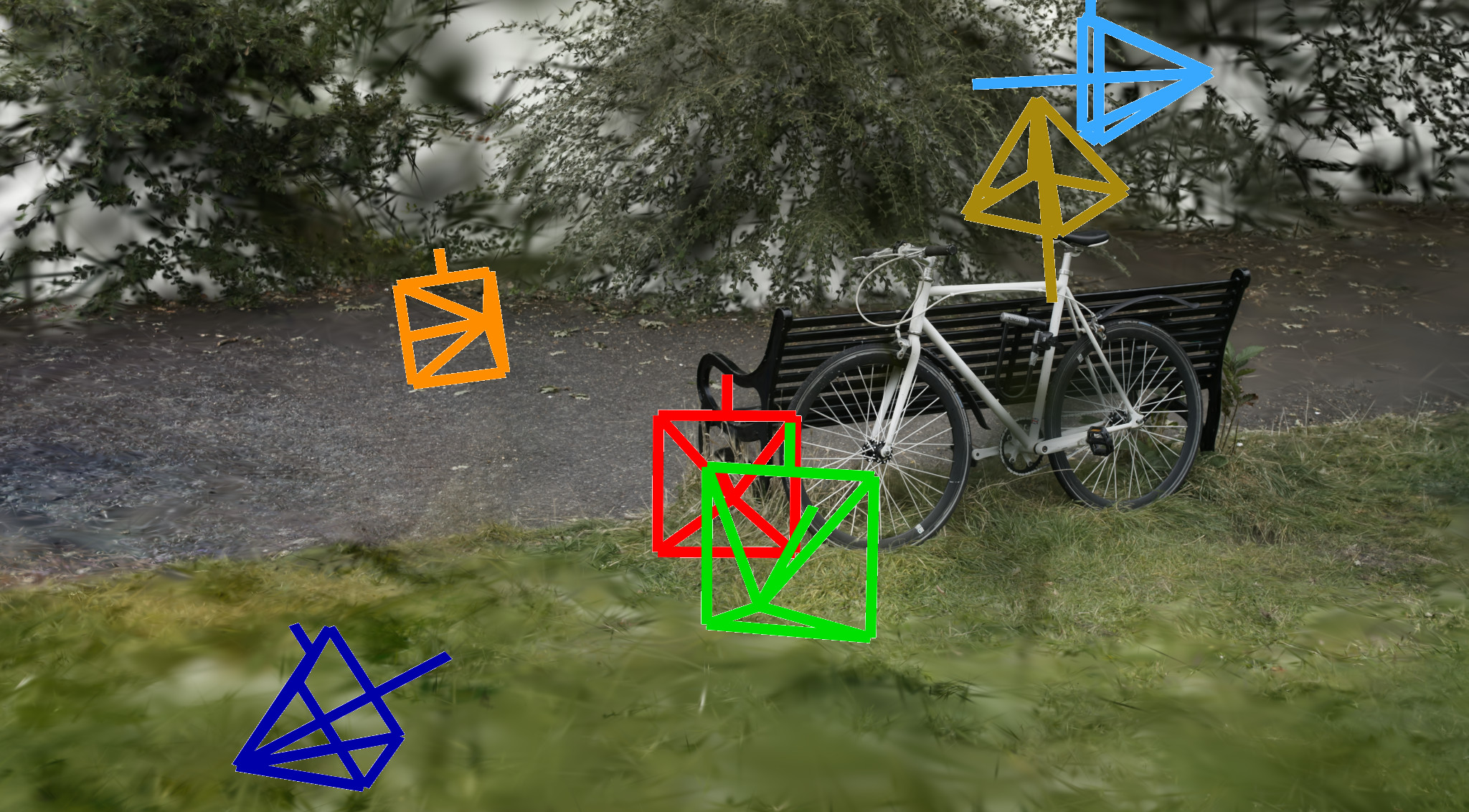} \\
    \end{minipage}
    \begin{minipage}[m]{\textwidth}
        \hspace{+7pt}
        \begin{minipage}[m]{0.49\textwidth}
        \centering
            {\scriptsize Target image} \\
            \includegraphics[width =0.7\textwidth]{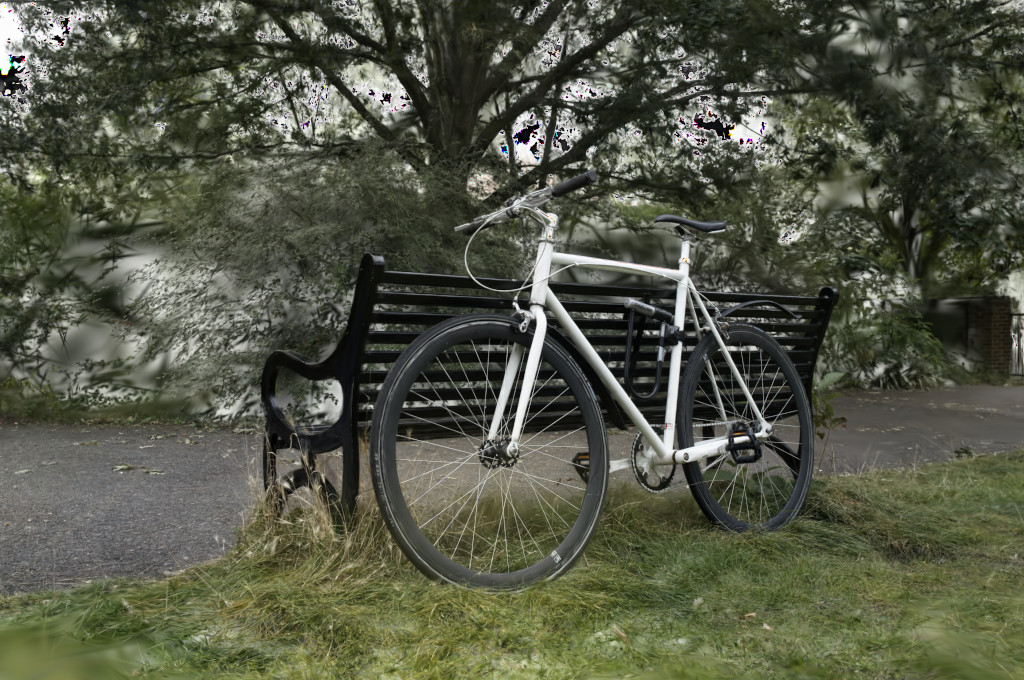}
        \end{minipage}
        \hspace{-25pt}
        \begin{minipage}[m]{0.49\textwidth}
        \centering
        {\scriptsize Estimated NVS} \\
        \includegraphics[width =0.7\textwidth]{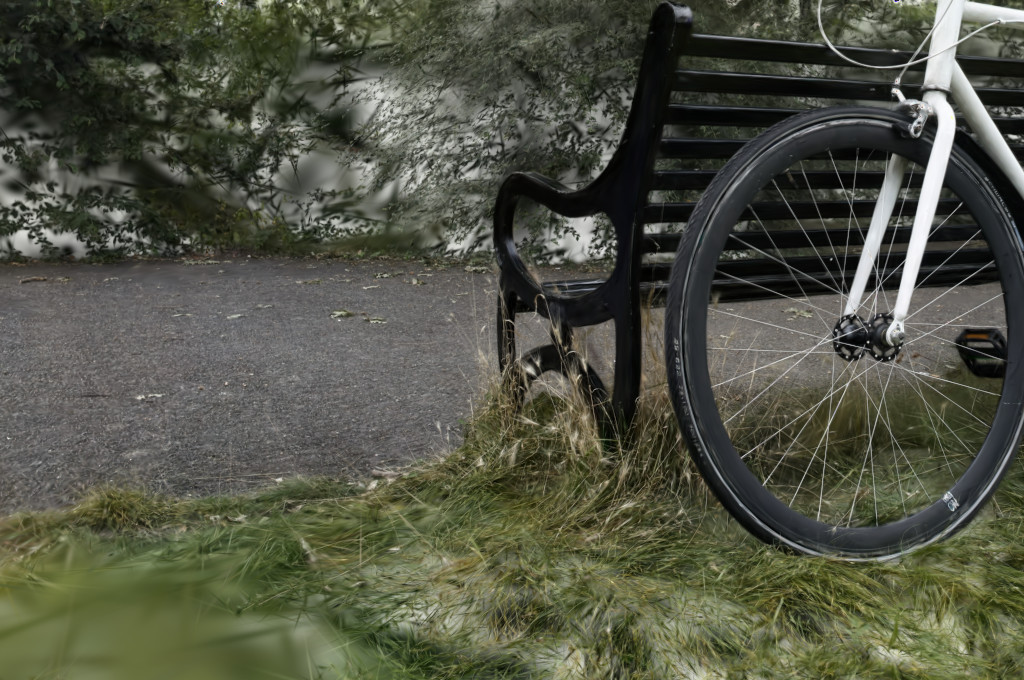}
        \end{minipage}\\[0.5ex]
    \end{minipage}\hfill
    \begin{minipage}[m]{0.6\textwidth}
    \centering
    \end{minipage}\hfill
\end{minipage}
\begin{minipage}[outer sep=0]{0.48\textwidth}
    \centering
    {\footnotesize Counter}
    \vspace{.5ex}
    \hrule
    \vspace{.5ex}
    \begin{minipage}[m]{\textwidth}
        \centering
        \includegraphics[width=0.8\textwidth]{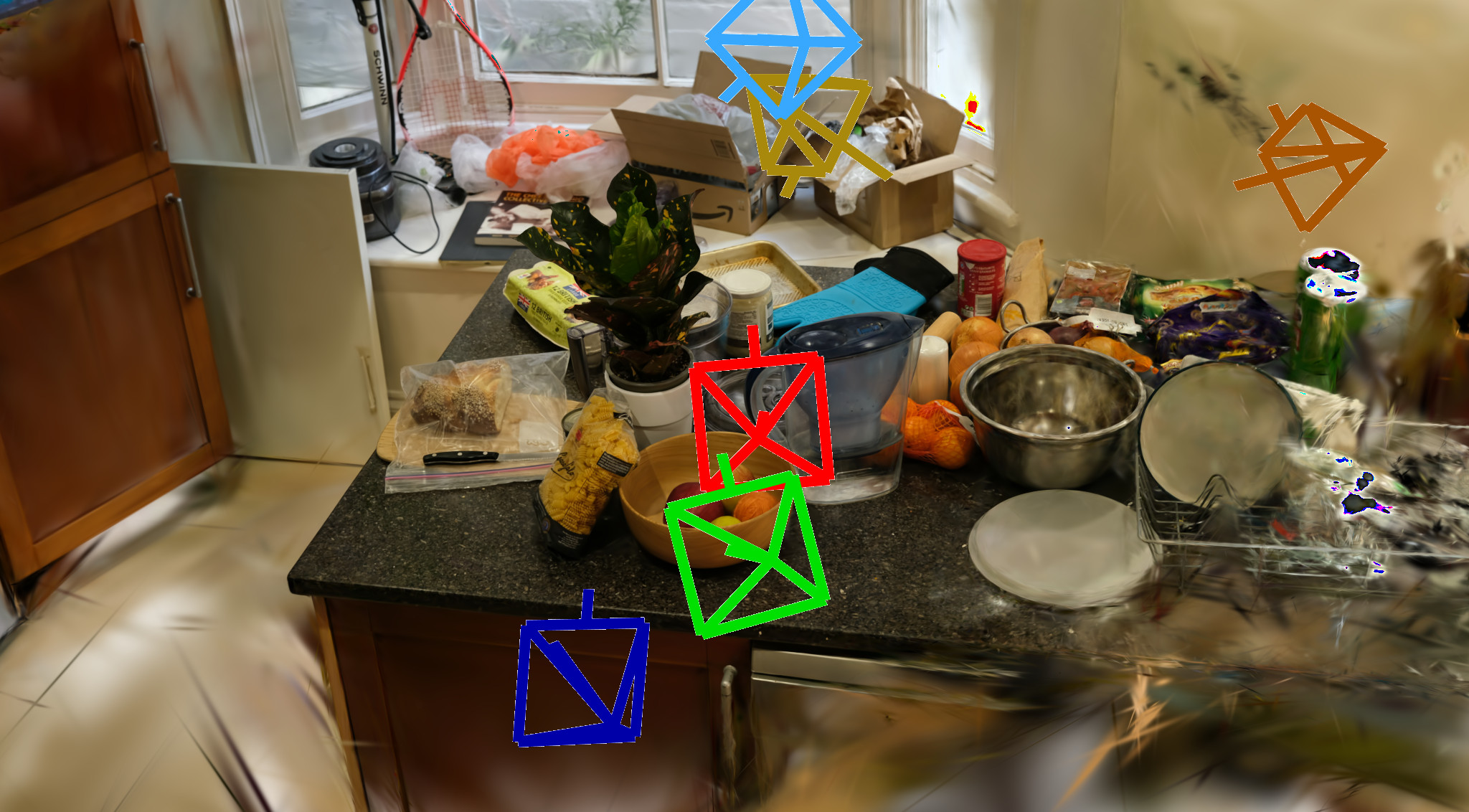} \\
    \end{minipage}
    \begin{minipage}[m]{\textwidth}
        \hspace{+7pt}
        \begin{minipage}[m]{0.49\textwidth}
        \centering
            {\scriptsize Target image} \\
            \includegraphics[width =0.7\textwidth]{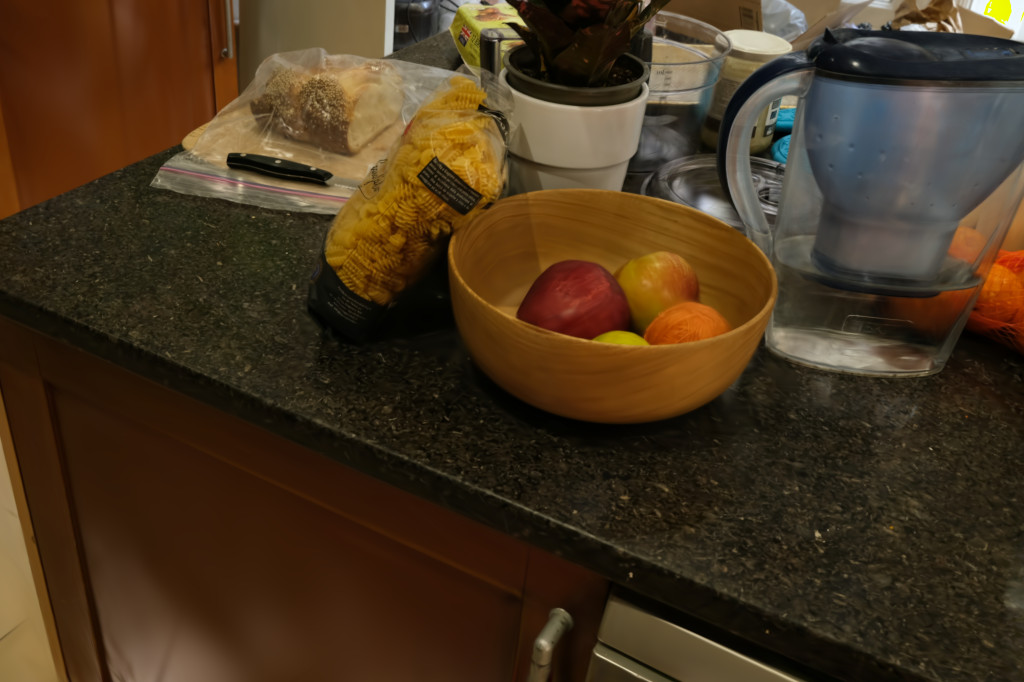}
        \end{minipage}
        \hspace{-25pt}
        \begin{minipage}[m]{0.49\textwidth}
        \centering
        {\scriptsize Estimated NVS} \\
        \includegraphics[width =0.7\textwidth]{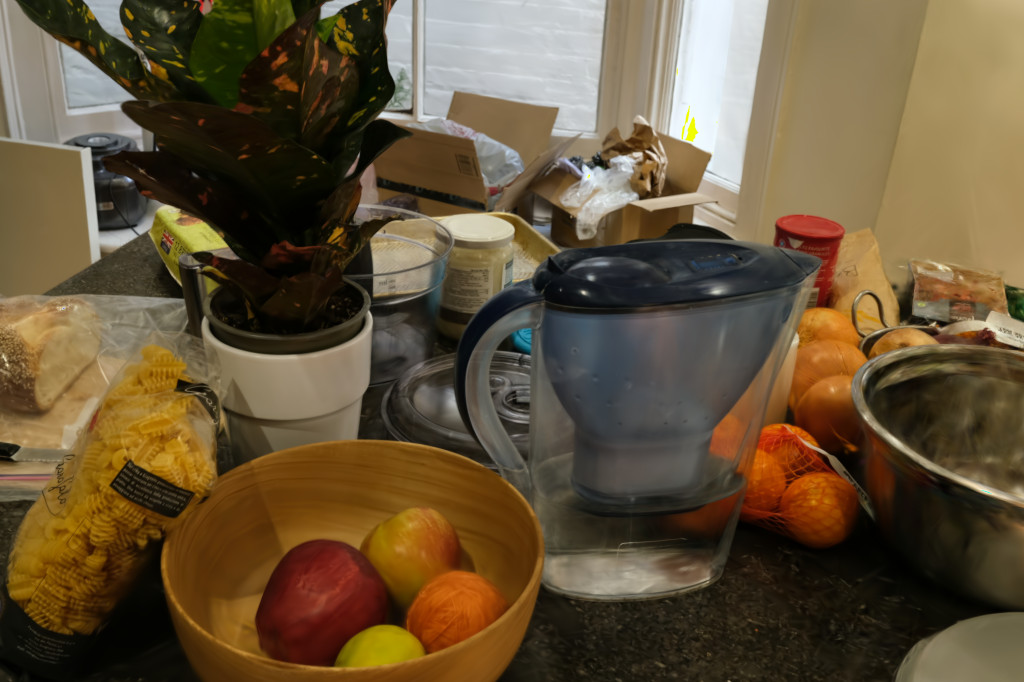}
        \end{minipage}\\[0.5ex]
    \end{minipage}\hfill
    \begin{minipage}[m]{0.6\textwidth}
    \centering
    \end{minipage}\hfill
\end{minipage}\hfill
    
        \begin{center}
    {\tiny
    \begin{tabular}{ c P{0.03\textwidth} c P{0.05\textwidth} c P{0.06\textwidth} c P{0.07\textwidth} c P{0.06\textwidth} c P{0.07\textwidth} c P{0.09\textwidth} c P{0.09\textwidth} }
     \includegraphics[width=0.04\textwidth]{imgs/qualitative_visualization/cameras/gt.png} & \vspace{-0.03\textwidth} GT & \includegraphics[width=0.04\textwidth]{imgs/qualitative_visualization/cameras/ours.png} & \vspace{-0.03\textwidth}\ourmethod (Ours) & \includegraphics[width=0.04\textwidth]{imgs/qualitative_visualization_2/inerf_wprior.png} & \vspace{-0.03\textwidth}iNeRF w/ prior & 
     \includegraphics[width=0.04\textwidth]{imgs/qualitative_visualization/cameras/inerf_random.png} & \vspace{-0.03\textwidth}iNeRF w/o prior & 
     \includegraphics[width=0.04\textwidth]{imgs/qualitative_visualization_2/pinerf_wprior.png} & \vspace{-0.03\textwidth} Parallel iNeRF w/ prior & 
     \includegraphics[width=0.04\textwidth]{imgs/qualitative_visualization_2/pinerf_woprior.png} & \vspace{-0.03\textwidth}Parallel iNeRF w/o prior & 
     \includegraphics[width=0.04\textwidth]{imgs/qualitative_visualization/cameras/nemo_voge_init.png} & \vspace{-0.03\textwidth}NeMo + VoGE w/ prior & \includegraphics[width=0.04\textwidth]{imgs/qualitative_visualization/cameras/nemo_voge_random.png} & \vspace{-0.03\textwidth}NeMo + VoGE w/o prior
     \\ 
    \end{tabular}
    }
    \end{center}
    
    }
    \vspace{-2.0ex}
    \caption[XYZ]{\label{fig:supp_qualitative_results_mip} 
    Additional scenes from the Mip-NeRF 360\textdegree dataset. 
    For each scene, we show a visualization of the camera poses in regards to the model (top) for \ourmethod as well as the baselines, which are visualized with different colors as indicated in the image legend. 
    In addition, for each scene, we showcase the target image (bottom left) along with their corresponding 3DGS Novel View Synthesis (NVS) output (bottom right) of the estimated camera pose by \ourmethod.   
    }
    \vspace{-8pt}
\end{figure*}

In this section, we report additional qualitative results of other scenes not presented in the paper.
Fig.~\ref{fig:supp_qualitative_results_tt} shows the Tanks\&Temples dataset~\cite{Knapitsch2017TanksAndTemples} scenes \textit{Barn}, \textit{Caterpillar}, \textit{Ignatius}, and \textit{Truck}.
On the other hand, Fig.~\ref{fig:supp_qualitative_results_mip} presents views from the \textit{Garden}, \textit{Kitchen}, \textit{Stump}, \textit{Room}, \textit{Bicycle}, and \textit{Counter} scenes of the Mip-NeRF 360\textdegree dataset~\cite{barron2022mipnerf360}. 
In each figure, we show the ground truth camera position (green) and our estimated camera pose (red) in addition to the baseline methods (see legend for camera colors and details).
In addition to the 3D camera positions estimated by our method and the other baselines, we also show the rendered images using the 3DGS model from ground truth and estimated \ourmethod camera pose. From the rendered images, we can assess visually how the synthesized image viewpoint is different from the tested target image. 

Looking closer at Fig.~\ref{fig:supp_qualitative_results_tt}, our method predicts poses with a more accurate rotation, aligning closer to the ground truth. 
The translation can be slightly less accurate, resulting in either being slightly closer, \eg for the \textit{Caterpillar}, \textit{Ignatius}, and \textit{Truck} scenes, or further, \eg for the \textit{Barn} scene, from the object. 
A similar estimation of accuracy in rotation can be seen for the scenes \textit{Kitchen, Stump} of the  Mip-NeRF 360\textdegree dataset, see in Fig.~\ref{fig:supp_qualitative_results_mip} while showing a closer camera position though in a more limited amount.
Mip-NeRF 360\textdegree also showcases a degradation in performance on camera rotation estimation related to the scenes \textit{Room, Stump, Garden}.
As for the baselines, in general, we can see how Parallel iNeRF tends to be more precise compared to the ones from iNeRF, but it can still fail in the complex scenes as shown.
Finally, for NeMo + VoGE, the qualitative examples are consistent with the quantitative results, which tend to produce poses with a higher error than the other approaches.

\end{document}